\documentclass[10pt,twocolumn,letterpaper]{article}

\usepackage{cvpr}              %

\providecommand{\OURS}{PhysHead\xspace}

\providecommand{\orange}[1]{{\color{orange}#1}}

\definecolor{r2color}{rgb}{0.0, 0.53, 0.74}

\definecolor{r3color}{rgb}{0.55, 0.71, 0.0}

\definecolor{cvprblue}{rgb}{0.21,0.49,0.74}
\usepackage[pagebackref,breaklinks,colorlinks,allcolors=cvprblue]{hyperref}
\usepackage{bm}
\usepackage{multirow}
\usepackage{colortbl}
\usepackage{comment}
\usepackage{tikz}
\usepackage{amsmath}
\usepackage[dvipsnames]{xcolor}
\usepackage{wrapfig}


\definecolor{green}{HTML}{E2EFDA}
\definecolor{yellow}{HTML}{FFF2CC}

\usepackage{tabularx}
\usepackage{array}
\newcolumntype{Y}{>{\centering\arraybackslash}X}

\def\paperID{36725} %
\def\confName{CVPR}
\def\confYear{2026}

\title{PhysHead: Simulation-Ready Gaussian Head Avatars}

\author{
Berna Kabadayi$^{1,3}$ \hspace{0.6em}
Vanessa Sklyarova$^{1,2}$ \hspace{0.6em}
Wojciech Zielonka$^{4}$\footnotemark[1] \hspace{0.6em}  %
Justus Thies$^{1,4}$ \hspace{0.6em}
Gerard Pons-Moll$^{3,5,6}$ \and
\vspace{0.1cm}\\
$^1$Max Planck Institute for Intelligent Systems \ \ $^2$ETH Zürich \ \ $^3$University of Tübingen \\ 
$^4$Technical University of Darmstadt \ \  $^5$Tübingen AI Center \ \ $^6$Max Planck Institute for Informatics
}

\begin{document}

\twocolumn[{%
\renewcommand\twocolumn[1][]{#1}%
\maketitle
\begin{center}
    \centering
    \captionsetup{type=figure}
    \vspace{-0.75cm}
    \includegraphics[width=1.0\textwidth]{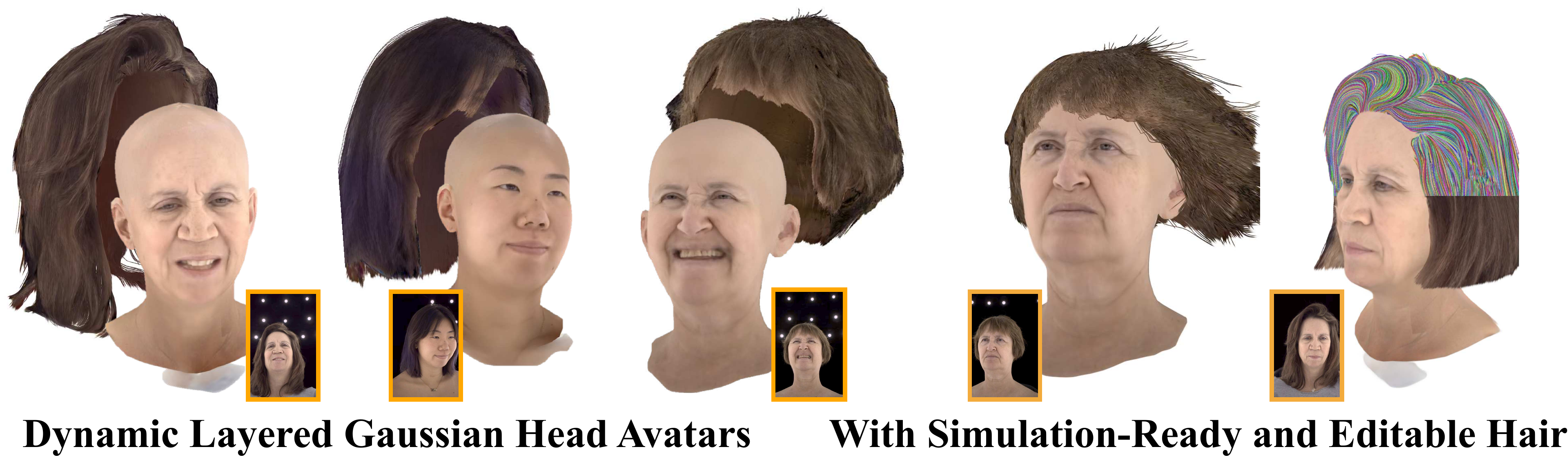}
    \caption{We introduce \OURS, a novel method for creating photorealistic head avatars with dynamic hair from multi-view videos. We leverage a vision language model to create a layered representation of the avatar, decomposed into a head and a hair layer (left). The two separate layers enable localized control of facial expressions with a 3DMM and physics-guided simulation of the hair. The strand-based representation of the hair works seamlessly with physics engines and enables hair geometry and appearance editing (right). Reference images are shown in \orange{orange}.} 
\end{center}%
}]

\footnotetext[1]{This work was conducted while Wojciech Zielonka was at TU Darmstadt. }
\begin{abstract}
Realistic digital avatars require expressive and dynamic hair motion; however, most existing head avatar methods assume rigid hair movement.
These methods often fail to disentangle hair from the head, representing it as a simple outer shell and failing to capture its natural volumetric behavior.
In this paper, we address these limitations by introducing \textbf{PhysHead}, a hybrid representation for animatable head avatars with realistic hair dynamics learned from multi-view video.
Our approach combines a 3D parametric mesh for the head with strand-based hair, which can be directly simulated using physics engines.
For the appearance model, we employ Gaussian primitives attached to both the head mesh and hair segments.
This representation enables the creation of photorealistic head avatars with dynamic hair behavior, such as wind-blown motion, overcoming the constraints of rigid hair in existing methods.
However, these animation capabilities also require new training schemes.
In particular, we propose the use of VLM-based models to generate appearance of regions that are occluded in the dynamic training sequences.
In quantitative and qualitative studies, we demonstrate the capabilities of the proposed model and compare it with existing baselines.
We show that our method can synthesize physically plausible hair motion besides expression and camera control.
For additional results and code, please refer to
\url{https://phys-head.github.io}.
\end{abstract}
    
\section{Introduction}
\label{sec:intro}
Human hair in real life is dynamic. It moves in different directions when we rotate our heads or blow in the wind.
Modeling hair dynamics for digital avatars is challenging as hair is complex, composed of thousands of tiny strands.
Thus, most data-driven approaches for animatable head avatar reconstruction often assume static hair. 
Despite being photorealistic~\cite{Giebenhain_2024, Qian2024gaussianavatars}, there is no clear separation between the head and the hair in these approaches. When the head moves, the hair moves rigidly with it. 
Recent works have addressed this problem by disentangling the hair from head~\cite{Feng2023DELTA, wang2024mega, kim2025haircup}, however, it does not solve the problem of the head avatar having static hair. 
Concurrent follow-up methods extend this approach by capturing hair dynamics data and learning hair dynamics from the data~\cite{liao2023hhavatar, liao2025hades}. While promising, it is simply not feasible to capture all possible hair dynamics and effects such as wind, since changing wind source direction would affect the hair dynamics. 
Unfortunately, these methods cannot generalize to unseen hair dynamics scenarios.

In contrast, explicit strand-based hair of digital avatars in movies and games is simulated using physical simulation systems~\cite{loki, unrealengine, maya} and the appearance is carefully crafted by artists. Strand-based representation gives artists full control over hair properties such as stiffness or damping. 
There are methods to obtain such a representation from multiview~\cite{sklyarova2023haar, zakharov2024gh, monohair, sklyarova_kabadayi_2025neuralfur} or one-shot images~\cite{hairnet, hairstep, wu2022neuralhdhair}.
Strand-based hair reconstruction methods typically focus on the geometry of the hair, and when hair strands are animated, a user-defined appearance model is manually incorporated~\cite{unrealengine}.
Since the focus is on hair geometry and appearance~\cite{zheng2025groomlighthybridinverserendering}, facial expressions are ignored.

Instead, we look for a solution with the following properties:
(1) a photorealistic head avatar which is driven by a standard 3D morphable model (3DMM), and
(2) a hair representation which is compatible with classical physics engines, while providing a photorealistic appearance.
To this end, we propose \OURS, a fully controllable head avatar. 
Specifically, we use a layered representation for the face and hair based on a 3D Gaussian appearance model. 
The dynamics of the face are driven by FLAME~\cite{FLAME:SiggraphAsia2017}, while the hair is represented by strands, and the hair dynamics are driven by a physics simulator~\cite{coumans2015bullet, maya}.
This layered representation is flexible, however, there are challenges:
(i) As our method, uses physics simulator, it generalizes to unseen head poses. However, these unseen head poses also reveal the unobserved head regions, such as ears and neck during animation. This is especially visible for actors with long hair, where the hair often covers the ears or side views.
(ii) During animation of the hair, previously occluded hair Gaussian primitives will become visible.
To address the first challenge, we propose to use a vision language model (VLM) to complete the head that is occluded by hair.
Specifically, we task the VLM~\cite{comanici2025gemini} to edit the captured images to remove the hair and generate bald images.
As the VLM does not guarantee the generation of a consistent appearance across all frames of our training sequence, where the subject shows different facial expressions, we first optimize for a shared texture of the FLAME model using differentiable rendering. We then generate bald training frames by blending the rendered shared texture with visible skin parts of the captured images.
We use these additional generated images to train a dynamic head avatar without hair.
Then, in a second stage, we optimize hair appearance as an additional layer on top of the bald avatar.
To address the second challenge, we design a color consistency loss for unobserved hair regions, ensuring that hair appearance remains consistent, even in areas like inner strands that are not directly observed in the images.

Extensive experiments show the effectiveness of our novel approach. We compare against state-of-the-art dynamic avatar methods, including the concurrent work HairCup~\cite{kim2025haircup} which also disentangles hair and skin using SDS-based training schemes. In summary, our main contributions are:
\begin{itemize}
\item A two-stage framework for reconstructing 3D head avatars with physics-guided animation of hair.
\item A method to generate dynamic bald avatars leveraging a vision language model (VLM) efficiently with differentiable rendering, then combining them with real captures through image-based blending, enabling generalization across diverse skin tones.
\item A strand-level regularization that allows occluded strands to transfer appearance from views where the hair is visible, yielding plausible hair appearance during animation.
\end{itemize}

\section{Related Work}
\label{sec:related work}
This work focuses on generating high-quality Gaussian avatars with realistic and dynamic hair, positions itself within the broader landscape of hair reconstruction, hair simulation, compositional head avatars, and holistic head avatar reconstruction.
For a comprehensive overview of photorealistic face and full-body avatars, we refer to the reports on face tracking and reconstruction~\cite{zollhoefer2018facestar}, morphable models~\cite{egger2020morphablemodels}, and neural rendering surveys~\cite{tewari2020neuralrendering,tewari2022advances}.
\subsection{Animatable Head Avatar}
A common approach to controllable avatar creation is directly optimizing the parameters of a 3DMM~\cite{FLAME:SiggraphAsia2017, Blanz1999AMM}.  Early works, such as Thies \etal~\cite{Thies2016Face2FaceRF, thies2015realtime}, employed custom second-order GPU optimizers for real-time performance, inspiring numerous follow-up methods~\cite{tewari2019fml, zielonka2022mica, qian2024versatile, taubner2024flowface}. However, PCA-based appearance models limit reconstruction quality, and typical 3DMMs omit hair geometry, hindering realistic novel-view synthesis.

To improve fidelity, several methods leverage NeRFs to capture fine structures like hair, skin, and eyes~\cite{shellnerf, sarkar2023litnerf, morf, Gafni2020DynamicNR, Zielonka2022InstantVH, nehvi_3vp, Lombardi2021MixtureOV, Gao2022nerfblendshape, An2023PanoHeadG3, Garbin2024VolTeMorph, Athar2022RigNeRF, Athar2023FLAMEinNeRF, kania2022conerf, kania2023blendfields, kabadayi2024ganavatar}. Gafni \etal~\cite{Gafni2020DynamicNR} introduced the first NeRF-based animatable avatar conditioned on expression coefficients, while others employ triangle-based deformations~\cite{Zielonka2022InstantVH}, triplanes~\cite{kabadayi2024ganavatar}, mixtures of volumetric primitives~\cite{Lombardi2021MixtureOV}, tetrahedral meshes~\cite{Garbin2024VolTeMorph}, or multi-level voxel blendshapes~\cite{Gao2022nerfblendshape}. Despite high-quality facial animation, these methods ignore hair dynamics, producing static hair during motion or lacking explicit strand modeling. \OURS addresses this by jointly modeling facial expressions and strand-based hair dynamics.

The emergence of 3D Gaussian Splatting (3DGS)~\cite{Kerbl20233DGS} led to Gaussian-based avatar methods~\cite{fan2023lightgaussian, zielonka2024gem, saito2024rgca, Giebenhain_2024, Pang2023ASHAG, navaneet2023compact3d, feng2024gaussian, Qian2024gaussianavatars, kirschstein2024ggheadfastgeneralizable3d, zielonka2025synshot, kirschstein2025avat3r, shao2024splattingavatar, xiang2024flashavatar, li2024animatablegaussians, xu2024gphm, liao2023hhavatar, wang2024mega, teotia2024gaussianheads, aneja2025scaffoldavatar}. 
Qian \etal~\cite{Qian2024gaussianavatars} rigs Gaussians to a 3DMM for real-time performance with high-quality rendering, while others embed Gaussians on the mesh surface with neural corrective fields for wrinkles and self-shadows~\cite{xiang2024flashavatar, shao2024splattingavatar, xu2024gphm, zielonka2024gem, zielonka2025synshot, kirschstein2025avat3r, taubner2024cap4, li2024animatablegaussians}. Gaussian Head Avatars (GHA)~\cite{xu2023gaussianheadavatar} first reconstructs per frame head meshes, and later deformations are learned by MLPs. An additional super-resolution module enables synthesizing high-quality avatars. Advanced methods like RGCA~\cite{saito2024rgca} learn relightable primitives for ultra-realistic rendering. However, these methods do not explicitly model hair strands, which limits their applicability for physics-based hair simulation.
This limitation extends to multi-person avatar models. GPHM~\cite{xu2024gphm} employs a neural parametric Gaussian model controlled by shape and expression coefficients, analogous to PCA-based 3DMMs. SynShot~\cite{zielonka2025synshot} leverages synthetic training data and bridges the domain gap through pivotal fine-tuning. CAP4D~\cite{taubner2024cap4} adopts a two-stage pipeline similar to~\cite{gao2024cat3d}, first regressing per-subject images using a multi-view diffusion model, then constructing drivable avatars. 
Avat3r~\cite{kirschstein2025avat3r} directly regresses Gaussian primitives in a single stage, producing high-quality, multi-view-consistent avatars. Despite these advances, none enforce strand-level hair modeling, making them unsuitable for physics-based hair simulation which is an aspect explicitly addressed by \OURS. 

\subsection{Compositional Human Modeling}

A growing body of work~\cite{zhang2023teca, Feng2023DELTA, Cha_2024_CVPR, wang2024mega, Cha_2025_CVPR, kim2025haircup, he2025head, simavatar2024, ostrek2025hairfree} decomposes human geometry and appearance into semantically distinct components (face, hair, and clothing) for modular and controllable representations. 
Independent modeling enables flexible editing and cross-person manipulations, such as hairstyle or outfit transfer.
PEGASUS~\cite{Cha_2024_CVPR} and PERSE~\cite{Cha_2025_CVPR} focus on generating animatable 3D personalized avatars with disentangled and editable facial attributes. 
DELTA~\cite{Feng2023DELTA} and TECA~\cite{zhang2023teca} leverage the parametric SMPL-X~\cite{SMPL-X:2019} model for body and face geometry while integrating volumetric representations~\cite{Mildenhall2020NeRF} to capture hair and clothing. 
More recently, MeGA~\cite{wang2024mega} represents the face using the FLAME~\cite{FLAME:SiggraphAsia2017} model with additional displacements and models hair with 3D Gaussians~\cite{Kerbl20233DGS}. 
Similarly, 3DGH~\cite{he2025head} introduces an unconditional generative model for 3D human heads with composable face and hair components using separate Gaussian primitives. Concurrent work HairCUP~\cite{kim2025haircup} builds a universal compositional prior by first modeling the bald head and then representing hair and face as disentangled Gaussian primitives. 
It uses Score Distillation Sampling (SDS)~\cite{poole2022dreamfusion} to obtain bald images. In contrast, our method benefits from recent VLM models and optimizes for shared texture UV maps from sparse views. It is personalized and employs a strand-based hair representation coupled with structured Gaussians, enabling physically plausible hair animation through a physics engine.
\subsection{Hair Modeling and Simulation} 
Strands are a widely used representation for hairstyles in modern computer graphics engines, essential for realistic rendering, editing~\cite{digitalsalon}, and simulation~\cite{Daviet2023InteractiveHS, digitalsalon, stuyck2024quaffure, zhou2023groomgen}. 
However, creating accurate strand-based hair geometry is a challenging task that typically requires skilled artists.
To simplify this process, various methods have been developed to reconstruct realistic strand-based hairstyles from monocular video~\cite{sklyarova2023neural, zakharov2024gh, luo2024gaussianhair, monohair, neuralstrands, drhair}, from multi-view images~\cite{deepmvshair, GroomCap, nam2019strand}, from single images~\cite{hairnet, hairstep, wu2022neuralhdhair, he2025perm, difflocks2025, Sklyarova2025Im2haircut}, and even from CT scans~\cite{shen2023CT2Hair}. 
For hairstyle simulation, earlier works such as~\cite{sklyarova2023haar, zakharov2024gh} leverage Unreal Engine~\cite{unrealengine} to generate realistic hair motion. 
However, these approaches often derive the appearance model directly from the simulator, which limits their ability to capture person-specific characteristics. 
Data-driven methods~\cite{hvh, neuwigs, liao2023hhavatar, liao2025hades} model hair motion by learning from multi-view images. 
While effective, these approaches are often hairstyle-specific, require large and diverse training datasets, and may fail in out-of-distribution scenarios. 
In addition, most do not incorporate dynamic physics-based priors, which can result in unrealistic motion and inaccurate collision handling.
GroomGen~\cite{zhou2023groomgen} trains a neural hair simulator using synthetic data.
In contrast, Quaffure~\cite{stuyck2024quaffure} employs physics-based self-supervised losses~\cite{Santesteban2021CVPR, Santesteban2022Snug, caphy_su2023}, removing the need for data generation. 
Although these methods achieve realistic results and generalize well across different poses and hairstyles, they are not personalized and do not explicitly model hair appearance.  

\begin{figure*}[h]
  \centering
  \includegraphics[width=\textwidth]{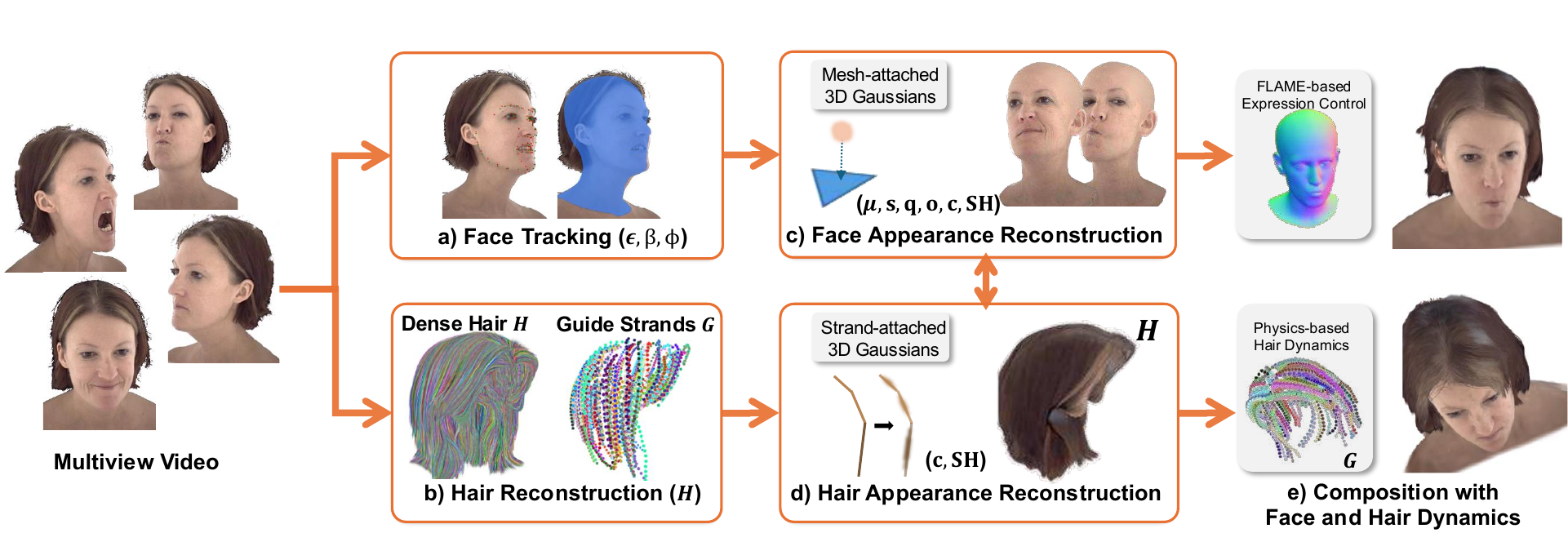}
  \caption{\textbf{Overview.} 
  PhysHead reconstructs an animatable 3D human head avatar (e) from a multiview input video. It is based on a 3D Gaussian appearance representation that is split into a face (c) and a hair region (d). The face region uses 3D Gaussians that are attached to a 3DMM-based mesh (FLAME~\cite{FLAME:SiggraphAsia2017}), which allows for parametric facial expression as well as head pose control (a,c). To enable physics-based animation of the hair region, we rely on a strand-based hair model (b). The appearance of the individual hair strands is represented as structured 3D Gaussians attached to each hair strand segment (d). 
  }
  \label{fig:pipeline}
\end{figure*}

\section{Method}
\label{sec:method}
Our method takes multiview video to learn facial expressions, along with a 360-degree static capture of a human head for hair reconstruction. 
From this data, we reconstruct a disentangled, animatable head avatar that can change expressions, be rendered from different camera views, and support physics-based animations through an attached physics simulator. 
Figure~\ref{fig:pipeline} shows an overview of our method.
As described in Section \ref{sec:layered}, we propose a layered optimization to disentangle the face from the hair.
Our representation consists of two key components: 
a strand-based hair $H$ and a dynamic face module $D$, where Gaussian primitives are attached to the parametric face model FLAME~\cite{FLAME:SiggraphAsia2017} and move consistently with different expressions, similar to recent work Gaussian Avatars (GA)~\cite{Qian2024gaussianavatars}.
As described in Section~\ref{sec:optimization}, we use photometric and silhouette-based losses and we regularize the appearance of unseen hair Gaussians using nearest neighbor strands.
After obtaining the appearance model, we use guiding hair strands and a physics engine to simulate head pose-dependent effects and transfer the motion to the dense 3D Gaussian primitives, see Section ~\ref{sec:sim}.

\subsection{Layered Representation of Human Head}
\label{sec:layered}
An essential aspect of our method is to disentangle the face and hair from real-world captures.
This is challenging, as the exact separation between head and hair in real data is unknown, and image captures show head and hair together.
Modeling head with a single layer of 3D Gaussian primitives is insufficient for disentanglement and results in artifacts, such as skin peeling off when animating the hair.
To achieve a clean layered representation, we propose a two-stage optimization of the Gaussians.
In the first stage, only the regions represented by a tracked FLAME~\cite{FLAME:SiggraphAsia2017} mesh are optimized, while regions outside FLAME’s representation, such as hair, are optimized in the second stage.
In the following, we detail the two layers, namely the expression-dependent head part and the physics-driven hair part.

\paragraph{Expression-dependent Head Part \( D \)} is represented by a set of 3D Gaussians, denoted as follows:
\[
\mathcal{G}_k = \{ \mathbf{\mu}_k, s_k, q_k, o_k, c_k, \text{SH}_k \}
\] 
The parameters are mean position \(\mathbf{\mu}_k\), scale \(s_k\), rotation \(q_k\) represented as a quaternion, opacity \(o_k\), color \(c_k\), and spherical harmonics coefficients \(\text{SH}_k\).
Following Gaussian Avatars (GA) ~\cite{Qian2024gaussianavatars}, we attach the 3D Gaussian primitives to the FLAME mesh.
Instead of optimizing the global positions of the Gaussians, we attach them to the FLAME triangles and move the triangles, allowing them to move with FLAME’s expression and motion parameters.
This binding strategy is simple but effective for modeling the face part, as it allows us full control of the head avatar using FLAME.

\paragraph{Hair Part $ H $} is represented as a collection of strands, initially obtained by NeuralHaircut~\cite{sklyarova2023neural}. 
This gives us the vertices of the hair strand polylines defined as \( H \in \mathbb{R}^{N_d \times N_{\text{seg}} \times 3} \), where \( H \) represents the dense set of hair strands, \( N_d \) is the number of dense strands, and \( N_{\text{seg}} \) is the number of segments per strand.

This strand-based reconstruction results in a large number of points.
These points are not uniformly distributed along the curves, which causes problems for simulation.
To address this, we redistribute the points along the strand and uniformly sample \( m = 60,000 \) 3D hair strands and \( n = 16 \) points per strand while preserving the strand shape.

We assign a 3D Gaussian primitive to each hair strand segment following~\cite{zakharov2024gh, GroomCap, luo2024gaussianhair} and use Frenet–Serret frames (TNB frames) to compute their rotations ($\mathbf{g_R}=[TNB]$) using the tangent $\mathbf{T}$, normal $\mathbf{N}$, and binormal $\mathbf{B}$.
Given two points \( p_1 \) and \( p_2 \) of a segment, we define an elongated Gaussian with the following properties: 
\[
\mathbf{g_{\text{mean}}}(p_1, p_2) = (p_1 + p_2)/2, \mathbf{g_{\text{scale}}} = \left( \|p_2 - p_1\|, k, k \right)
\] where \( k = 0.0001 \).
These Gaussians are splatted using a differentiable tile-based rasterizer~\cite{Kerbl20233DGS,Zwicker2001SurfaceS} and are supervised by ground truth images.
In addition to color and perceptual losses, we use a strand-level color consistency loss to regularize the appearance of unseen hair, see Section~\ref{sec:optimization}.
\subsection{Generation of Bald Training Images}
\label{sec:vlm_training_data_generation}
\begin{figure}[tb]
  \centering
    \includegraphics[width=\linewidth, keepaspectratio]{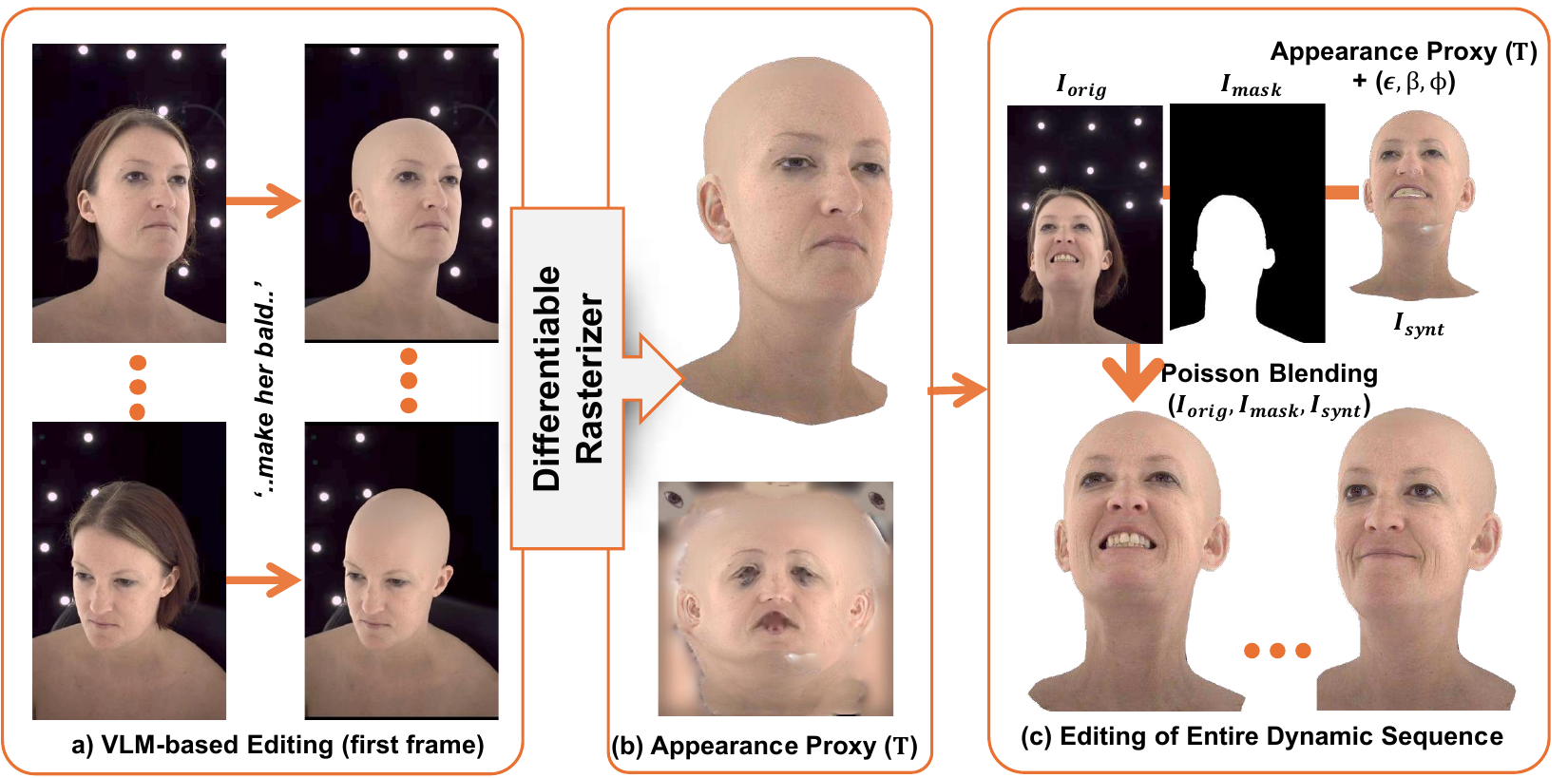}
  \caption{We use a VLM-based editing to remove the hair from the first frame of the multi-view input sequence (a). From this, we construct an appearance proxy (b) based on differentiable rendering of the FLAME head model. We use this appearance proxy to remove the hair part in the entire dynamic sequence (c) by employing Poisson image editing~\cite{perez03poisson}. These new images are used as targets for the Gaussian splatting based appearance optimization of the head layer.}
  \label{fig:vlm_target_views}
\end{figure}

To train the head layer, we remove hair from the training sequences using VLM-based editing~\cite{comanici2025gemini}, shown in \Cref{fig:vlm_target_views}.
Specifically, we use Nano-Banana~\cite{comanici2025gemini} to automatically remove the hair from the first frame and filter views that are not multiview consistent. 
Based on these sparse multiview consistent views, we use a differentiable rasterizer~\cite{qian2024versatile, Laine2020diffrast} to optimize a shared texture map of the FLAME model, referred to as \( T \in \mathbb{R}^{2048 \times 2048 \times 3} \). 

As the FLAME model~\cite{FLAME:SiggraphAsia2017} does not properly represent the geometry of the mouth interior, the textures also might contain artifacts in these regions (e.g., baked-in teeth on the lips, etc.).
Despite this, texture map \( T \) serves as an appearance proxy of unobserved regions, i.e., regions covered by hair such as the ears or the scalp, as these regions do not deform a lot compared to the facial expressions.
To this end, we render the appearance proxy \( T \) with the head pose and expression of a particular video frame and blend it with the face region of the original image using dilated and blurred segmentation masks. 
We employ a blending scheme that is based on Poisson image editing~\cite{hairmapper, perez03poisson}.
In contrast to HairCUP~\cite{kim2025haircup}, this scheme allows optimization for a head layer without strong boundary artifacts and generalizes to diverse skin tones.

\subsection{Simulation of Hair Strands}
\label{sec:sim}
For simulation of the hair, we use sparse guiding strands attached to the scalp, and create a hair particle system.
We use tracked FLAME head poses to guide our character. In our framework, we employ a physics engine \cite{coumans2015bullet} that integrates rigid-body dynamics using a semi-implicit Euler method. Newton's second law governs the translational dynamics:
\[
\frac{d\mathbf{v}}{dt} = \frac{\mathbf{F}}{m}, \quad \frac{d\mathbf{x}}{dt} = \mathbf{v} .
\]
with discrete updates given by:
\begin{align}
\mathbf{v}(t+\Delta t) &= \mathbf{v}(t) + \frac{\mathbf{F}(t)}{m}\,\Delta t, \\
\mathbf{x}(t+\Delta t) &= \mathbf{x}(t) + \mathbf{v}(t+\Delta t)\,\Delta t.
\end{align}
An iterative constraint solver is utilized to resolve collisions and to enforce joint or contact constraints, ensuring a stable and realistic simulation. Here, \( \mathbf{x} \) is the particle position, \( \mathbf{v} \) is the velocity, and \( \mathbf{F} \) is the applied force vector.

\begin{figure}[tb]
  \centering
    \includegraphics[width=1.0\linewidth, keepaspectratio]{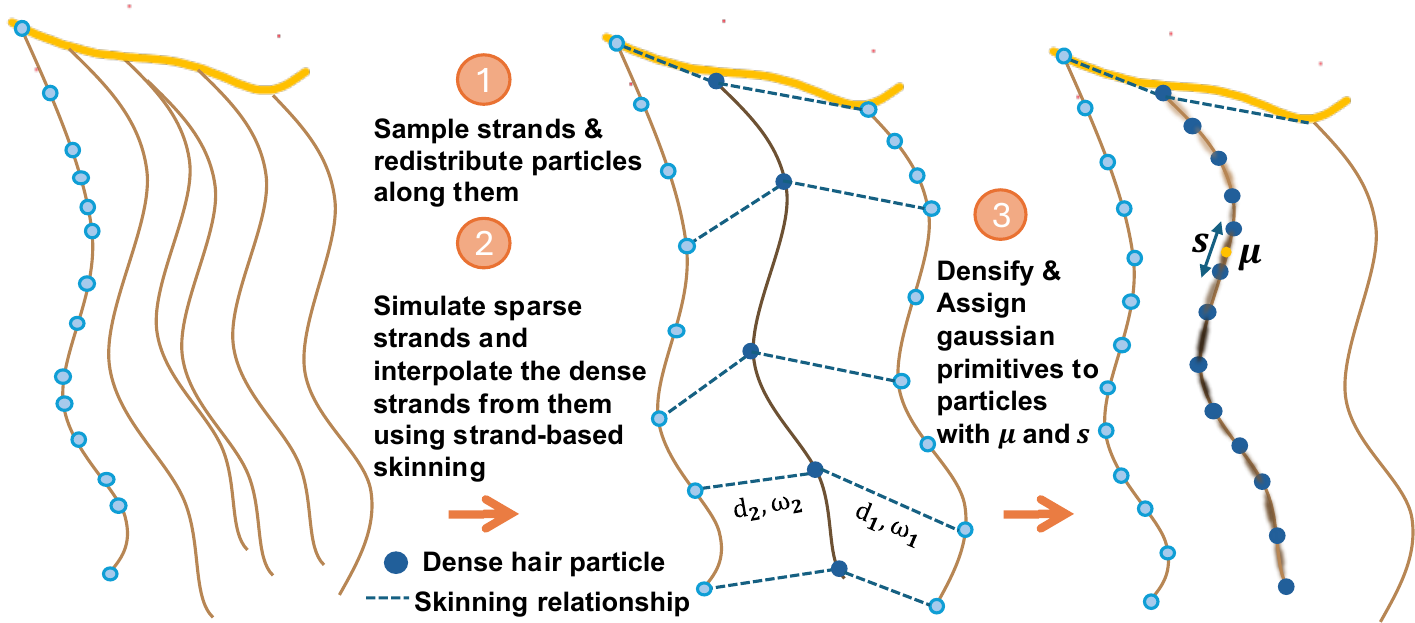}
  \caption{Using strand-skinning~\cite{chai2016adaptive}, the dense strands are affected by \( k = 10 \) sparse strands. The skinning weights are inversely proportional to the distance from the sparse hair strands. Specifically, if \( d_1 \) is smaller than \( d_2 \) in the figure, then \( \omega_1 \) is larger than \( \omega_2 \).}
  \label{fig:skinning}
\end{figure}

\paragraph{From sparse simulation to a dense hairstyle.}  
The physics simulator~\cite{maya} gives us sparse segment locations per frame that require interpolating guide strands to complete the groom~\cite{lyu2020real}. 
Between the dense hair strands and the sparse guiding hair strands, we find the k-nearest neighbor strands and transfer only the relative displacement instead of the absolute locations of the guiding hair particles, see ~\Cref{fig:skinning}.

\subsection{Optimization and Regularization}
\label{sec:optimization}
First, we only optimize the head appearance layer with the generated training sequence (\Cref{sec:vlm_training_data_generation}). Then, both the head and the hair appearance layer are rendered  jointly, and only the hair part is optimized to match the original training data, see Figure~\ref{fig:pipeline}.
For the head layer, we optimize FLAME-attached 3D Gaussians similar to ~\cite{Qian2024gaussianavatars}.
However, we incorporate facial masks and ignore the regions which FLAME cannot represent (i.e. hair).
Specifically, we adapted the tracker from GA~\cite{Qian2024gaussianavatars} to only track the facial part without hair.
The optimization of the 3D Gaussian primitives is supervised with a combination of $\mathcal{L}_1$ and D-SSIM terms.
\begin{equation}\label{eq:rgb_loss}
\mathcal{L}_{\text{rgb}} = (1 - \lambda) \cdot \mathcal{L}_1 + \lambda \cdot \mathcal{L}_{\text{D-SSIM}}, \quad  \text{with}  \quad \lambda = 0.2.
\end{equation}
Similar to GA~\cite{Qian2024gaussianavatars}, to reduce spiky Gaussian primitives, we apply the position and scaling regularizations, resulting in the loss function for the bald head part: 
\begin{equation}\label{eq:total_loss}
\mathcal{L} = \mathcal{L}_{\text{rgb}} + \lambda_{\text{pos}}\mathcal{L}_{\text{pos}} + \lambda_{\text{scaling}}\mathcal{L}_{\text{scaling}}.
\end{equation}
From this optimization stage, we obtain the head layer, which is denoted as  $ D $, where all 3D Gaussian parameters \( \mathcal{G}_k = \{ \mathbf{\mu}_k, s_k, q_k, o_k, c_k, \text{SH}_k \} \) are optimized for.

\medskip

To recover the hair appearance, we initialize the geometry with NeuralHaircut~\cite{sklyarova2023neural}. 
We begin optimizing hair appearance using the photometric rendering loss for 3000 iterations. As we have strand-based representation and hair is animatable by a physics-based engine, having skin color in the hair region is undesired. To do so, we compute the loss \ref{eq:rgb_loss} only on the masked hair region. This prevents having fewer artifacts near hairtips, in contrast to \cite{kim2025haircup, zakharov2024gh}. Given $M_{\text{hair}}$ is the binary hair mask, $I^{\text{rend}}$ is rendered 3D Gaussians, $I^{\text{gt}}$ ground truth image, formally the loss is defined as:
\begin{equation}\label{eq:rgb_loss}
\mathcal{L}_{\text{rgb}}
=
\left\|
\left( I^{\text{rend}} - I^{\text{gt}} \right) \odot M_{\text{hair}}
\right\|_{1} .
\end{equation}
While this recovers the appearance of the outer hair layer, it leads to arbitrary colors in unobserved regions—such as back views or interior strands—due to the lack of supervision.
To mitigate random colors of hidden strands, we introduce a regularization that encourages neighboring strands 
$j \in \mathcal{N}(i)$ of each strand $i \in \mathcal{S}$ to have similar colors.
The hair appearance is being diffused into unseen regions with:
\begin{equation}\label{eq:color_consistency}
\mathcal{L}_{\text{consistency}}
= 
\sum_{i \in \mathcal{S}}
\sum_{j \in \mathcal{N}(i)}
\left\| \mathbf{c}_i - \mathbf{c}_j \right\|_2^2,
\end{equation} where $\mathbf{c}_i \in \mathbb{R}^{49 \times 3}$ is the color of strand $i$.
The final objective for hair appearance optimization is:
\begin{equation}\label{eq:total_loss}
\mathcal{L}_{\text{hair}} = \mathcal{L}_{\text{rgb}} + \lambda_{\text{consistency}}\mathcal{L}_{\text{consistency}} .
\end{equation}
During the optimization, the 3D Gaussian parameters \( s_k \), and \( q_k \) are initialized and kept fixed from the TNB frames~\cite{wang2008computation}, \( o_k = 1 \), and only \( c_k \) and  \( \text{SH}_k \) are optimized.

\begin{figure}[tb]
    \centering
    \begin{minipage}{0.05\linewidth}
        \centering
        \footnotesize
        \rotatebox{90}{t=7 \hspace{5em} t=0 \hspace{7em} t=23 \hspace{5em} t=0}
    \end{minipage}%
    \begin{minipage}{0.9\linewidth} %
        \centering
        \includegraphics[width=\linewidth, keepaspectratio]{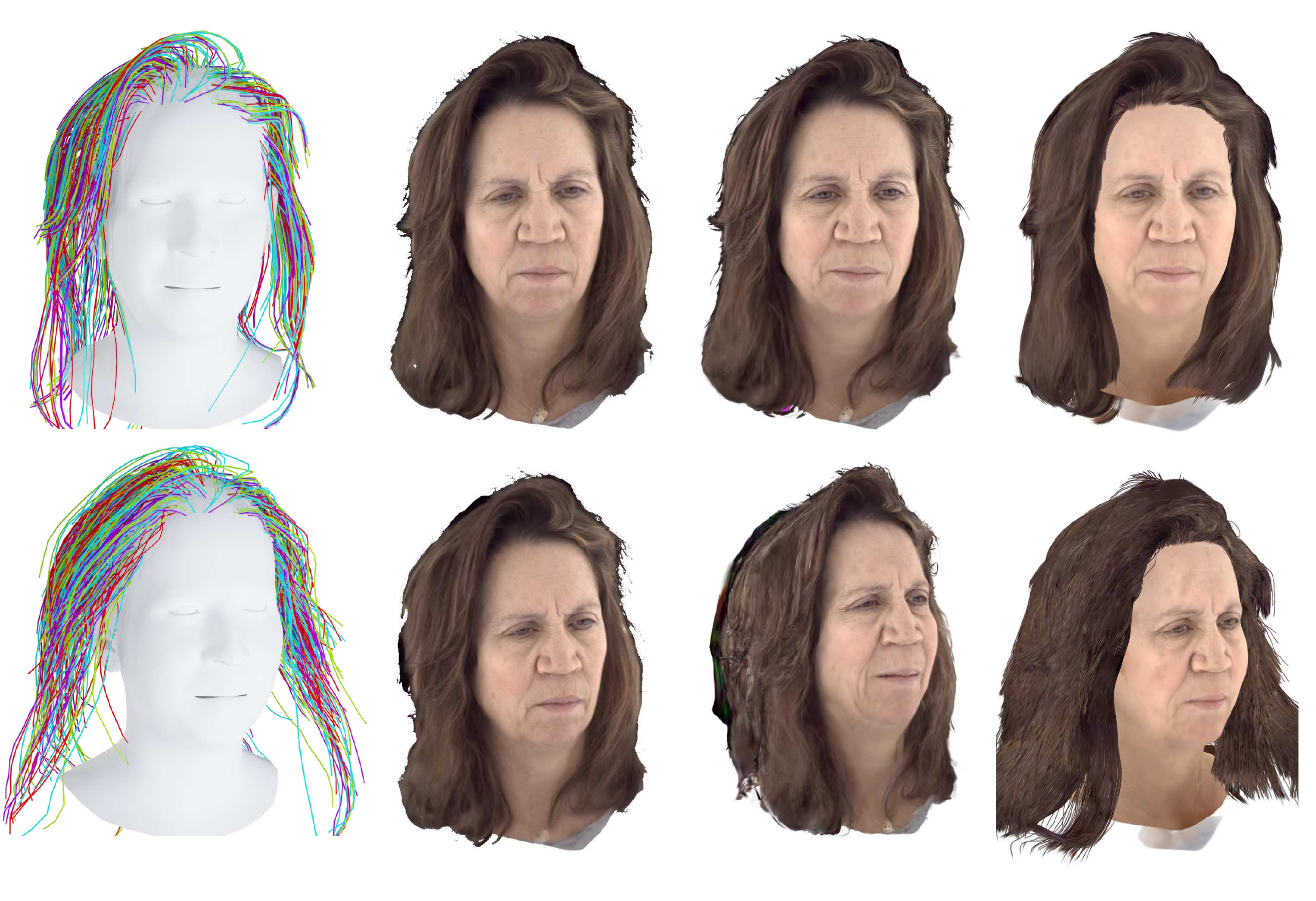}\\
        \includegraphics[width=\linewidth, keepaspectratio]{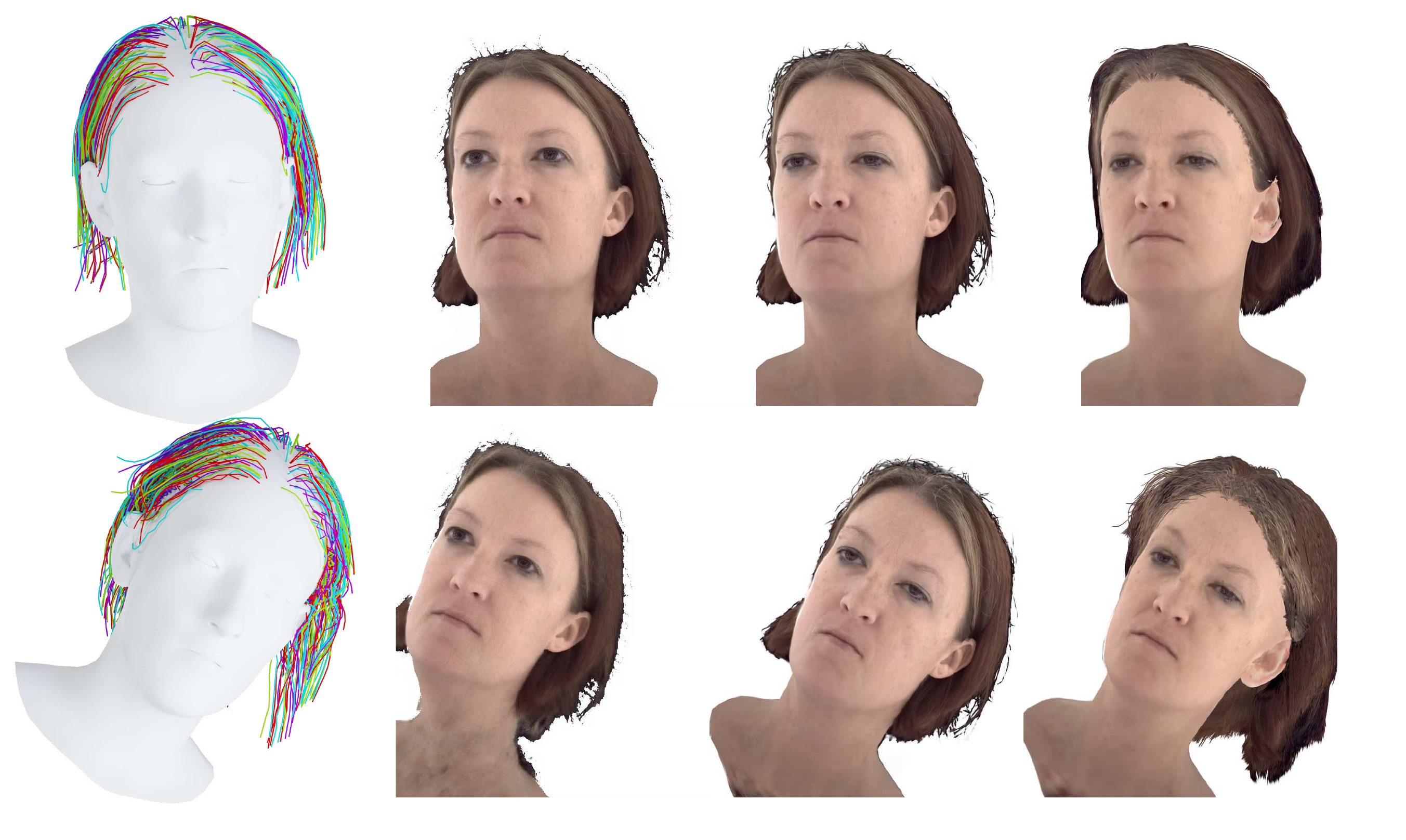}
    \end{minipage}
    
    \vspace{0.5em}
    
    \resizebox{\linewidth}{!}{
    \begin{tabularx}{\linewidth}{YYYY}
        Driving & 
        GHA~\cite{xu2023gaussianheadavatar} & 
        GA~\cite{Qian2024gaussianavatars}& 
        Ours 
    \end{tabularx}
    }
    
    \caption{\textbf{Qualitative Comparison} All methods synthesize photorealistic avatars. However, both GHA~\cite{xu2023gaussianheadavatar} and GA~\cite{Qian2024gaussianavatars} have rigid hair when animated, as Gaussians are unstructured. In contrast, our method, PhysHead, benefitting from strand representation, generalizes under novel driving signals.}
    \label{fig:maincomparison}
\end{figure}

\section{Results}
\label{sec:result}

\begin{figure*}[ht!]
  \centering    
  \includegraphics[width=1\linewidth, trim=0cm 40cm 0cm 0cm, clip]{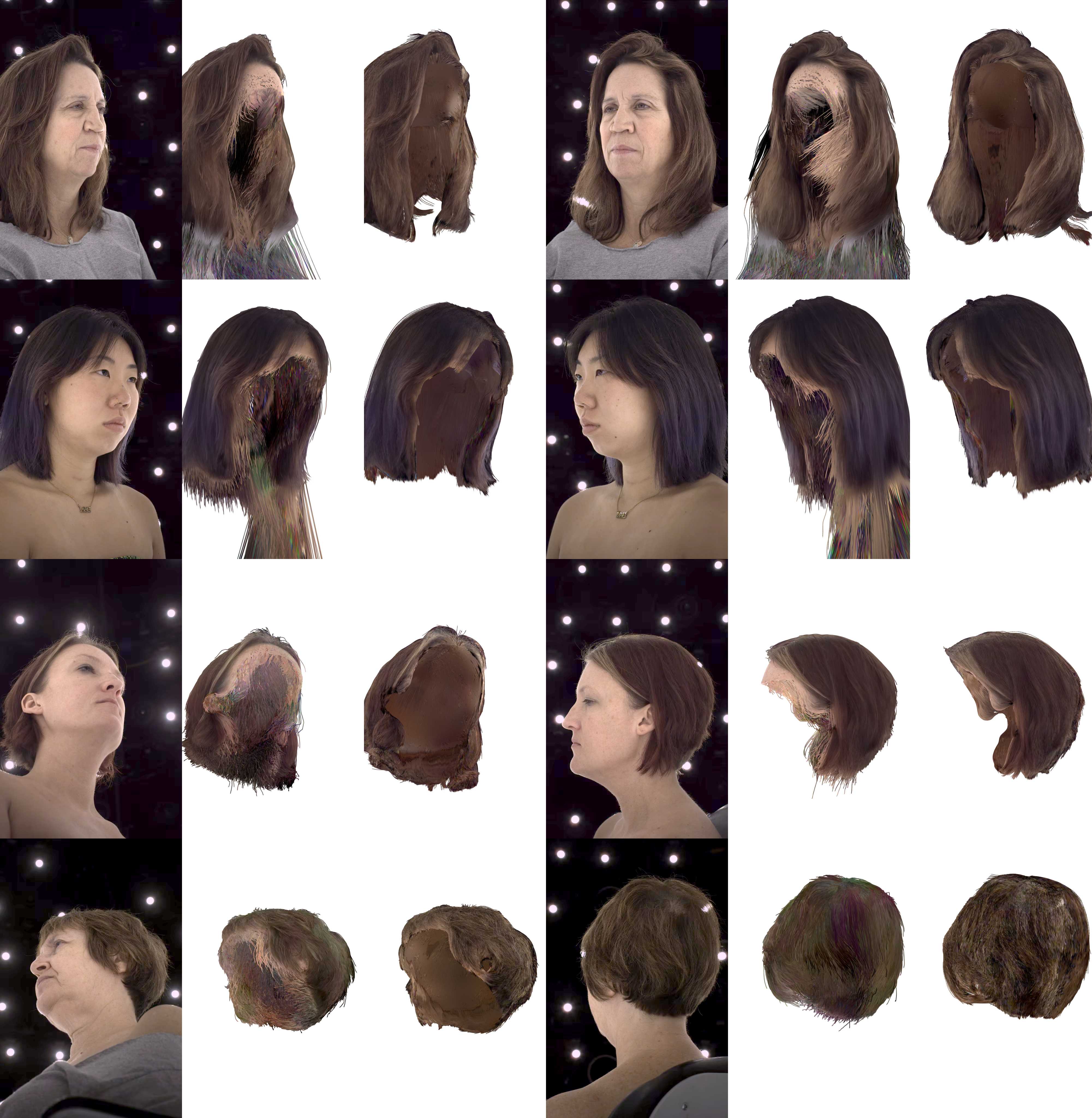}
  \resizebox{\linewidth}{!}{
    \begin{tabularx}{\linewidth}{YYYYYY}
        Ground Truth & 
        GaussianHaircut~\cite{zakharov2024gh} & 
        Ours & 
        Ground Truth & 
        GaussianHaircut~\cite{zakharov2024gh} & 
        Ours
    \end{tabularx}
    }
  \caption{ \textbf{GaussianHaircut~\cite{zakharov2024gh} comparison.}
 As can be seen, GaussianHaircut is not yet suitable for simulation with learned appearance, as some hair strands penetrate the head and are assigned skin colors. In contrast, our reconstructed hair strands exhibit more consistent coloring due to the proposed consistency loss and can be used directly in simulation. %
  }
  \label{fig:comparisonGH_inner}
\end{figure*}

\begin{table}[b]
    \caption{Comparison to GaussianHaircut ~\cite{zakharov2024gh}. \colorbox{green}{Green} indicates the best and \colorbox{yellow}{yellow} indicates the second.}
\centering
\resizebox{0.8\linewidth}{!}{

    \begin{tabular}{lccc}
          &  PSNR↑& SSIM↑ & LPIPS↓ \\
        \hline
        GH~\cite{zakharov2024gh} & \cellcolor[HTML]{E2EFDA}30.55 & \cellcolor[HTML]{FFF2CC}0.915 & \cellcolor[HTML]{FFF2CC}0.071  \\ 
        Ours w/$\mathcal{L}_{\text{consistency}}$ & 28.24 & 0.893 &  0.078  \\
        Ours &\cellcolor[HTML]{FFF2CC}30.28 & \cellcolor[HTML]{E2EFDA}0.946 & \cellcolor[HTML]{E2EFDA}0.061  \\
        \hline
    \end{tabular}
    }
    \label{tab:comparison_GH}
\end{table}

\begin{figure}[tb]
  \centering
    \begin{minipage}{0.05\linewidth}
        \centering
        \footnotesize
        \rotatebox{90}{~~~~~~~~~~~~~~~~Ours ~~~~~~~~~~~~~~~~~~~~~ HairCUP~\cite{kim2025haircup}}
    \end{minipage}
    \begin{minipage}{0.9\linewidth}
        \includegraphics[width=1.0\linewidth,keepaspectratio=true, trim=0cm 2cm 0cm 0cm, clip]{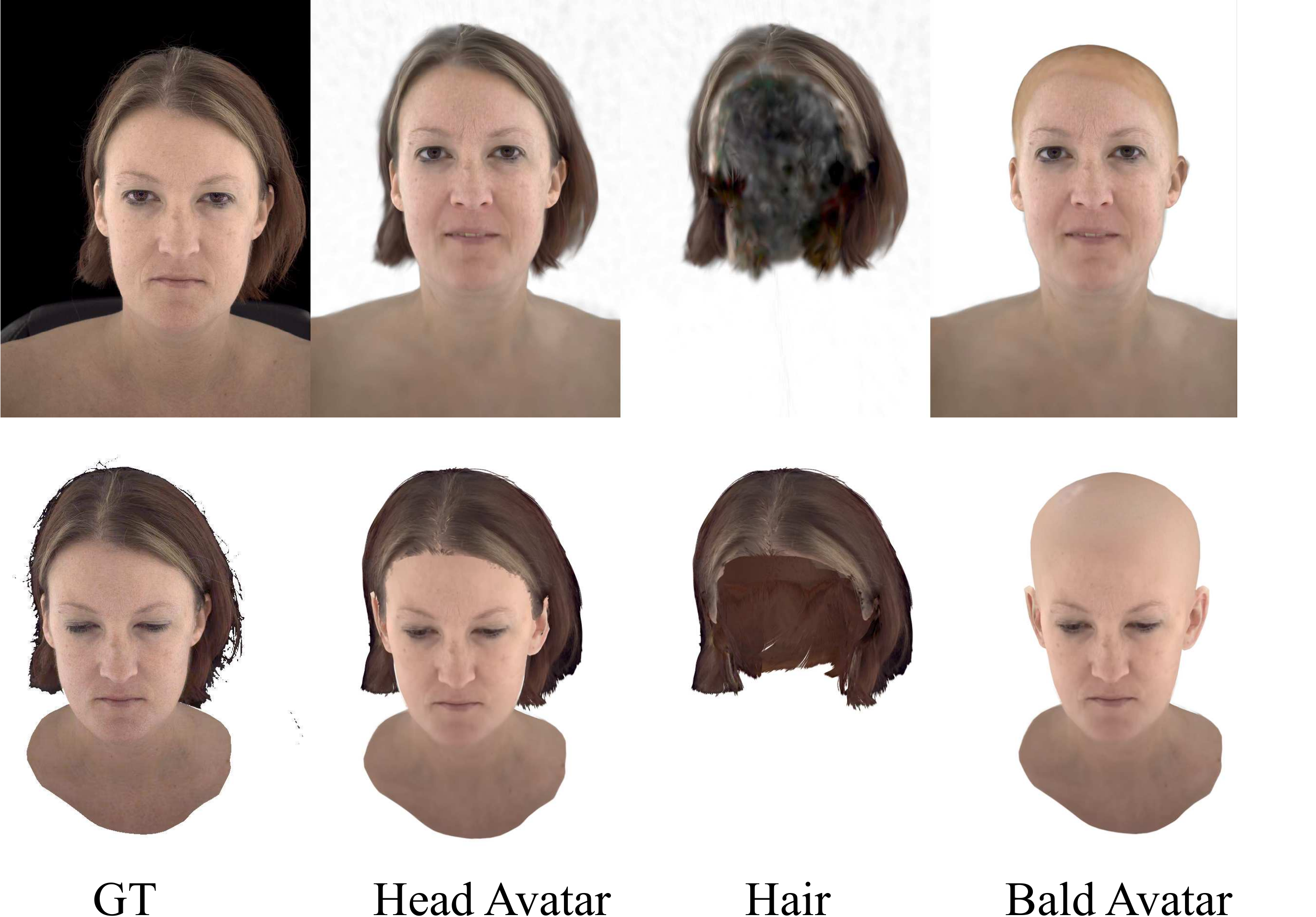}
    
        \footnotesize
        \begin{tabularx}{\linewidth}{YYYY}
            Ground Truth & 
            Composed Avatar & 
            Hair Layer &
            Head Layer        
        \end{tabularx}
    \end{minipage}
  \caption{ \textbf{HairCUP~\cite{kim2025haircup} comparison.} Since haircup uses SDS, it exhibits artifacts and has tendency to generate the same type of skin color. In contrast, we use VLM model for image generation which is trained on a lot of data. This helps with generalization to different skin colors. Despite being compositional, it has unstructured gaussians which does not allow for physics simulation. For more results, see supp mat. 
  \textit{HairCUP images taken from their original paper.} 
  }
  \label{fig:haircup}
\end{figure}

\paragraph{Dataset.} We use the Ava-256 dataset~\cite{martinez2024codec} with 16 cameras per subject, selected via the Hungarian algorithm to match the Nersemble dataset~\cite{Kirschstein2023NeRSembleMR} distribution. Our subset includes 8 expressions with one sequence held out for expression synthesis and one frontal camera for novel view synthesis. See suppl. doc. for actor details.

\paragraph{Baselines.} 

\noindent\textit{Gaussian Avatars (GA)~\cite{Qian2024gaussianavatars}} represents appearance using FLAME-rigged Gaussian primitives, with control driven directly by the FLAME model. It is computationally efficient since no neural network is queried for appearance. However, the hair geometry is modeled through 3DMM offsets, resulting in rigid and non-dynamic hair.

\noindent\textit{Gaussian Head Avatars (GHA)~\cite{xu2023gaussianheadavatar}} is a Gaussian-based head avatar approach comprising two stages. First, a mesh head is optimized at each time step from multiview videos. Then, a deformation MLP learns temporal deformations, followed by a super-resolution module for refinement. GHA uses BFM~\cite{bfm09} for expression control and produces rigid hair, as hair geometry is not explicitly modeled.

\noindent\textit{Gaussian Haircut (GH)~\cite{zakharov2024gh}} optimizes hair appearance by attaching Gaussian primitives to the midpoints of hair segments and using photometric losses combined with orientation-based regularization. Our method builds on Neural Haircut by leveraging strand-based hair geometry and optimizing appearance to achieve realistic hair rendering during physical simulations.

\noindent\textit{HairCUP~\cite{kim2025haircup}}, a concurrent universal compositional avatar method built on UrAvatar~\cite{li2024uravatar}, also models the head and hair separately, enabling hair transfer in addition to expression control. However, bald avatars synthesized by HairCUP exhibit artifacts in unseen regions. Moreover, although the representation is decompositional, the hair is modeled as unstructured Gaussians, leading to rigid and non-dynamic behavior.

\medskip

\subsection{Comparison}
\paragraph{Hair reconstruction evaluation.}
We evaluate the quality of our obtained hair appearance model in comparison to GaussianHaircut.
In \Cref{fig:comparisonGH_inner}, we show novel views of the hair region, which is represented with hair strands and attached 3D Gaussian primitives.
As can be seen, GaussianHaircut has hair strands inside the head region that learn skin color, which restricts it from being animated through simulation with learned colors.
The overall static appearance of the GaussianHaircut reconstruction and ours is similar, see \Cref{tab:comparison_GH}.

\paragraph{Cross-reenactment evaluation.}
In \Cref{fig:maincomparison} we show a comparison to GHA~\cite{xu2023gaussianheadavatar} and GA~\cite{Qian2024gaussianavatars}. When there is not much hair motion, the baseline methods show similar quality as our method (t=0). However, the difference gets clear in a motion sequence that results in dynamically moving hair. As our method, Physhead completes missing parts during capture; it generalizes to very diverse motions such as rotating head, nodding, effects like wind while baseline methods remains with rigid hair.

\paragraph{Comparison of layered avatar.}
As shown in \Cref{fig:haircup}, the concurrent work HairCUP is also compositional. Despite being realistic when head and hair layer are combined, the head layer has artifacts due to the underlying SDS-based procedure. In contrast, our method utilizes an appearance proxy generated with a vision language model and differentiable rendering which is then blended with the original image content to faithfully preserve its details and skin color.

\begin{figure}[tb]
  \centering
    \includegraphics[width=1.0\linewidth,keepaspectratio=true]{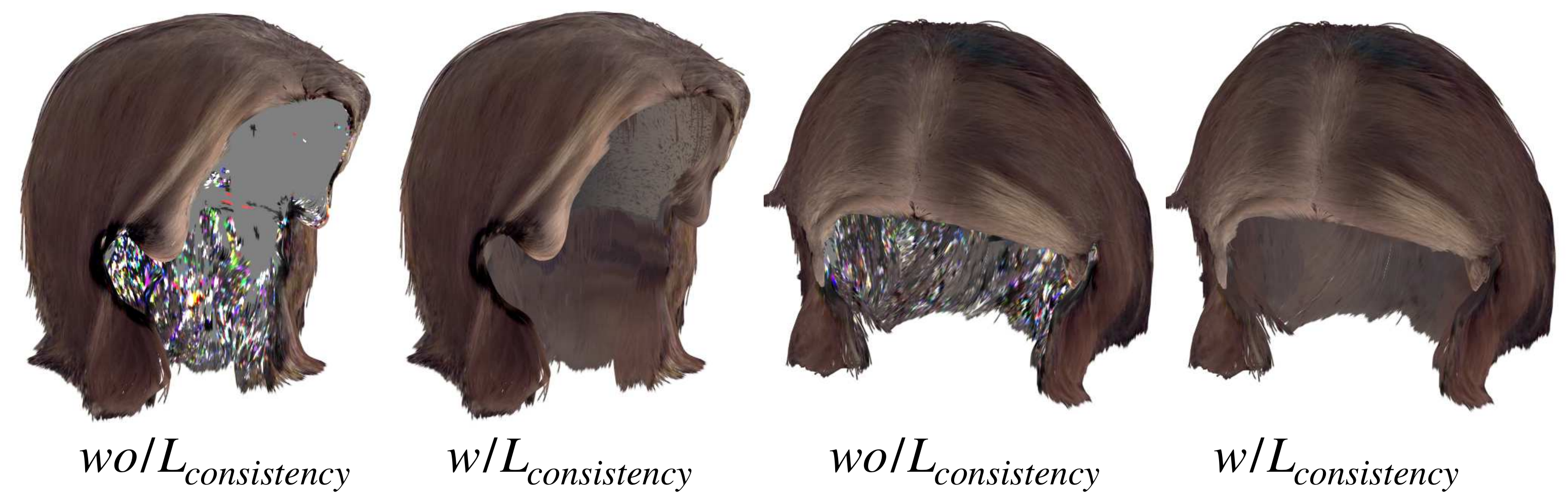}
  \caption{ The consistency loss helps to assign meaningful colors to hidden 3D gaussian primitives.}
  \label{fig:ablationconsistency}
\end{figure}

\subsection{Ablation studies}
In the supplemental document, we show ablation studies with respect to our proposed layered representation.
It is key for being able to simulate the hair.
A single-layer approach would lead to artifacts such as skin peeling.
In addition, hidden regions of skin and hair would not be handled well.
Besides the layered representation, we have introduced a color consistency regularization.
Without this regularization, the inner hair strands will have random colors, restricting it from being animated.
We enable this regularization for the hair strands after 5k iterations. The intuition behind this is that the visible strands are initially optimized based on the RGB images, allowing their colors to propagate naturally to the hidden strands, see \Cref{fig:ablationconsistency}.

\subsection{Applications}
\begin{figure}[tb]
    \centering
    \includegraphics[width=1.0\linewidth,keepaspectratio=true]{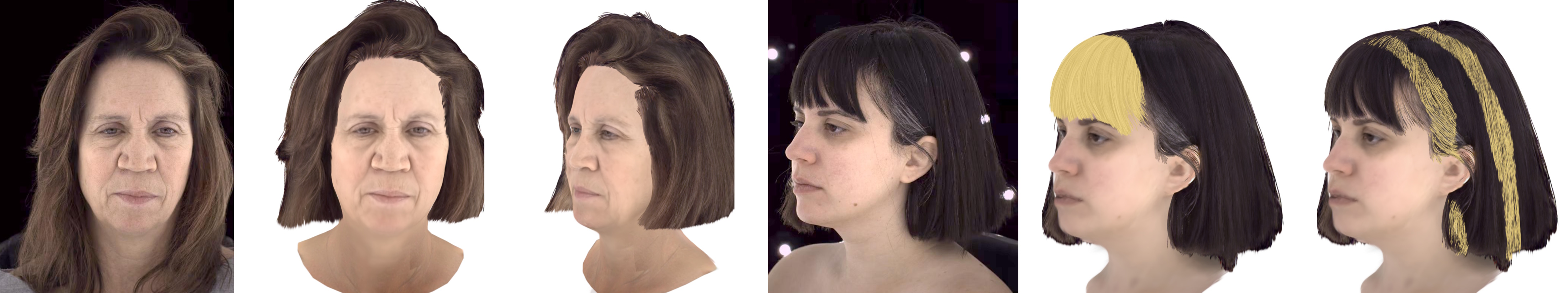}
    \footnotesize
        \begin{tabularx}{\linewidth}{YY}
            Hair Geometry Editing &
            Hair Appearance Editing
        \end{tabularx}
    \caption{The strand-based hair layer can be edited by the user to change the geometry and appearance.}
    \label{fig:editing}
\end{figure}

\begin{figure}[tb]
  \centering
      \begin{minipage}{0.05\linewidth}
        \centering
        \footnotesize
        \rotatebox{90}{~Short $\rightarrow$ Long~~~~~~ Long $\rightarrow$ Short}
    \end{minipage}
        \begin{minipage}{0.9\linewidth}
        \includegraphics[width=1.0\linewidth,keepaspectratio=true]{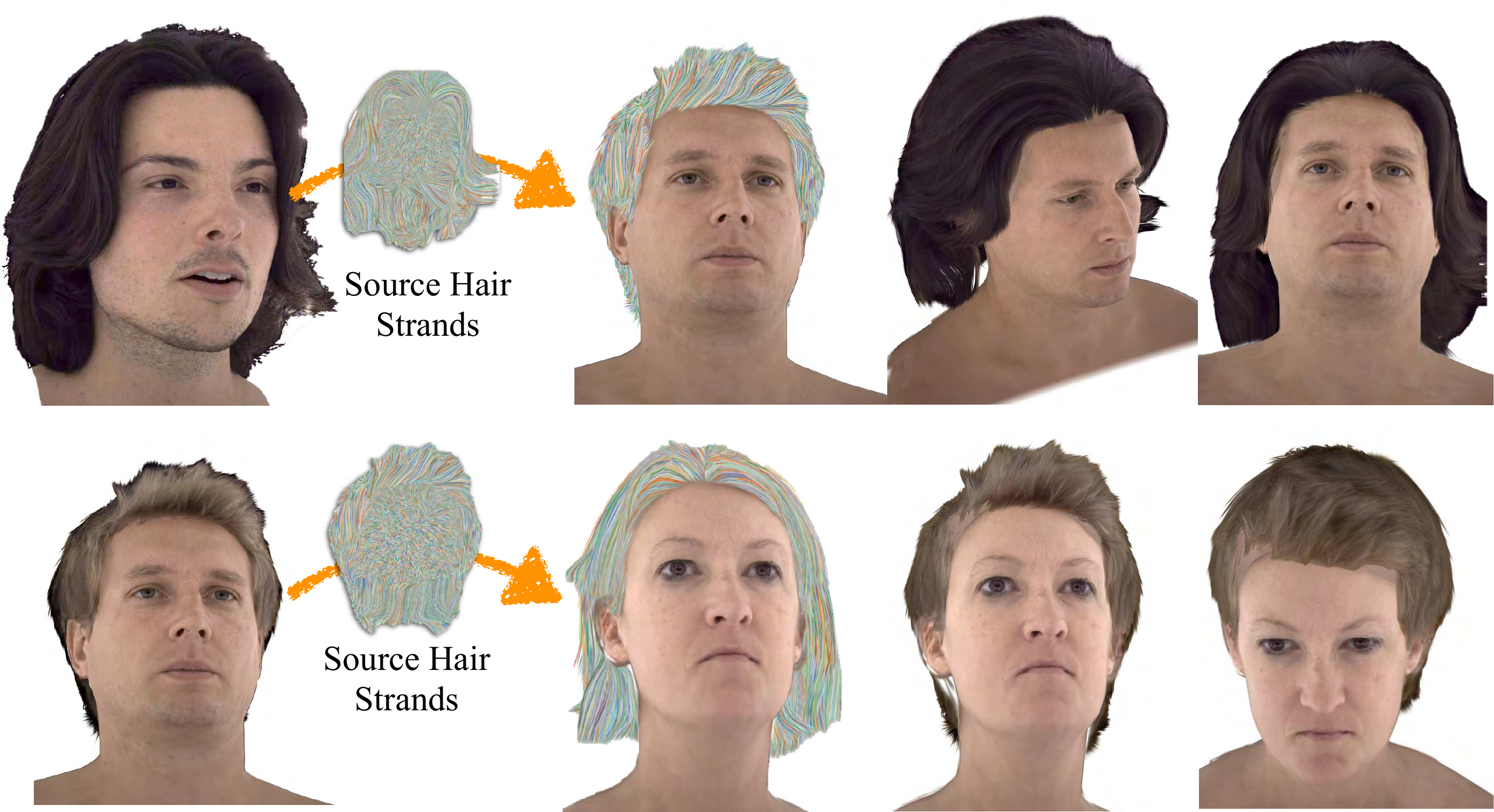}
        \footnotesize
        \begin{tabularx}{\linewidth}{YYY}
            Source Actor &
            Target Actor &
            Transferred Hair
        \end{tabularx}
    \end{minipage}
  \caption{The layered avatar representation allows for editing operations like hair-swapping.}
  \label{fig:hairstransfer}
\end{figure}

Our method uses dense explicit strands, which consist of connected 3D points. Thus, it can be used in several applications, such as hairstyle transfer \Cref{fig:hairstransfer}, hair geometry and appearance editing \Cref{fig:editing}.

\subsection{Limitations}

While \OURS is able to produce physically plausible hair dynamics for animatable 3D head avatars, the appearance quality is dependent on the quality of the foreground and hair masks.
Imperfect masks might result in wrong disentanglement between hair and head.
In addition, \OURS relies on the hair geometry quality of NeuralHaircut~\cite{sklyarova2023neural}, thus, inheriting its limitations (e.g., curly hair).

\section{Conclusion}
\label{sec:conclusion}

In this paper, we have presented the first method that combines physics-driven hair dynamics with an animatable 3D Gaussian-splatting-based head avatar.
It is enabled by a layered representation that disentangles the face region from hair.
The face region is modeled with 3D Gaussian primitives that are attached to a 3DMM which allows for facial expression control.
To handle the hair region and its dynamics that are mainly dependent on the head pose, a strand-based hairstyle reconstruction is executed to provide a representation that is compatible to physics simulation.
On top of this strand-based representation, 3D Gaussian primitives are attached which are optimized to reproduce the appearance of the hair from the input images.
This layered representation allows us to demonstrate physically plausible hair motion, as well as hair editing.
We believe that \OURS is a stepping stone towards highly realistic human head avatars that generalize to novel poses.

\section{Acknowledgments}
\label{sec:result}
We thank Y. Feng, J.~E.~Lenssen, D.~Eskandar, S. Qian, S. Giebenhain, K.~Cetinkaya, N.~Kister, G.~Tiwari, and S.~Bharadwaj for helpful discussions and help during the project.
Berna Kabadayi is supported by the International Max Planck Research School for Intelligent Systems (IMPRS-IS).
Vanessa Sklyarova is supported by the Max Planck ETH Center for Learning Systems. 
Justus Thies is supported by the ERC Starting Grant 101162081 ``LeMo'' and the DFG Excellence Strategy EXC-3057 – Project No. 533677015. 
Gerard Pons-Moll is a member of the Machine Learning Cluster of Excellence, EXC number 2064/1 – Project No. 390727645 and is endowed by the Carl Zeiss Foundation. 
This work was supported by the German Federal Ministry of Education and Research (BMBF): Tübingen AI Center, FKZ: 01IS18039A and funded by the Deutsche Forschungsgemeinschaft (DFG, German Research Foundation) – 409792180 (Emmy Noether Programme, project: Real Virtual Humans).

{
    \small
    \bibliographystyle{ieeenat_fullname}
    \bibliography{main}

@String(CVPR= {IEEE Conf. Comput. Vis. Pattern Recog.})

@String(ICCV= {Int. Conf. Comput. Vis.})

@String(ECCV= {Eur. Conf. Comput. Vis.})

@String(TOG= {ACM Trans. Graph.})

@String(CGF  = {Comput. Graph. Forum})

@String(CVPR  = {CVPR})

@String(ICCV  = {ICCV})

@String(ECCV  = {ECCV})

@String(TOG   = {ACM TOG})

@article{loki,
author = {Lesser, Steve and Stomakhin, Alexey and Daviet, Gilles and Wretborn, Joel and Edholm, John and Lee, Noh-Hoon and Schweickart, Eston and Zhai, Xiao and Flynn, Sean and Moffat, Andrew},
title = {Loki: a unified multiphysics simulation framework for production},
year = {2022},
issue_date = {July 2022},
publisher = {Association for Computing Machinery},
address = {New York, NY, USA},
volume = {41},
number = {4},
issn = {0730-0301},
url = {https://doi.org/10.1145/3528223.3530058},
doi = {10.1145/3528223.3530058},
journal = {ACM Trans. Graph.},
month = jul,
articleno = {50},
numpages = {20}
}

@article{kim2025haircup,
  title={HairCUP: Hair Compositional Universal Prior for 3D Gaussian Avatars},
  author={Kim, Byungjun and Saito, Shunsuke and Nam, Giljoo and Simon, Tomas and Saragih, Jason and Joo, Hanbyul and Li, Junxuan},
  journal={arXiv preprint arXiv:2507.19481},
  year={2025}
}

@misc{zheng2025groomlighthybridinverserendering,
      title={GroomLight: Hybrid Inverse Rendering for Relightable Human Hair Appearance Modeling}, 
      author={Yang Zheng and Menglei Chai and Delio Vicini and Yuxiao Zhou and Yinghao Xu and Leonidas Guibas and Gordon Wetzstein and Thabo Beeler},
      year={2025},
      eprint={2503.10597},
      archivePrefix={arXiv},
      primaryClass={cs.GR},
      url={https://arxiv.org/abs/2503.10597}, 
}

@article{zhou2023groomgen,
author = {Zhou, Yuxiao and Chai, Menglei and Pepe, Alessandro and Gross, Markus and Beeler, Thabo},
title = {GroomGen: A High-Quality Generative Hair Model Using Hierarchical Latent Representations},
year = {2023},
issue_date = {December 2023},
publisher = {Association for Computing Machinery},
address = {New York, NY, USA},
volume = {42},
number = {6},
issn = {0730-0301},
url = {https://doi.org/10.1145/3618309},
doi = {10.1145/3618309},
journal = {ACM Trans. Graph.},
month = dec,
articleno = {270},
numpages = {16}
}

@inproceedings{liao2025hades,
  title={HADES: Human Avatar with Dynamic Explicit Hair Strands},
  author={Liao, Zhanfeng and Tu, Hanzhang and Peng, Cheng and Zhang, Hongwen and Zhou, Boyao and Liu, Yebin},
  booktitle={Proceedings of the IEEE/CVF International Conference on Computer Vision},
  pages={12318--12327},
  year={2025}
}

@article{Kerbl20233DGS,
  title={{3D} Gaussian Splatting for Real-Time Radiance Field Rendering},
  author={Bernhard Kerbl and Georgios Kopanas and Thomas Leimkuehler and George Drettakis},
  journal=TOG,
  year={2023},
  volume={42},
  pages={1 - 14},
}

@inproceedings{Qian2024gaussianavatars,
  title={{GaussianAvatars}: {P}hotorealistic Head Avatars with Rigged {3D} Gaussians},
  author={Qian, Shenhan and Kirschstein, Tobias and Schoneveld, Liam and Davoli, Davide and Giebenhain, Simon and Nie{\ss}ner, Matthias},
  booktitle=CVPR,
  pages        = {20299--20309},
  year={2024}
}

@inproceedings{Zielonka2022InstantVH,
  title={Instant Volumetric Head Avatars},
  author={Wojciech Zielonka and Timo Bolkart and Justus Thies},
  booktitle = {Proceedings of the IEEE/CVF Conference on Computer Vision and Pattern Recognition (CVPR)},
  year={2022},
  pages={4574-4584},
}

@article{Kirschstein2023NeRSembleMR,
  title={{NeRSemble}: {M}ulti-view Radiance Field Reconstruction of Human Heads},
  author={Tobias Kirschstein and Shenhan Qian and Simon Giebenhain and Tim Walter and Matthias Nie{\ss}ner},
  journal=TOG,
  year={2023},
  volume={42},
  pages={1 - 14},
}

@inproceedings{Mildenhall2020NeRF,
  title        = {{NeRF}: {R}epresenting Scenes as Neural Radiance Fields for View Synthesis},
  author       = {Ben Mildenhall and Pratul P. Srinivasan and Matthew Tancik and Jonathan T. Barron and Ravi Ramamoorthi and Ren Ng},
  booktitle    = ECCV,
  volume       = {12346},
  pages        = {405--421},
  year         = {2020},
}

@inproceedings{Blanz1999AMM,
  author    = {Volker Blanz and Thomas Vetter},
  title     = {A Morphable Model for the Synthesis of {3D} Faces},
  booktitle = "SIGGRAPH",
  pages     = {187--194},
  year      = {1999}
}

@inproceedings{Pang2023ASHAG,
  author       = {Haokai Pang and Heming Zhu and Adam Kortylewski and Christian Theobalt and Marc Habermann},
  title        = {{ASH:} Animatable Gaussian Splats for Efficient and Photoreal Human Rendering},
  booktitle    = CVPR,
  pages        = {1165--1175},
  publisher    = {{IEEE}},
  year         = {2024},
}

@inproceedings{saito2024rgca,
  author = {Shunsuke Saito and Gabriel Schwartz and Tomas Simon and Junxuan Li and Giljoo Nam},
  title = {Relightable Gaussian Codec Avatars}, 
  booktitle = CVPR,
  pages        = {130--141},
  year = {2024},
}

@article{Kopanas2021PointBasedNR,
  title={Point‐Based Neural Rendering with Per‐View Optimization},
  author={Georgios Kopanas and Julien Philip and Thomas Leimk{\"u}hler and George Drettakis},
  journal=CGF,
  year={2021},
  volume={40},
}

@article{Wang2019DifferentiableSS,
  title={Differentiable surface splatting for point-based geometry processing},
  author={Yifan Wang and Felice Serena and Shihao Wu and Cengiz {\"O}ztireli and Olga Sorkine-Hornung},
  journal=TOG,
  year={2019},
  volume={38},
  pages={1 - 14},
}

@article{Zwicker2001SurfaceS,
  title={Surface splatting},
  author={Matthias Zwicker and Hans R{\"u}diger Pfister and Jeroen van Baar and Markus H. Gross},
  journal="SIGGRAPH",
  pages        = {371--378},
  year={2001},
}

@article{Ramamoorthi2001AnER,
  title={An efficient representation for irradiance environment maps},
  author={Ravi Ramamoorthi and Pat Hanrahan},
  journal="SIGGRAPH",
  year={2001},
}

@article{Goral1984ModelingTI,
  title={Modeling the interaction of light between diffuse surfaces},
  author={Cindy M. Goral and Kenneth E. Torrance and Donald P. Greenberg and Bennett Battaile},
  journal="SIGGRAPH",
  year={1984},
}

@article{tewari2020neuralrendering,
	author = {A. Tewari and O. Fried and J. Thies and V. Sitzmann and S. Lombardi and K. Sunkavalli and R. Martin-Brualla and T. Simon and J. Saragih and M. Nie{\ss}ner and R. Pandey and S. Fanello and G. Wetzstein and J.-Y. Zhu and C. Theobalt and M. Agrawala and E. Shechtman and D. B. Goldman and M. Zollh{\"o}fer},
	title = {State of the Art on Neural Rendering}, 
	journal = {EG},
	year = {2020}
}

@article{tewari2022advances,
      title={Advances in Neural Rendering}, 
      author={Ayush Tewari and Justus Thies and Ben Mildenhall and Pratul Srinivasan and Edgar Tretschk and Yifan Wang and Christoph Lassner and Vincent Sitzmann and Ricardo Martin-Brualla and Stephen Lombardi and Tomas Simon and Christian Theobalt and Matthias Niessner and Jonathan T. Barron and Gordon Wetzstein and Michael Zollhoefer and Vladislav Golyanik},
      year={2022},
      journal      = CGF,
      pages        = {703--735},
}

@article{egger2020morphablemodels,
    author = {Egger, Bernhard and Smith, William A. P. and Tewari, Ayush and Wuhrer, Stefanie and Zollhoefer, Michael and Beeler, Thabo and Bernard, Florian and Bolkart, Timo and Kortylewski, Adam and Romdhani, Sami and Theobalt, Christian and Blanz, Volker and Vetter, Thomas},
    title = {{3D} Morphable Face Models—Past, Present, and Future},
    year = {2020},
    volume = {39},
    number = {5},
    issn = {0730-0301},
    journal = TOG,
}

@article{zollhoefer2018facestar,
  author = {Michael Zollh{\"o}fer and Justus Thies and Darek Bradley and Pablo Garrido and Thabo Beeler and Patrick P{\'e}erez and Marc Stamminger and Matthias Nie{\ss}ner and Christian Theobalt},
  title = {State of the Art on Monocular {3D} Face Reconstruction, Tracking, and Applications}, 
  journal      = CGF,
  volume       = {37},
  number       = {2},
  pages        = {523--550},
  year         = {2018},
}

@article{Kingma2014AdamAM,
  title={Adam: A Method for Stochastic Optimization},
  author={Diederik P. Kingma and Jimmy Ba},
  journal={CoRR},
  year={2014},
  volume={abs/1412.6980},
}

@article{Thies2016Face2FaceRF,
  title={{Face2Face}: {R}eal-Time Face Capture and Reenactment of {RGB} Videos},
  author={Justus Thies and Michael Zollh{\"o}fer and Marc Stamminger and Christian Theobalt and Matthias Nie{\ss}ner},
  journal=CVPR,
  year={2016},
  pages={2387-2395},
}

@article{FLAME:SiggraphAsia2017, 
  title = {Learning a model of facial shape and expression from {4D} scans}, 
  author = {Li, Tianye and Bolkart, Timo and Black, Michael. J. and Li, Hao and Romero, Javier}, 
  journal = SIGGRAPHASIA, 
  volume = {36}, 
  number = {6}, 
  year = {2017}, 
  pages = {194:1--194:17},
}

@article{Gafni2020DynamicNR,
  title={Dynamic Neural Radiance Fields for Monocular {4D} Facial Avatar Reconstruction},
  author={Guy Gafni and Justus Thies and Michael Zollhofer and Matthias Nie{\ss}ner},
  journal=CVPR,
  year={2020},
  pages={8645-8654},
}

@article{An2023PanoHeadG3,
  title={{PanoHead}: {G}eometry-Aware {3D} Full-Head Synthesis in 360°},
  author={Sizhe An and Hongyi Xu and Yichun Shi and Guoxian Song and Umit Y. Ogras and Linjie Luo},
  journal=CVPR,
  year={2023},
  pages={20950-20959},
}

@article{Lombardi2021MixtureOV,
  title={Mixture of volumetric primitives for efficient neural rendering},
  author={Stephen Lombardi and Tomas Simon and Gabriel Schwartz and Michael Zollhoefer and Yaser Sheikh and Jason M. Saragih},
  journal=TOG,
  year={2021},
  volume={40},
  pages={1 - 13},
}

@misc{feng2024gaussian,
      title={Gaussian Splashing: {D}ynamic Fluid Synthesis with Gaussian Splatting}, 
      author={Yutao Feng and Xiang Feng and Yintong Shang and Ying Jiang and Chang Yu and Zeshun Zong and Tianjia Shao and Hongzhi Wu and Kun Zhou and Chenfanfu Jiang and Yin Yang},
      year={2024},
      eprint={2401.15318},
      archivePrefix={arXiv},
      primaryClass={cs.GR}
}

@article{bfm09,
    title={A {3D} Face Model for Pose and Illumination Invariant Face Recognition},
    author={P. Paysan and R. Knothe and B. Amberg
            and S. Romdhani and T. Vetter},
    journal={IEEE International Conference on Advanced Video and Signal based Surveillance (AVSS)},
    year={2009},
    }

@inproceedings{SMPL-X:2019,
  title = {Expressive Body Capture: {3D} Hands, Face, and Body from a Single Image},
  author = {Pavlakos, Georgios and Choutas, Vasileios and Ghorbani, Nima and Bolkart, Timo and Osman, Ahmed A. A. and Tzionas, Dimitrios and Black, Michael J.},
  booktitle = CVPR,
  pages     = {10975--10985},
  year = {2019}
}

@inproceedings{Giebenhain_2024, series={SA ’24},
   title={NPGA: Neural Parametric Gaussian Avatars},
   url={http://dx.doi.org/10.1145/3680528.3687689},
   DOI={10.1145/3680528.3687689},
   booktitle={SIGGRAPH Asia 2024 Conference Papers},
   publisher={ACM},
   author={Giebenhain, Simon and Kirschstein, Tobias and Rünz, Martin and Agapito, Lourdes and Nießner, Matthias},
   year={2024},
   month=dec, pages={1–11},
   collection={SA ’24} }

@misc{kirschstein2024ggheadfastgeneralizable3d,
      title={GGHead: Fast and Generalizable 3D Gaussian Heads}, 
      author={Tobias Kirschstein and Simon Giebenhain and Jiapeng Tang and Markos Georgopoulos and Matthias Nießner},
      year={2024},
      eprint={2406.09377},
      archivePrefix={arXiv},
      primaryClass={cs.CV},
}

@article{navaneet2023compact3d,
  title={CompGS: Smaller and Faster Gaussian Splatting with Vector Quantization},
  author={Navaneet, KL and Meibodi, Kossar Pourahmadi and Koohpayegani, Soroush Abbasi and Pirsiavash, Hamed},
  journal={ECCV},
  year={2024},
}

@misc{fan2023lightgaussian,
  author    = {Zhiwen Fan and Kevin Wang and Kairun Wen and Zehao Zhu and Dejia Xu and Zhangyang Wang},
  title     = {LightGaussian: Unbounded 3D Gaussian Compression with 15x Reduction and 200+ FPS},
  year      ={2023}, 
  eprint    ={2311.17245}, 
  archivePrefix={arXiv}, 
  primaryClass={cs.CV},
}

@inproceedings{zielonka2025synshot,
    title={Synthetic Prior for Few-Shot Drivable Head Avatar Inversion},
    author={Wojciech Zielonka and Stephan J. Garbin and Alexandros Lattas 
                    and George Kopanas and Paulo Gotardo and Thabo Beeler 
                    and Justus Thies and Timo Bolkart},
    booktitle = {Proceedings of the IEEE/CVF Conference on Computer Vision and Pattern Recognition (CVPR)},
    month = {June},
    year={2025},
}

@inproceedings{nehvi_3vp,
author = {Nehvi, Jalees and Kabadayi, Berna and Valentin, Julien and Thies, Justus},
title = { Volumetric Portrait Avatar},
year = {2025},
isbn = {978-3-031-85186-5},
url = {https://doi.org/10.1007/978-3-031-85187-2_1},
doi = {10.1007/978-3-031-85187-2_1},
booktitle = {Pattern Recognition: 46th DAGM German Conference, DAGM GCPR 2024, Munich, Germany, September 10–13, 2024, Proceedings, Part II},
pages = {3–19},
numpages = {17},
}

@inproceedings{zielonka2024gem,
    title={Gaussian Eigen Models for Human Heads},
    author={Wojciech Zielonka and Timo Bolkart and Thabo Beeler and Justus Thies},
    booktitle = {Proceedings of the IEEE/CVF Conference on Computer Vision and Pattern Recognition (CVPR)},
    month = {June},
    year={2025},
}

@article{Feng2023DELTA,
    author = {Feng, Yao and Liu, Weiyang and Bolkart, Timo and Yang, Jinlong and Pollefeys, Marc and Black, Michael J.},
    title = {Learning Disentangled Avatars with Hybrid 3D Representations},
    journal={arXiv},
    year = {2023}
}

@inproceedings{sklyarova2023haar,
  title = {Text-Conditioned Generative Model of 3D Strand-based Human Hairstyles},
  booktitle = {IEEE/CVF Conf.~on Computer Vision and Pattern Recognition (CVPR)},
  month = jun,
  year = {2024},
  author = {Sklyarova, Vanessa and Zakharov, Egor and Hilliges, Otmar and Black, Michael J. and Thies, Justus},
  url = {https://haar.is.tue.mpg.de/},
  month_numeric = {6}
}

@article{zakharov2024gh,
    title={Human Hair Reconstruction with Strand-Aligned 3D Gaussians},
    author={Zakharov, Egor and Sklyarova, Vanessa and Black, Michael J and Nam, Giljoo and Thies, Justus and Hilliges, Otmar},
    journal={ArXiv},
    month={Sep}, 
    year={2024} 
}

@article {shellnerf,
journal = {Computer Graphics Forum},
title = {{ShellNeRF: Learning a Controllable High-resolution Model of the Eye and Periocular Region}},
author = {Li, Gengyan and Sarkar, Kripasindhu and Meka, Abhimitra and Buehler, Marcel and Mueller, Franziska and Gotardo, Paulo and Hilliges, Otmar and Beeler, Thabo},
year = {2024},
publisher = {The Eurographics Association and John Wiley & Sons Ltd.},
ISSN = {1467-8659},
DOI = {10.1111/cgf.15041}
}

@proceedings{sarkar2023litnerf,          
  author = {Kripasindhu Sarkar and Marcel C. Buehler and Gengyan Li and Daoye Wang and  Delio Vicini and  Jérémy Riviere and Yinda Zhang
                    and Sergio Orts-Escolano and Paulo Gotardo and Thabo Beeler and Abhimitra Meka},
  title = {LitNeRF: Intrinsic Radiance Decomposition for High-Quality View Synthesis and Relighting of Faces},
  booktitle = {ACM SIGGRAPH Asia 2023 Conference Papers, December 12--15, 2023, Sydney, NSW, Australia},
  url = {https://doi.org/10.1145/3550469},
  doi = {10.1145/3610548.3618210},
  isbn = {979-8-4007-0315-7/23/12},
  year={2023}}

@inproceedings{morf,
author = {Wang, Daoye and Chandran, Prashanth and Zoss, Gaspard and Bradley, Derek and Gotardo, Paulo},
title = {MoRF: Morphable Radiance Fields for Multiview Neural Head Modeling},
year = {2022},
isbn = {9781450393379},
publisher = {Association for Computing Machinery},
address = {New York, NY, USA},
url = {https://doi.org/10.1145/3528233.3530753},
doi = {10.1145/3528233.3530753},
articleno = {55},
numpages = {9},
location = {Vancouver, BC, Canada},
series = {SIGGRAPH '22}
}

@article{Gao2022nerfblendshape,
 author = {Xuan Gao and Chenglai Zhong and Jun Xiang and Yang Hong and Yudong Guo and Juyong Zhang}, 
 title = {Reconstructing Personalized Semantic Facial NeRF Models From Monocular Video}, 
 journal = {ACM Transactions on Graphics (Proceedings of SIGGRAPH Asia)}, 
 volume = {41}, 
 number = {6}, 
 year = {2022}, 
 doi = {10.1145/3550454.3555501} }

@article{thies2015realtime,
	author = {Thies, J. and Zollh{\"o}fer, M. and Nie{\ss}ner, M. and Valgaerts, L. and Stamminger, M. and Theobalt, C.},
	title = {Real-time Expression Transfer for Facial Reenactment},
	journal = {ACM Transactions on Graphics (TOG)},
	volume={34},
	number={6},
	year = {2015},
	publisher = {ACM}
}

@article{Garbin2024VolTeMorph,
  author    = {Stephan J. Garbin and Marek Kowalski and Virginia Estellers and Stanislaw Szymanowicz and Shideh Rezaeifar and Jingjing Shen and Matthew Johnson and Julien Valentin},
  title     = {{VolTeMorph}: Real-time, Controllable and Generalizable Animation of Volumetric Representations},
  journal   = {Computer Graphics Forum},
  year      = {2024},
  note      = {To appear},
  url       = {https://arxiv.org/abs/2208.00949}
}

@inproceedings{Athar2022RigNeRF,
  author    = {ShahRukh Athar and Zexiang Xu and Kalyan Sunkavalli and Eli Shechtman and Zhixin Shu},
  title     = {{RigNeRF}: Fully Controllable Neural 3D Portraits},
  booktitle = {Proceedings of the IEEE/CVF Conference on Computer Vision and Pattern Recognition (CVPR)},
  year      = {2022},
  pages     = {20364--20373},
  url       = {https://openaccess.thecvf.com/content/CVPR2022/papers/Athar_RigNeRF_Fully_Controllable_Neural_3D_Portraits_CVPR_2022_paper.pdf}
}

@inproceedings{Athar2023FLAMEinNeRF,
  author    = {ShahRukh Athar and Zhixin Shu and Dimitris Samaras},
  title     = {{FLAME-in-NeRF}: Neural Control of Radiance Fields for Free View Face Animation},
  booktitle = {IEEE International Conference on Automatic Face and Gesture Recognition (FG)},
  year      = {2023},
  url       = {https://arxiv.org/abs/2108.04913}
}

@inproceedings{kania2022conerf,
  title     = {{CoNeRF: Controllable Neural Radiance Fields}},
  author    = {Kania, Kacper and Yi, Kwang Moo and Kowalski, Marek and Trzci{\'n}ski, Tomasz and Tagliasacchi, Andrea},
  booktitle = {Proceedings of the IEEE Conference on Computer Vision and Pattern Recognition (CVPR)},
  year      = {2022},
  pages     = {20314--20323},
  url       = {https://arxiv.org/abs/2112.01983}
}

@inproceedings{kania2023blendfields,
  title     = {{BlendFields: Few-Shot Example-Driven Facial Modeling}},
  author    = {Kania, Kacper and Garbin, Stephan J. and Tagliasacchi, Andrea and Estellers, Virginia and Yi, Kwang Moo and Valentin, Julien and Trzci{\'n}ski, Tomasz and Kowalski, Marek},
  booktitle = {Proceedings of the IEEE Conference on Computer Vision and Pattern Recognition},
  year      = {2023},
  url       = {https://arxiv.org/abs/2305.07514}
}

@article{lyu2020real,
  title={Real-time hair simulation with neural interpolation},
  author={Lyu, Qing and Chai, Menglei and Chen, Xiang and Zhou, Kun},
  journal={IEEE Transactions on Visualization and Computer Graphics},
  volume={28},
  number={4},
  pages={1894--1905},
  year={2020},
  publisher={IEEE}
}

@article{martinez2024codec,
  author = {Julieta Martinez and Emily Kim and Javier Romero and Timur Bagautdinov and Shunsuke Saito and Shoou-I Yu and Stuart Anderson and Michael Zollhöfer and Te-Li Wang and Shaojie Bai and Chenghui Li and Shih-En Wei and Rohan Joshi and Wyatt Borsos and Tomas Simon and Jason Saragih and Paul Theodosis and Alexander Greene and Anjani Josyula and Silvio Mano Maeta and Andrew I. Jewett and Simon Venshtain and Christopher Heilman and Yueh-Tung Chen and Sidi Fu and Mohamed Ezzeldin A. Elshaer and Tingfang Du and Longhua Wu and Shen-Chi Chen and Kai Kang and Michael Wu and Youssef Emad and Steven Longay and Ashley Brewer and Hitesh Shah and James Booth and Taylor Koska and Kayla Haidle and Matt Andromalos and Joanna Hsu and Thomas Dauer and Peter Selednik and Tim Godisart and Scott Ardisson and Matthew Cipperly and Ben Humberston and Lon Farr and Bob Hansen and Peihong Guo and Dave Braun and Steven Krenn and He Wen and Lucas Evans and Natalia Fadeeva and Matthew Stewart and Gabriel Schwartz and Divam Gupta and Gyeongsik Moon and Kaiwen Guo and Yuan Dong and Yichen Xu and Takaaki Shiratori and Fabian Prada and Bernardo R. Pires and Bo Peng and Julia Buffalini and Autumn Trimble and Kevyn McPhail and Melissa Schoeller and Yaser Sheikh},
  title = {{Codec Avatar Studio: Paired Human Captures for Complete, Driveable, and Generalizable Avatars}},
  year = {2024},
  journal = {NeurIPS Track on Datasets and Benchmarks},
}

@inproceedings{coumans2015bullet,
  author    = {Erwin Coumans},
  title     = {Bullet Physics Simulation},
  booktitle = {ACM SIGGRAPH 2015 Courses},
  year      = {2015},
  pages     = {1},
  doi       = {10.1145/2776880.2792704},
  url       = {https://dl.acm.org/doi/10.1145/2776880.2792704}
}

@inproceedings{sklyarova2023neural,
  title     = {{Neural Haircut: Prior-Guided Strand-Based Hair Reconstruction}},
  author    = {Sklyarova, Vanessa and Chelishev, Jenya and Dogaru, Andreea and Medvedev, Igor and Lempitsky, Victor and Zakharov, Egor},
  booktitle = {Proceedings of the IEEE/CVF International Conference on Computer Vision (ICCV)},
  year      = {2023},
  url       = {https://arxiv.org/abs/2306.05872}
}

@misc{kirschstein2025avat3r,
      title={Avat3r: Large Animatable Gaussian Reconstruction Model for High-fidelity 3D Head Avatars},
      author={Tobias Kirschstein and Javier Romero and Artem Sevastopolsky and Matthias Nie\ss{}ner and Shunsuke Saito},
      year={2025},
      eprint={2502.20220},
      archivePrefix={arXiv},
      primaryClass={cs.CV},
      url={https://arxiv.org/abs/2502.20220},
}

@inproceedings{xiang2024flashavatar,
    author    = {Jun Xiang and Xuan Gao and Yudong Guo and Juyong Zhang},
    title     = {FlashAvatar: High-fidelity Head Avatar with Efficient Gaussian Embedding},
    booktitle = {The IEEE Conference on Computer Vision and Pattern Recognition (CVPR)},
    year      = {2024},
}

@inproceedings{shao2024splattingavatar,
  title = {{SplattingAvatar: Realistic Real-Time Human Avatars with Mesh-Embedded Gaussian Splatting}},
  author = {Shao, Zhijing and Wang, Zhaolong and Li, Zhuang and Wang, Duotun and Lin, Xiangru and Zhang, Yu and Fan, Mingming and Wang, Zeyu},
  booktitle = {Proceedings of the IEEE/CVF Conference on Computer Vision and Pattern Recognition (CVPR)},
  year = {2024}
}

@inproceedings{li2024animatablegaussians,
  title={Animatable Gaussians: Learning Pose-dependent Gaussian Maps for High-fidelity Human Avatar Modeling},
  author={Li, Zhe and Zheng, Zerong and Wang, Lizhen and Liu, Yebin},
  booktitle={Proceedings of the IEEE/CVF Conference on Computer Vision and Pattern Recognition (CVPR)},
  year={2024}
}

@inproceedings{xu2024gphm,
    title={3D Gaussian Parametric Head Model},
    author={Xu, Yuelang and Wang, Lizhen and Zheng, Zerong and Su, Zhaoqi and Liu, Yebin},
    booktitle={Proceedings of the European Conference on Computer Vision (ECCV)},
    year={2024}
}

@article{liao2023hhavatar,
  title={HHAvatar: Gaussian Head Avatar with Dynamic Hairs},
  author={Liao, Zhanfeng and Xu, Yuelang and Li, Zhe and Li, Qijing and Zhou, Boyao and Bai, Ruifeng and Xu, Di and Zhang, Hongwen and Liu, Yebin},
  journal={arXiv e-prints},
  pages={arXiv--2312},
  year={2023}
}

@inproceedings{zielonka2022mica,
  title        = {Towards Metrical Reconstruction of Human Faces},
  author       = {Wojciech Zielonka and Timo Bolkart and Justus Thies},
  booktitle    = {European Conference on Computer Vision},
  year         = {2022}
}

@misc{qian2024versatile,
  title   = "VHAP: Versatile Head Alignment with Adaptive Appearance Priors",
  author  = "Qian, Shenhan",
  year    = "2024",
  month   = "September",
  url     = "https://github.com/ShenhanQian/VHAP"
}

@InProceedings{taubner2024flowface,
  author    = {Taubner, Felix and Raina, Prashant and Tuli, Mathieu and Teh, Eu Wern and Lee, Chul and Huang, Jinmiao},
  title     = {{3D} Face Tracking from {2D} Video through Iterative Dense {UV} to Image Flow},
  booktitle = {Proceedings of the IEEE/CVF Conference on Computer Vision and Pattern Recognition (CVPR)},
  month     = {June},
  year      = {2024},
  pages     = {1227-1237}
}

@misc{taubner2024cap4,
      title={CAP4D: Creating Animatable 4D Portrait Avatars with Morphable Multi-View Diffusion Models}, 
      author={Felix Taubner and Ruihang Zhang and Mathieu Tuli and David B. Lindell},
      year={2024},
      eprint={2412.12093},
}

@inproceedings{li2024uravatar,
      author = {Junxuan Li and Chen Cao and Gabriel Schwartz and Rawal Khirodkar and Christian Richardt and Tomas Simon and Yaser Sheikh and Shunsuke Saito},
      title = {URAvatar: Universal Relightable Gaussian Codec Avatars}, 
      booktitle = {ACM SIGGRAPH 2024 Conference Papers},
      year = {2024},
}

@article{gao2024cat3d,
    title={CAT3D: Create Anything in 3D with Multi-View Diffusion Models},
    author={Ruiqi Gao* and Aleksander Holynski* and Philipp Henzler and Arthur Brussee and Ricardo Martin-Brualla and Pratul P. Srinivasan and Jonathan T. Barron and Ben Poole*
    },
    journal={Advances in Neural Information Processing Systems},
    year={2024}
}

@inproceedings{wang2024mega,
  title={MeGA: Hybrid Mesh-Gaussian Head Avatar for High-Fidelity Rendering and Head Editing},
  author={Wang, Cong and Kang, Di and Sun, He-Yi and Qian, Shen-Han and Wang, Zi-Xuan and Bao, Linchao and Zhang, Song-Hai},
  booktitle = {Proceedings of the IEEE/CVF Conference on Computer Vision and Pattern Recognition (CVPR)},
  month     = {June},
  year      = {2025},
}

@inproceedings{neuralstrands,
    title={Neural Strands: Learning Hair Geometry and 
           Appearance from Multi-View Images},
    author={Radu Alexandru Rosu and Shunsuke Saito and Ziyan Wang and Chenglei Wu and Sven Behnke and Giljoo Nam},
    booktitle={Computer Vision – ECCV 2018: 15th European Conference},
    year={2022}
}

@article{GroomCap,
author = {Zhou, Yuxiao and Chai, Menglei and Wang, Daoye and Winberg, Sebastian and Wood, Erroll and Sarkar, Kripasindhu and Gross, Markus and Beeler, Thabo},
title = {{GroomCap}: High-Fidelity Prior-Free Hair Capture},
year = {2024},
issue_date = {December 2024},
publisher = {Association for Computing Machinery},
address = {New York, NY, USA},
volume = {43},
number = {6},
issn = {0730-0301},
url = {https://doi.org/10.1145/3687768},
doi = {10.1145/3687768},
journal = {ACM Trans. Graph.},
month = nov,
articleno = {254},
numpages = {15}
}

@inproceedings{deepmvshair,
author = {Kuang, Zhiyi and Chen, Yiyang and Fu, Hongbo and Zhou, Kun and Zheng, Youyi},
title = {{DeepMVSHair}: Deep Hair Modeling from Sparse Views},
year = {2022},
isbn = {9781450394703},
publisher = {Association for Computing Machinery},
address = {New York, NY, USA},
url = {https://doi.org/10.1145/3550469.3555385},
doi = {10.1145/3550469.3555385},
booktitle = {SIGGRAPH Asia 2022 Conference Papers},
articleno = {10},
numpages = {8},
location = {Daegu, Republic of Korea},
series = {SA '22}
}

@inproceedings{monohair,
    author = {Keyu Wu and Lingchen Yang and Zhiyi Kuang and Yao Feng and Xutao Han and Yuefan Shen and Hongbo Fu and Kun Zhou and Youyi Zheng},
    title = {MonoHair: High-Fidelity Hair Modeling from a Monocular Video},
    booktitle = {2023 IEEE/CVF Conference on Computer Vision and Pattern Recognition (CVPR)},
    year = {2024}
}

@article{luo2024gaussianhair,
  title={{GaussianHair}: Hair Modeling and Rendering with Light-aware Gaussians},
  author={Luo, Haimin and Ouyang, Min and Zhao, Zijun and Jiang, Suyi and Zhang, Longwen and Zhang, Qixuan and Yang, Wei and Xu, Lan and Yu, Jingyi},
  journal={arXiv preprint arXiv:2402.10483},
  year={2024}
}

@InProceedings{drhair,
  author    = {Takimoto, Yusuke and Takehara, Hikari and Sato, Hiroyuki and Zhu, Zihao and Zheng, Bo},
  title     = {Dr.Hair: Reconstructing Scalp-Connected Hair Strands without Pre-Training via Differentiable Rendering of Line Segments},
  booktitle = {Proceedings of the IEEE/CVF Conference on Computer Vision and Pattern Recognition (CVPR)},
  month     = {June},
  year      = {2024},
  pages     = {20601-20611}
}

@article{Daviet2023InteractiveHS,
  title={Interactive Hair Simulation on the GPU using ADMM},
  author={Gilles Daviet},
  journal={ACM SIGGRAPH 2023 Conference Proceedings},
  year={2023},
  url={https://api.semanticscholar.org/CorpusID:259767889}
}

@software{unrealengine,
  author = {{Epic Games}},
  title = {Unreal Engine},
  url = {https://www.unrealengine.com},
  version = {4.22.1},
  date = {2019-04-25},
}

@article{digitalsalon,
    title={Digital Salon: An AI and Physics-Driven Tool for 3D Hair Grooming and Simulation},
    author={He, Chengan and Herrera, Jorge Alejandro Amador and Zhou, Yi and Shu, Zhixin and Sun, Xin and Feng, Yao and Pirk, S{\"o}ren and Michels, Dominik L and Zhang, Meng and Wang, Tuanfeng Y and Rushmeier, Holly},
    journal={ACM SIGGRAPH Asia 2024 Real-Time Live!},
    url={https://digital-salon.github.io/},
    year={2024}
}

@inproceedings{stuyck2024quaffure,
  title={Quaffure: Real-Time Quasi-Static Neural Hair Simulation},
  author={Stuyck, Tuur and Lin, Gene Wei-Chin and Larionov, Egor and Chen, Hsiao-yu and Bozic, Aljaz and Sarafianos, Nikolaos and Roble, Doug},
  booktitle={Proceedings of the IEEE/CVF Conference on Computer Vision and Pattern Recognition},
  year={2025}
}

@InProceedings{HVH,
    author    = {Wang, Ziyan and Nam, Giljoo and Stuyck, Tuur and Lombardi, Stephen and Zollh\"ofer, Michael and Hodgins, Jessica and Lassner, Christoph},
    title     = {HVH: Learning a Hybrid Neural Volumetric Representation for Dynamic Hair Performance Capture},
    booktitle = {Proceedings of the IEEE/CVF Conference on Computer Vision and Pattern Recognition (CVPR)},
    month     = {June},
    year      = {2022},
    pages     = {6143-6154}
}

@article{neuwigs,
  title={NeuWigs: A Neural Dynamic Model for Volumetric Hair Capture and Animation},
  author={Wang, Ziyan and Nam, Giljoo and Stuyck, Tuur and Lombardi, Stephen and Cao, Chen and Saragih, Jason and Zollhoefer, Michael and Hodgins, Jessica and Lassner, Christoph},
  journal={arXiv preprint arXiv:2212.00613},
  year={2022}
}

@inproceedings{nam2019strand,
  title={Strand-accurate multi-view hair capture},
  author={Nam, Giljoo and Wu, Chenglei and Kim, Min H and Sheikh, Yaser},
  booktitle={Proceedings of the IEEE/CVF Conference on Computer Vision and Pattern Recognition},
  pages={155--164},
  year={2019}
}

@InProceedings{Santesteban2022Snug,
    author    = {Santesteban, Igor and Otaduy, Miguel A. and Casas, Dan},
    title     = {SNUG: Self-Supervised Neural Dynamic Garments},
    booktitle = {Proceedings of the IEEE/CVF Conference on Computer Vision and Pattern Recognition (CVPR)},
    month     = {June},
    year      = {2022},
    pages     = {8140-8150}
}

@InProceedings{Santesteban2021CVPR,
    author    = {Santesteban, Igor and Thuerey, Nils and Otaduy, Miguel A. and Casas, Dan},
    title     = {Self-Supervised Collision Handling via Generative 3D Garment Models for Virtual Try-On},
    booktitle = {Proceedings of the IEEE/CVF Conference on Computer Vision and Pattern Recognition (CVPR)},
    month     = {June},
    year      = {2021},
    pages     = {11763-11773}
}

@inproceedings{caphy_su2023,
  author = {Su, Zhaoqi and Hu, Liangxiao and Lin, Siyou and Zhang, Hongwen and Zhang, Shengping and Thies, Justus and Liu, Yebin},
  booktitle = {IEEE/CVF International Conference on Computer Vision (ICCV)},
  title = {CaPhy: Capturing Physical Properties for Animatable Human Avatars},
  year = {2023},
}

@InProceedings{hairnet,
author = {Zhou, Yi and Hu, Liwen and Xing, Jun and Chen, Weikai and Kung, Han-Wei and Tong, Xin and Li, Hao},
title = {HairNet: Single-View Hair Reconstruction using Convolutional Neural Networks},
booktitle = {Proceedings of the European Conference on Computer Vision (ECCV)},
month = {September},
year = {2018}
}

@InProceedings{hairstep,
    author    = {Zheng, Yujian and Jin, Zirong and Li, Moran and Huang, Haibin and Ma, Chongyang and Cui, Shuguang and Han, Xiaoguang},
    title     = {HairStep: Transfer Synthetic to Real Using Strand and Depth Maps for Single-View 3D Hair Modeling},
    booktitle = {Proceedings of the IEEE/CVF Conference on Computer Vision and Pattern Recognition (CVPR)},
    month     = {June},
    year      = {2023},
    pages     = {12726-12735}
}

@inproceedings{wu2022neuralhdhair,
  title={Neuralhdhair: Automatic high-fidelity hair modeling from a single image using implicit neural representations},
  author={Wu, Keyu and Ye, Yifan and Yang, Lingchen and Fu, Hongbo and Zhou, Kun and Zheng, Youyi},
  booktitle={Proceedings of the IEEE/CVF Conference on Computer Vision and Pattern Recognition},
  pages={1526--1535},
  year={2022}
}

@software{maya,
  author = {{Autodesk, INC.}},
  title = {Maya},
  url = {https:/ autodesk.com/maya},
  version = {2019},
  date = {2019-01-15},
}

@article{chai2016adaptive,
  title={Adaptive skinning for interactive hair-solid simulation},
  author={Chai, Menglei and Zheng, Changxi and Zhou, Kun},
  journal={IEEE transactions on visualization and computer graphics},
  volume={23},
  number={7},
  pages={1725--1738},
  year={2016},
  publisher={IEEE}
}

@article{shen2023CT2Hair,
  title={CT2Hair: High-Fidelity 3D Hair Modeling using Computed Tomography},
  author={Shen, Yuefan and Saito, Shunsuke and Wang, Ziyan and Maury, Olivier and Wu, Chenglei and Hodgins, Jessica and Zheng, Youyi and Nam, Giljoo},
  journal={ACM Transactions on Graphics},
  volume={42},
  number={4},
  articleno={75},
  pages={1--13},
  year={2023},
  publisher={ACM New York, NY, USA}
}

@inproceedings{difflocks2025,
  title = {{DiffLocks}: Generating 3D Hair from a Single Image using Diffusion Models},
  author = {Rosu, Radu Alexandru and Wu, Keyu and Feng, Yao and Zheng, Youyi and Black, Michael J.},
  booktitle = {IEEE/CVF Conf.~on Computer Vision and Pattern Recognition(CVPR)},
  year = {2025}
}

@inproceedings{Sklyarova2025Im2haircut,
  title = {{Im2Haircut:} Single-view Strand-based Hair Reconstruction for Human Avatars},
  booktitle = {Proceedings of the IEEE/CVF International Conference on Computer Vision (ICCV)},
  address = {Honolulu, USA},
  month = oct,
  year = {2025},
  author = {Vanessa, Sklyarova and Egor, Zakharov and Malte, Prinzler and Giorgio, Becherini and Michael, Black and Justus, Thies},
  month_numeric = {10}
}

@inproceedings{he2025perm,
    title={Perm: A Parametric Representation for Multi-Style 3D Hair Modeling},
    author={He, Chengan and Sun, Xin and Shu, Zhixin and Luan, Fujun and Pirk, S\"{o}ren and Herrera, Jorge Alejandro Amador and Michels, Dominik L. and Wang, Tuanfeng Y. and Zhang, Meng and Rushmeier, Holly and Zhou, Yi},
    booktitle={International Conference on Learning Representations},
    year={2025}
}

@InProceedings{Cha_2024_CVPR,
    author    = {Cha, Hyunsoo and Kim, Byungjun and Joo, Hanbyul},
    title     = {PEGASUS: Personalized Generative 3D Avatars with Composable Attributes},
    booktitle = {Proceedings of the IEEE/CVF Conference on Computer Vision and Pattern Recognition (CVPR)},
    month     = {June},
    year      = {2024},
    pages     = {1072-1081}
}

@article{zhang2023teca,
      author    = {Zhang, Hao and Feng, Yao and Kulits, Peter and Wen, Yandong and Thies, Justus and Black, Michael J.},
      title     = {TECA: Text-Guided Generation and Editing of Compositional 3D Avatars},
      journal   = {arXiv},
      year      = {2023}
}

@InProceedings{Cha_2025_CVPR,
    author    = {Cha, Hyunsoo and Lee, Inhee and Joo, Hanbyul},
    title     = {PERSE: Personalized 3D Generative Avatars from A Single Portrait},
    booktitle = {Proceedings of the Computer Vision and Pattern Recognition Conference (CVPR)},
    month     = {June},
    year      = {2025},
    pages     = {15953-15962}
}

@article{he2025head,
    title={3DGH: 3D Head Generation with Composable Hair and Face},
    author={He, Chengan and Li, Junxuan and Kirschstein, Tobias and Sevastopolsky, Artem and Saito, Shunsuke and Tan, Qingyang and Romero, Javier and Cao, Chen and Rushmeier, Holly and Nam, Giljoo},
    journal={ACM Transactions on Graphics},
    volume={44},
    number={4},
    pages={1--12},
    year={2025}
}

@article{simavatar2024,
  author    = {Li Xueting and Yuan, Ye and De Mello, Shalini and Daviet, Gilles and Leaf, Jonathan and Macklin, Miles and Kautz, Jan and Iqbal, Umar},
  title     = {SimAvatar: Simulation-Ready Avatars with Layered Hair and Clothing},
  journal   = {Arxiv},
  year      = {2024},
}

@inproceedings{ostrek2025hairfree,
    title={HairFree: Compositional 2D Head Prior for Text-Driven 360{\textdegree} Bald Texture Synthesis},
    author={Mirela Ostrek and Michael J. Black and Justus Thies},
    booktitle={The Thirty-ninth Annual Conference on Neural Information Processing Systems},
    year={2025},
    url={https://openreview.net/forum?id=ja3p9Dylmh}
}

@inproceedings{perez03poisson,
author = {P\'{e}rez, Patrick and Gangnet, Michel and Blake, Andrew},
title = {Poisson image editing},
year = {2003},
isbn = {1581137095},
publisher = {Association for Computing Machinery},
address = {New York, NY, USA},
url = {https://doi.org/10.1145/1201775.882269},
doi = {10.1145/1201775.882269},
booktitle = {ACM SIGGRAPH 2003 Papers},
pages = {313–318},
numpages = {6},
location = {San Diego, California},
series = {SIGGRAPH '03}
}

@inproceedings{xu2023gaussianheadavatar,
  title={Gaussian Head Avatar: Ultra High-fidelity Head Avatar via Dynamic Gaussians},
  author={Xu, Yuelang and Chen, Benwang and Li, Zhe and Zhang, Hongwen and Wang, Lizhen and Zheng, Zerong and Liu, Yebin},
  booktitle={Proceedings of the IEEE/CVF Conference on Computer Vision and Pattern Recognition (CVPR)},
  year={2024}
}

@article{poole2022dreamfusion,
  author = {Poole, Ben and Jain, Ajay and Barron, Jonathan T. and Mildenhall, Ben},
  title = {DreamFusion: Text-to-3D using 2D Diffusion},
  journal = {arXiv},
  year = {2022},
}

@article{Laine2020diffrast,
  title   = {Modular Primitives for High-Performance Differentiable Rendering},
  author  = {Samuli Laine and Janne Hellsten and Tero Karras and Yeongho Seol and Jaakko Lehtinen and Timo Aila},
  journal = {ACM Transactions on Graphics},
  year    = {2020},
  volume  = {39},
  number  = {6}
}

@InProceedings{hairmapper,
    author    = {Wu, Yiqian and Yang, Yong-Liang and Jin, Xiaogang},
    title     = {HairMapper: Removing Hair From Portraits Using GANs},
    booktitle = {Proceedings of the IEEE/CVF Conference on Computer Vision and Pattern Recognition (CVPR)},
    month     = {June},
    year      = {2022},
    pages     = {4227-4236}
}

@article{wang2008computation,
  title={Computation of rotation minimizing frames},
  author={Wang, Wenping and J{\"u}ttler, Bert and Zheng, Dayue and Liu, Yang},
  journal={ACM Transactions on Graphics (TOG)},
  volume={27},
  number={1},
  pages={1--18},
  year={2008},
  publisher={ACM New York, NY, USA}
}

@misc{rvm,
      title={Robust High-Resolution Video Matting with Temporal Guidance}, 
      author={Shanchuan Lin and Linjie Yang and Imran Saleemi and Soumyadip Sengupta},
      year={2021},
      eprint={2108.11515},
      archivePrefix={arXiv},
      primaryClass={cs.CV}
}

@article{comanici2025gemini,
  title        = {Gemini 2.5: Pushing the frontier with advanced reasoning, multimodality, long context, and next generation agentic capabilities},
  author       = {Comanici, Gheorghe and Bieber, Eric and Schaekermann, Mike and Pasupat, Ice and Sachdeva, Noveen and others},
  year         = 2025,
  journal      = {arXiv preprint arXiv:2507.06261}
}

@article{sklyarova_kabadayi_2025neuralfur,
    title={NeuralFur: NeuralFur: Animal Fur Reconstruction from Multi-view Images},
    author={Sklyarova, Vanessa and Kabadayi, Berna and Yiannakidis, Anastasios and Becherini, Giorgio and Black, Michael J. and Thies, Justus},
    journal={ArXiv},
    month={January}, 
    year={2026} 
}

@inproceedings{kabadayi2024ganavatar,
  title={Gan-avatar: Controllable personalized gan-based human head avatar},
  author={Kabadayi, Berna and Zielonka, Wojciech and Bhatnagar, Bharat Lal and Pons-Moll, Gerard and Thies, Justus},
  booktitle={2024 International Conference on 3D Vision (3DV)},
  pages={882--892},
  year={2024},
  organization={IEEE}
}

@inproceedings{tewari2019fml,
  title={Fml: Face model learning from videos},
  author={Tewari, Ayush and Bernard, Florian and Garrido, Pablo and Bharaj, Gaurav and Elgharib, Mohamed and Seidel, Hans-Peter and P{\'e}rez, Patrick and Zollhofer, Michael and Theobalt, Christian},
  booktitle={Proceedings of the IEEE/CVF Conference on Computer Vision and Pattern Recognition},
  pages={10812--10822},
  year={2019}
}

@article{teotia2024gaussianheads,
  title={Gaussianheads: End-to-end learning of drivable gaussian head avatars from coarse-to-fine representations},
  author={Teotia, Kartik and Kim, Hyeongwoo and Garrido, Pablo and Habermann, Marc and Elgharib, Mohamed and Theobalt, Christian},
  journal={ACM Transactions on Graphics (ToG)},
  volume={43},
  number={6},
  pages={1--12},
  year={2024},
  publisher={ACM New York, NY, USA}
}

@inproceedings{aneja2025scaffoldavatar,
  title={Scaffoldavatar: High-fidelity gaussian avatars with patch expressions},
  author={Aneja, Shivangi and Weiss, Sebastian and Baeza, Irene and Chandran, Prashanth and Zoss, Gaspard and Niessner, Matthias and Bradley, Derek},
  booktitle={Proceedings of the Special Interest Group on Computer Graphics and Interactive Techniques Conference Conference Papers},
  pages={1--11},
  year={2025}
}
}

\clearpage
\setcounter{page}{1}
\maketitlesupplementary

\providecommand{\OURS}{PhysHead\xspace}

\providecommand{\orange}[1]{{\color{orange}#1}}

\definecolor{r2color}{rgb}{0.0, 0.53, 0.74}

\definecolor{r3color}{rgb}{0.55, 0.71, 0.0}

This supplementary material provides additional details and results for PhysHead.
\section{Evaluation of Hair Dynamics and User Study}
\begin{figure}[h]
  \centering
  \vspace{-0.2cm}
  \includegraphics[width=0.8\linewidth]{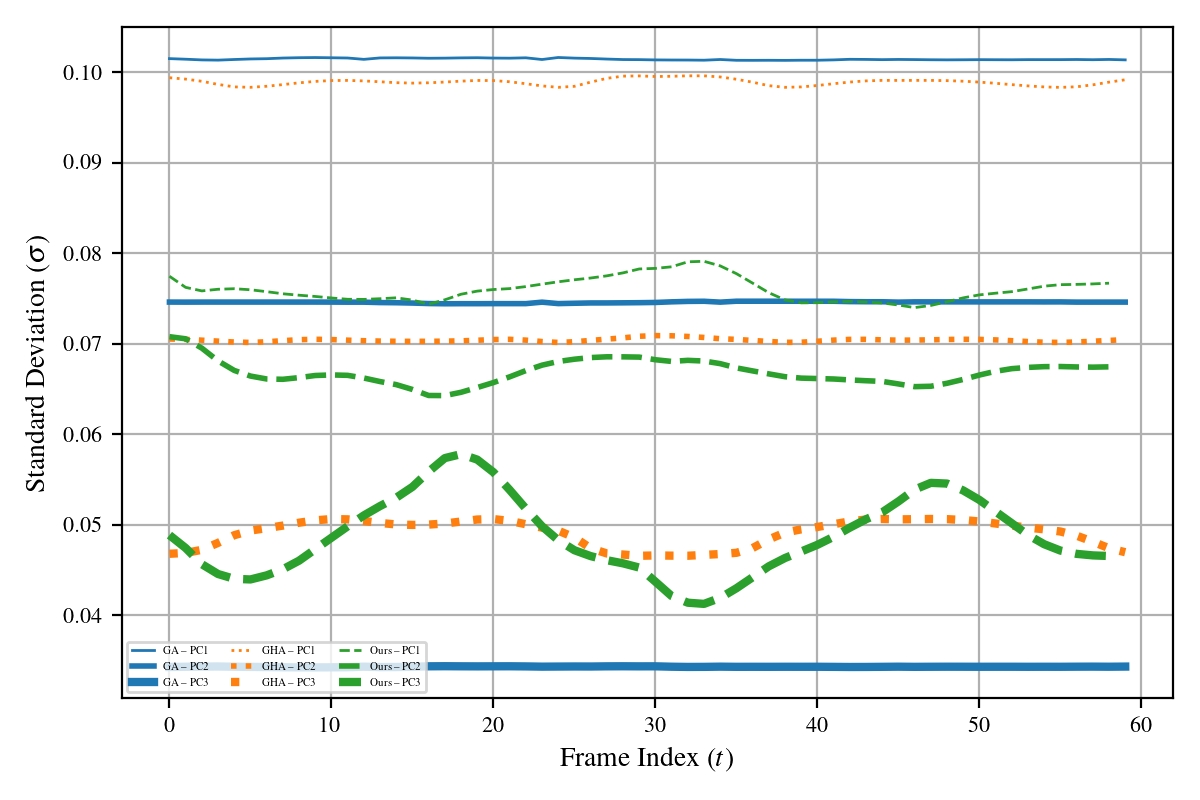}
  \caption{
    Analysis of the riggidity of the hair geometry on a nodding sequence.
    Our method (green) handles dynamic effects, others are static resulting in a constant std. dev. of geometry.
  }
   \label{fig:std}
\end{figure}

We conducted a preliminary user study with 26 participants which evaluated \textit{physical plausibility} and \textit{structural coherence}, finding our method significantly more plausible and coherent (\cref{tab:pcatime}).
In addition, we provide quantitative metrics that capture the non-rigid behavior of dynamic hair using  \textit{E. Variance PC1 (\%)} (lower indicates dynamic hair) and \textit{Temporal Smoothness} in PCA temporal space.
In \cref{fig:std}, we also show a graph of the std. dev. of the hair geometry in PCA space which effectively factors out rigid transformations and gives an indication of the hair deformations.
Especially, the std. dev. of the third principle component shows the hair deformation during the nodding sequence.
GA~\cite{Qian2024gaussianavatars} and GHA~\cite{xu2023gaussianheadavatar}) have almost static hair (std. dev. does not change), while ours is dynamic.

\begin{table}[h]
\centering
\setlength{\tabcolsep}{3pt}
\resizebox{0.8\columnwidth}{!}{%
\begin{tabular}{lccc}
Metric & GA\cite{Qian2024gaussianavatars} & GHA\cite{xu2023gaussianheadavatar} & Ours \\
\hline
Explained Variance PC1 (\%) $\downarrow$ &  95.42 &  96.16 & \textbf{93.28} \\
Temporal Smoothness (TS) $\downarrow$ & 6.23  & 2.13 &  \textbf{0.863} \\
\hline
User Study -- Physical Plausibility $\uparrow$ & 0  &  1.5 \% & \textbf{98.5 \%} \\
User Study -- Structural Coherence $\uparrow$ & 8.5 \%  & 3.1 \% & \textbf{88.5 \%}  \\
\end{tabular}
}
\caption{E. Variance PC1 (\%) shows how much of the motion of hair 3D Gaussians is captured by a single principle component. TS measures frame to frame change in PCA space. %
} 
\label{tab:pcatime}
\vspace{-0.2cm}
\end{table}

\section{Implementation Details}

\subsection{Optimization} Following Gaussian Avatars~\cite{Qian2024gaussianavatars}, we train the dynamic face region for 600k iterations using the Adam optimizer~\cite{Kingma2014AdamAM}, adopting their hyperparameter configuration. The hair region is optimized separately for 50k iterations. We first optimize colors in the visible regions to obtain stable estimates, then introduce a color consistency loss to propagate these colors to the invisible regions after 3k iterations.

\subsection{Dataset} We conduct our experiments on the Ava-256 dataset~\cite{martinez2024codec}. The list of actors used for our method is provided in Table~\ref{tab:actors}.

\begin{table}[h]
    \centering
    \caption{\textbf{Actors IDs.} List of actors from the Ava-256~\cite{martinez2024codec} dataset used in our experiments.}
    \renewcommand{\arraystretch}{1.2}
    \begin{tabular}{|c|}
        \hline
        \textbf{Actors IDs.} \\
        \hline
        20230313--0819--GCZ208 \\
        20230313--1653--RHL466 \\
        20230512--1009--ZEL435 \\
        20230810--1355--AJR151 \\
        20211006--0836--EID363 \\
        20230124--1415--KKF424 \\
        20230310--1106--FCT871 \\
        20230502--0816--ORZ494 \\
        20230602--1453--FMB793 \\
        20230308--1352--BDF920 \\
        20230726--1657--AYE877 \\
        \hline
    \end{tabular}
    \label{tab:actors}
\end{table}

\subsection{Baselines} For comparison with GH~\cite{zakharov2024gh}, for each actor in Ava-256~\cite{martinez2024codec}, we split the cameras into train and test sets, resulting in 64 and 16 cameras, respectively, with every 5th camera used for testing. We use the expression \textit{cheek002}. 
Gaussian Haircut~\cite{zakharov2024gh} results are provided by the corresponding authors.
We optimize the appearance of hairstyles obtained from Neural Haircut~\cite{sklyarova2023neural} on the training cameras and evaluate appearance results on the test set. 
Gaussian Avatars~\cite{Qian2024gaussianavatars} and Gaussian Head Avatar~\cite{xu2023gaussianheadavatar} are run using the public repositories and trained on the same expressions and cameras.

\subsection{Hair Simulation}
\noindent\textbf{Runtime of physics-based hair animation.}
Our hair simulation and sparse-to-dense interpolation run offline. Simulating 700 guide strands (16 points each) in Maya takes approximately 1s per frame on a MacBook Air (M3, 16GB RAM). The rendering of the Gaussian-based hair animation is real-time capable, taking roughly 0.0033s per frame, which is comparable to GA’s~\cite{Qian2024gaussianavatars} 0.0017s per frame for the same actor.

\noindent\textbf{Hair Simulation Parameters. } The physical parameters used for dynamic hair simulation are summarized in Table~\ref{tab:hair_params}.

\begin{table}[h]
    \centering
    \caption{\textbf{Hair system parameters used in our simulation.} The \textit{Start Curve Attract} parameter varies slightly across different hairstyles.}
    \renewcommand{\arraystretch}{1.2}
    \begin{tabular}{l c}
        \hline
        \textbf{Parameter} & \textbf{Value} \\
        \hline
        Self Collide & 1 \\
        Friction & 0.51087 \\
        Stickiness & 0.51087 \\
        \hline
        Stretch Resistance & 600 \\
        Compression Resistance & 600.0 \\
        Bend Resistance & 10 \\
        Twist Resistance & 1.718 \\
        Extra Bend Links & 1.0 \\
        \hline
        Mass & 2 \\
        Drag & 0.65 \\
        Tangential Drag & 0.096 \\
        Damp & 0.25 \\
        Stretch Damp & 1 \\
        Dynamics Weight & 1 \\
        Static Cling & 0.025 \\
        Start Curve Attract & 2.0 \\
        \hline
    \end{tabular}
    \label{tab:hair_params}
\end{table}

\paragraph{From simulated guiding strands to dense hairstyle.}  
Given a sparse set of hair strands, the motion data is represented as:  
\begin{equation}
    S_t \in \mathbb{R}^{N_s \times N_{\text{seg}} \times 3} ~~,
\end{equation}  
where \( S_t \) represents the sparse strands at time \( t \), \( N_s \) is the number of sparse strands, \( N_{\text{seg}} \) is the number of segments per strand, and each segment is represented in 3D space. In our experiments, we use \( N_s = 700 \) sparse strands, each consisting of \( 16 \) points (i.e., \( 15 \) segments) for the simulation.

For each dense strand \( d_i \), we find the nearest \( k = 10 \) sparse strands based on their root positions (first segment at \( t=0 \)):  
\begin{equation}
    \mathcal{N}(d_i) = \operatorname{KNN}( S_0[:,0,:], d_i(0) ) ~~,
\end{equation}
where \( \operatorname{KNN}(\cdot) \) finds the nearest neighbors based on Euclidean distance.

The influence of each sparse strand on a dense strand is computed using inverse distance weighting:  
\begin{equation}
    w_{i,j} = \frac{1}{d_{i,j} + \epsilon}, \quad w_{i,j} \leftarrow \frac{w_{i,j}}{\sum_{j=1}^{k} w_{i,j}} ~~,
\end{equation}  
where \( d_{i,j} \) is the Euclidean distance between the root of dense strand \( i \) and the root of its \( j \)-th nearest sparse neighbor, and \( \epsilon \) is a small constant to avoid division by zero.

For each dense strand \( d_i \) at time \( t \), the motion is computed using the weighted displacement of its nearest sparse strands:  
\begin{equation}
    \Delta S_t (i) = \sum_{j=1}^{k} w_{i,j} \cdot \left( S_t(j) - S_{t-1}(j) \right)
\end{equation}
\begin{equation}
    D_t(i) = D_{t-1}(i) + \Delta S_t(i)
\end{equation}  
where \( \Delta S_t(i) \) is the weighted relative motion transferred from sparse to dense strands, and \( D_t(i) \) is the updated dense strand position at frame \( t \).  

For hair reconstruction, we use a dense hairstyle consisting of \( 60{,}000 \) strands, each with \( 50 \) points, resulting in \( 49 \) Gaussian primitives per strand.

\section{Preliminaries}
\noindent\textbf{3D Gaussian Splatting.}
3D Gaussian Splatting (3DGS)~\cite{Kerbl20233DGS} reconstructs and renders static multi-view scenes from novel viewpoints.
Kerbl et al.~\cite{Kerbl20233DGS} model the scene using 3D Gaussians~\cite{Wang2019DifferentiableSS, Kopanas2021PointBasedNR}, parameterized by a mean vector $\mathbf{\mu}$ and a 3D covariance matrix $\mathbf{\Sigma}$:
\begin{equation}
\label{formula: gaussian's formula}
    G(\mathbf{x})=e^{-\frac{1}{2}(\mathbf{x-\mu})^T\mathbf{\Sigma}^{-1}(\mathbf{x-\mu})}.
\end{equation}
To render these 3D Gaussians, they must be projected onto the 2D image plane \cite{Zwicker2001SurfaceS}. However, perspective projection is inherently non-linear due to division by depth, making direct computation of the 2D covariance challenging. To approximate this transformation, the first-order Taylor expansion is applied to compute the Jacobian matrix $\mathbf{J}$, which locally linearizes the mapping from 3D to 2D. The 2D covariance matrix is then obtained as:
\begin{equation}
    \mathbf{\Sigma}^{\prime} = \mathbf{J} \mathbf{\Sigma} \mathbf{J}^T,
\end{equation}
where $\mathbf{\Sigma}^{\prime}$ represents the projected 2D covariance matrix. The Jacobian matrix $\mathbf{J}$ is computed from the perspective projection function:
\begin{equation}
    \mathbf{J} = 
    \begin{bmatrix}
        \frac{f}{z} & 0 & -\frac{fx}{z^2} \\
        0 & \frac{f}{z} & -\frac{fy}{z^2}
    \end{bmatrix},
\end{equation}
In this formulation, the Jacobian $\mathbf{J}$ accounts for the local changes in image-space coordinates to variations in 3D space. 
Since direct optimization of the covariance matrix $\mathbf{\Sigma}$ is difficult due to the requirement that it remains positive semidefinite, Kerbl et al.~\cite{Kerbl20233DGS} instead decompose it into a scale matrix $\mathbf{S}$ and a rotation matrix $\mathbf{R}$, representing the 3D Gaussian as a 3D ellipsoid:
\begin{equation}
    \mathbf{\Sigma} = \mathbf{R}\mathbf{S}\mathbf{S}^T\mathbf{R}^T.
\end{equation}
Furthermore, to approximate the diffuse component of the Bidirectional Reflectance Distribution Function (BRDF)~\cite{Goral1984ModelingTI}, 3DGS employs spherical harmonics (SH) following the method of Ramamoorthi et al.~\cite{Ramamoorthi2001AnER}. 
Four bands of SH coefficients are used, leading to a 48-dimensional vector representation that captures view-dependent color and global illumination.

\begin{figure}[tb]
\centering

\begin{minipage}{0.06\linewidth}
    \centering
    \footnotesize
    \rotatebox{90}{HairCUP~\cite{kim2025haircup}} \\[22mm]    %
    \rotatebox{90}{Ours}
\end{minipage}
\begin{minipage}{0.9\linewidth}
    \includegraphics[width=\linewidth,keepaspectratio=true,
        trim=0cm 1cm 0cm 0cm, clip]{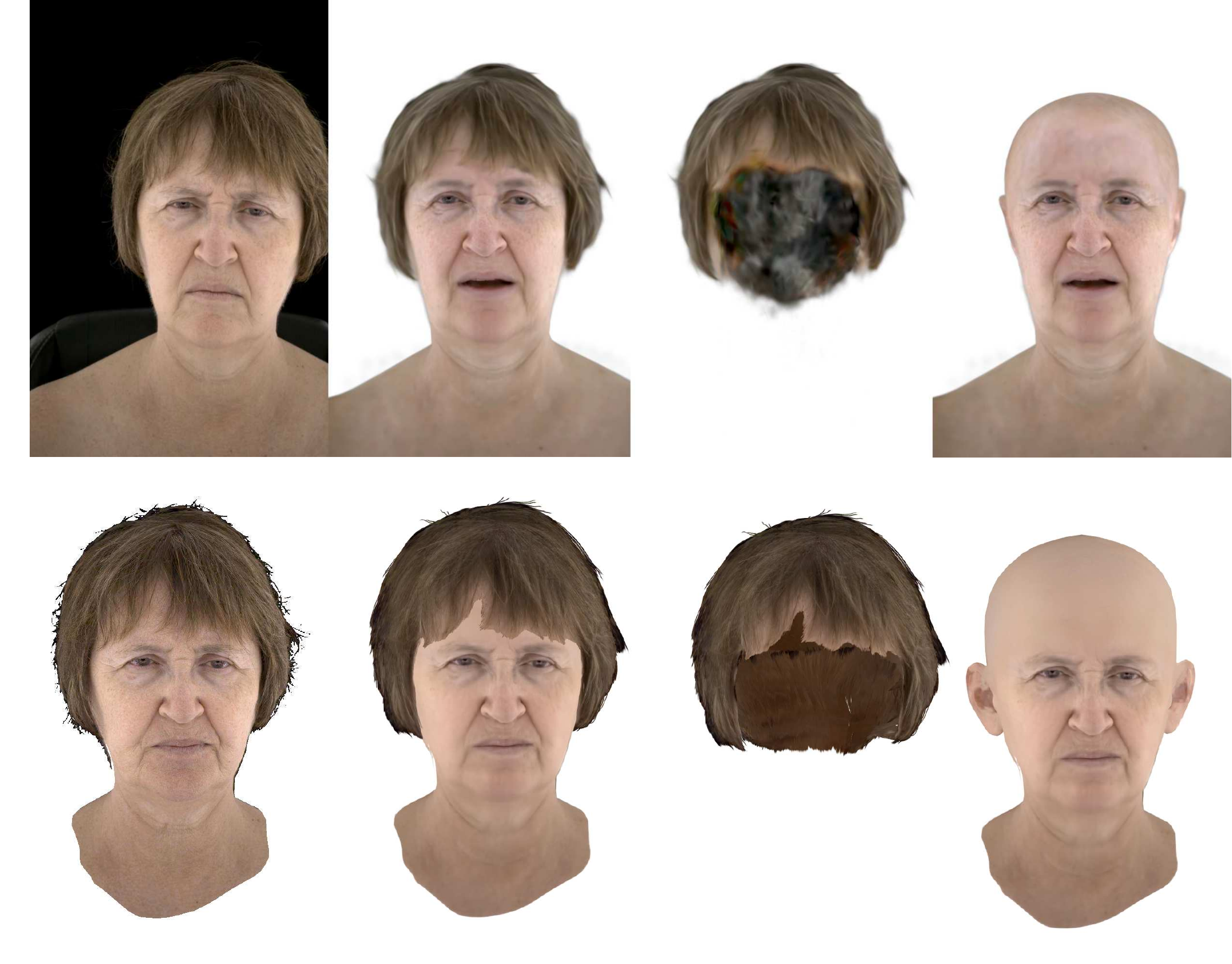}
\end{minipage}

\begin{minipage}{0.06\linewidth}
    \centering
    \footnotesize
    \rotatebox{90}{HairCUP~\cite{kim2025haircup}} \\[22mm]    %
    \rotatebox{90}{Ours}
\end{minipage}
\begin{minipage}{0.9\linewidth}
    \includegraphics[width=\linewidth]{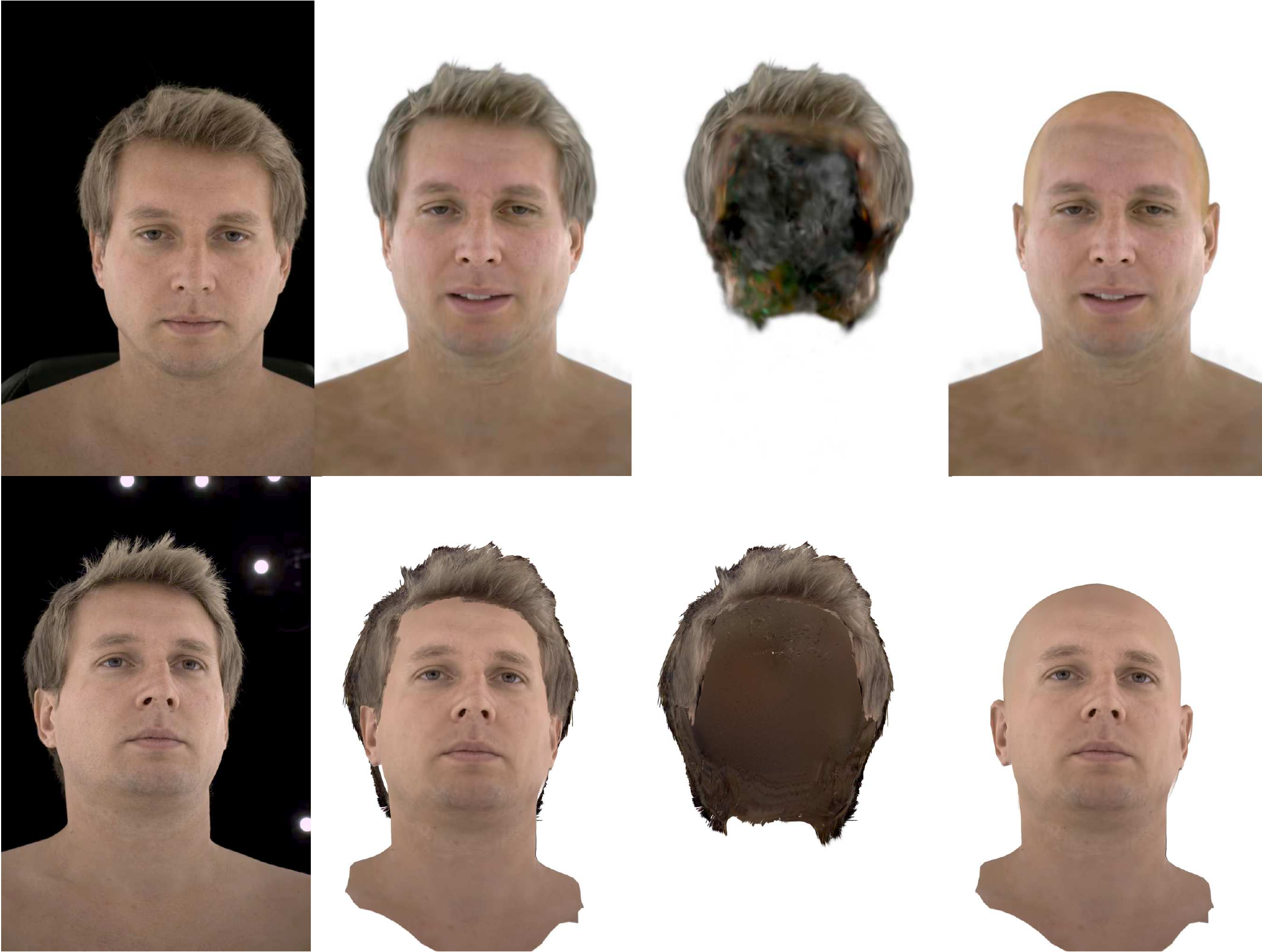}
\end{minipage}

\begin{minipage}{0.06\linewidth}
    \centering
    \footnotesize
    \rotatebox{90}{HairCUP~\cite{kim2025haircup}} \\[22mm]    %
    \rotatebox{90}{Ours}
\end{minipage}
\begin{minipage}{0.9\linewidth}
    \includegraphics[width=\linewidth]{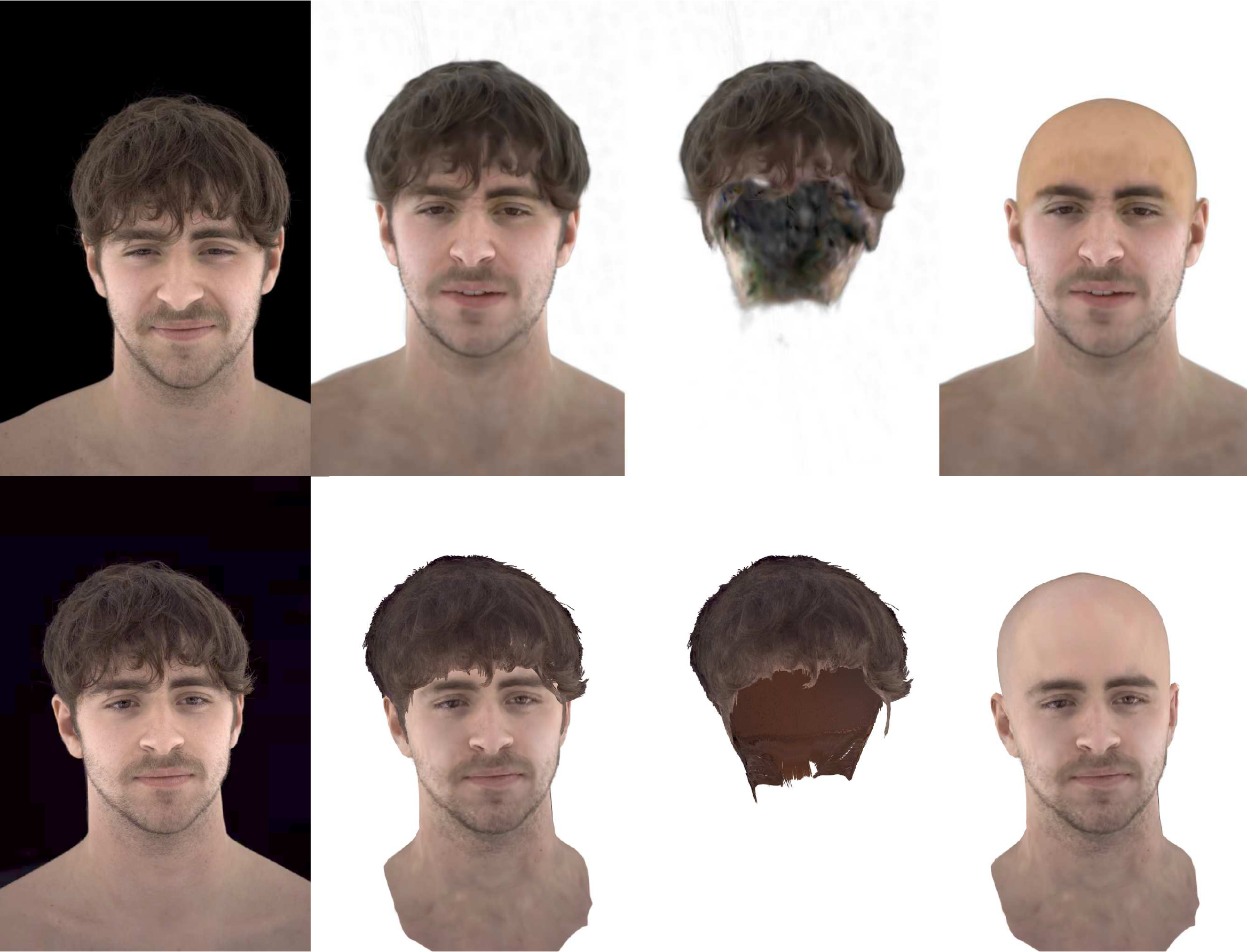}
\end{minipage}

\vspace{4pt}
\footnotesize
\begin{tabularx}{0.96\linewidth}{YYYY}
    Ground Truth &
    Composed Avatar &
    Hair Layer &
    Head Layer
\end{tabularx}

\caption{\textbf{Extended HairCUP~\cite{kim2025haircup} comparison.} HairCUP images are sourced from the original publication. As shown, HairCUP may produce artifacts and exhibits limited skin-tone variability across subjects, while our method mitigates these issues.}
\label{fig:haircup_supmat}
\end{figure}

\section{Additional Experiments}

\begin{figure}[tb]
    \centering
    \begin{minipage}{0.05\linewidth}
        \centering
        \footnotesize
        \rotatebox{90}{ t=65 \hspace{5em} t=13 \hspace{5em} t=0}
    \end{minipage}%
    \begin{minipage}{1.0\linewidth} %
        \centering
        \includegraphics[width=\linewidth, keepaspectratio]{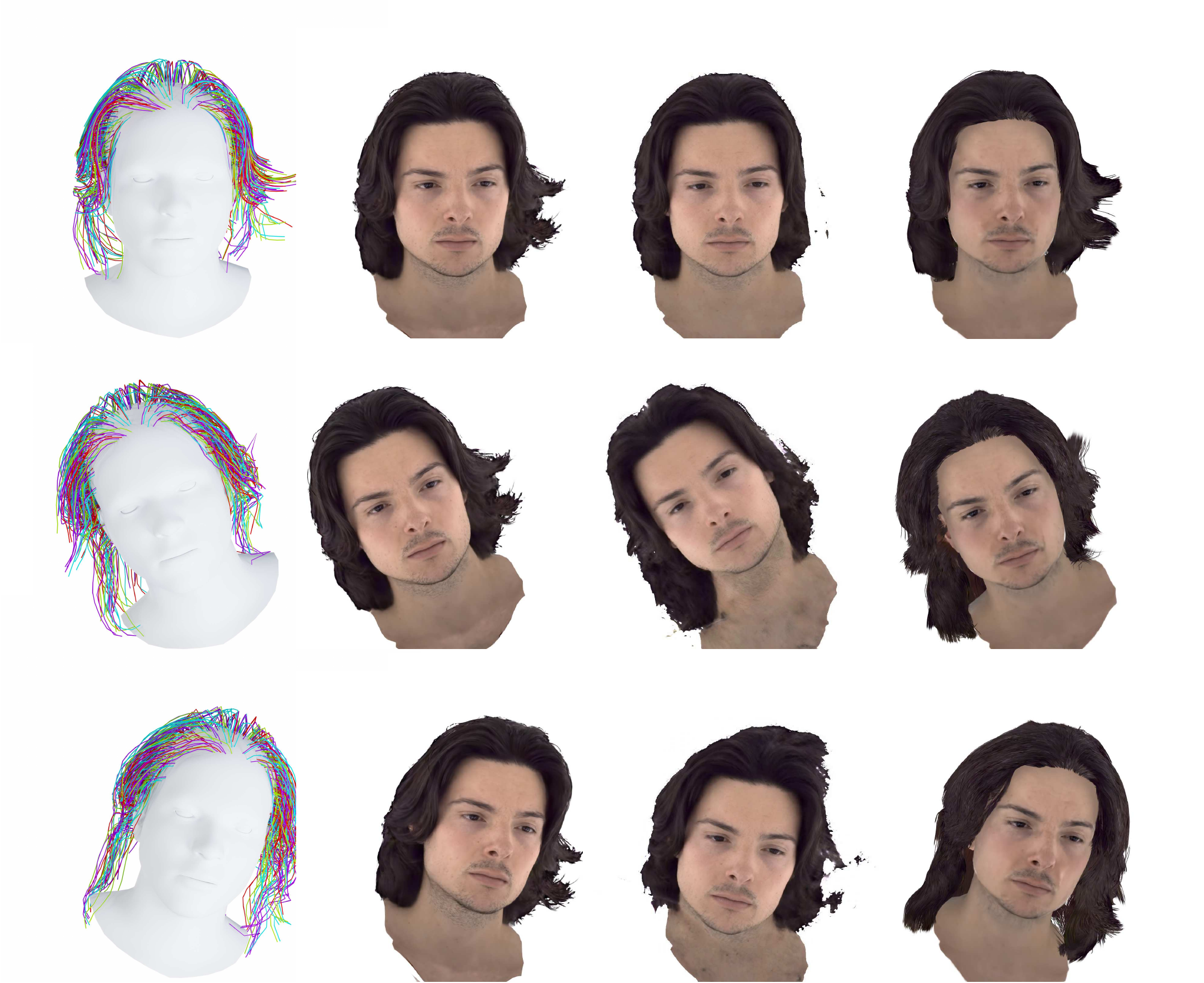}\\
    \end{minipage}

    \vspace{0.5em}
    
    \noindent
    \makebox[\linewidth][c]{%
        \parbox{0.24\linewidth}{\centering Driving}%
        \hfill
        \parbox{0.24\linewidth}{\centering GA~\cite{Qian2024gaussianavatars}}%
        \hfill
        \parbox{0.24\linewidth}{\centering GHA~\cite{xu2023gaussianheadavatar}}%
        \hfill
        \parbox{0.24\linewidth}{\centering Ours}%
    }
    
    \caption{\textbf{Extended qualitative comparison.} All methods synthesize photorealistic avatars. GA produces rigid hair that remains nearly identical in each frame. GHA exhibits artifacts due to being controlled by a BFM-like model. In contrast, our method, PhysHead, benefits from a strand-based representation and generalizes well under novel driving signals.}
    \label{fig:maincomparison_supmat}
\end{figure}

In this section, we provide additional comparisons of bald avatar appearance generation using our method and HairCUP~\cite{kim2025haircup} in Figure~\ref{fig:haircup_supmat}, along with extended results of our bald image generation pipeline in Figures~\ref{fig:bald1}, \ref{fig:bald2}, and \ref{fig:bald3}. Next, we compare hair dynamics and appearance with baseline methods in Figures~\ref{fig:maincomparison_supmat} and \ref{fig:gh_additional}. Finally, we present results of optimizing the head and hair as a single layer in Figure~\ref{fig:singlelayer}.

\subsection{Generation of Bald Avatar}
 
We use capabilities of Vision Language Model (VLM) to create realistic bald avatar.
For our experiments we use the \textit{gemini-2.5-flash-image-preview} model to process images of each actor captured from 16 different cameras at a single timestep. For each input image, our prompt is  \textit{make him/her bald, don’t change skin color}. Overall, our API calls works reliably, though in rare cases (2 or less images for couple of actors), the generation fails due to security policies of the model. Despite these, the approach is effective for producing consistent bald-head images. After generating bald images, we overlay resize them to original input size and overlay with the original images. If face region is sharp, we consider that the generated view is consistent with the original capture, remove the background of generated~\cite{rvm}. Finally, we optimize the joint FLAME UV map for a single timestep using landmark and photometric losses with a differentiable rasterizer. 

\begin{figure}[t]
\centering

\begin{minipage}{0.96\linewidth}
    \includegraphics[width=\linewidth,
        trim=0cm 3cm 0cm 2cm, clip]{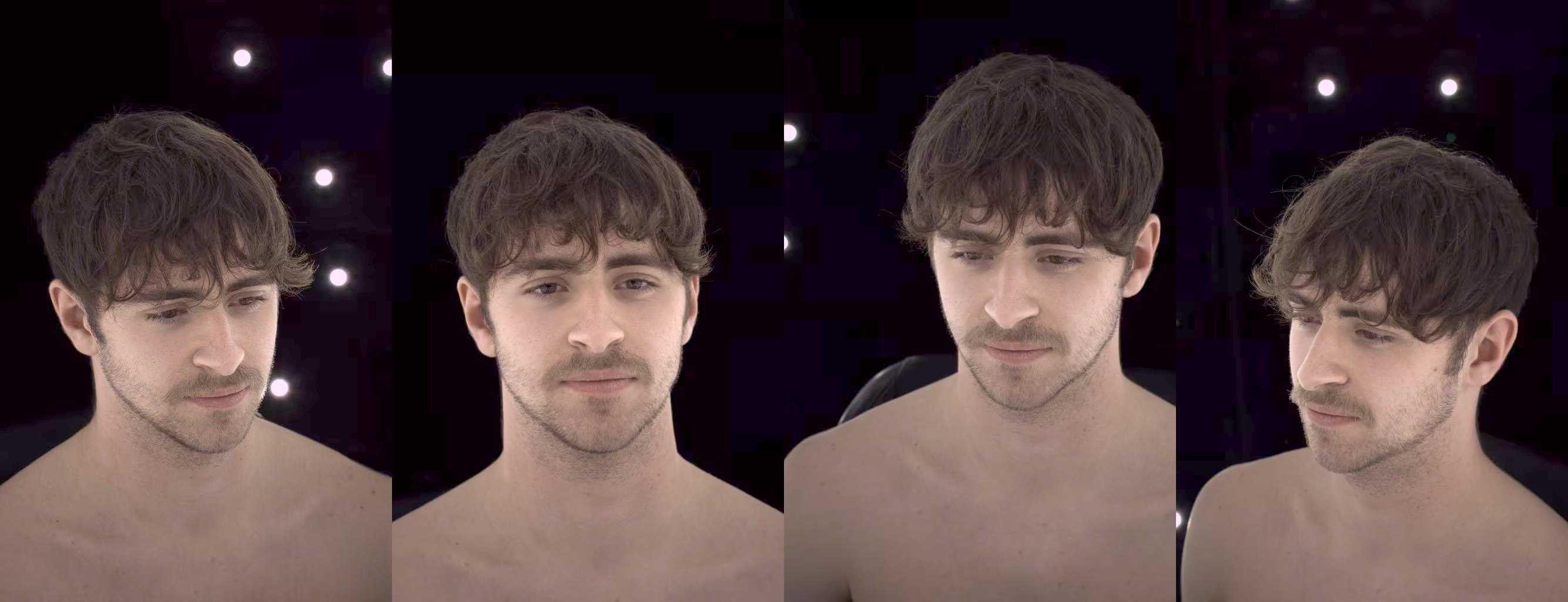}
\end{minipage}

\begin{minipage}{0.96\linewidth}
    \includegraphics[width=\linewidth,
        trim=0cm 3cm 0cm 2cm, clip]{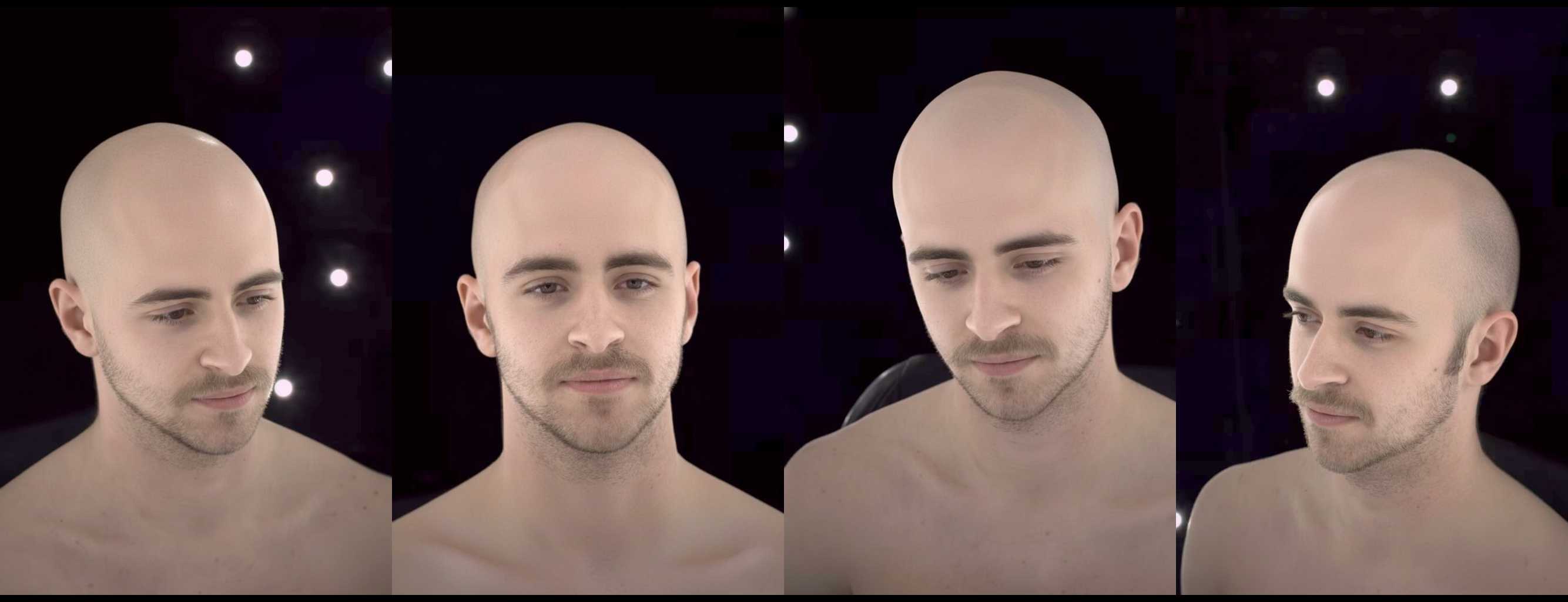}
\end{minipage}

\caption{\textbf{VLM~\cite{zakharov2024gh} failure cases.} Given the images in the first row, the VLM model generates bald images in the second row. However, the VLM model may alter the head pose (see the example in the third column). We filter out such cases and only use multiview-consistent views. Note that VLM-generated images are never used in our method; they are only employed to model regions such as the ear, hairline, or scalp.}

\label{fig:nanofail}

\end{figure}

\begin{figure*}[t]
  \centering
  \includegraphics[width=\linewidth]{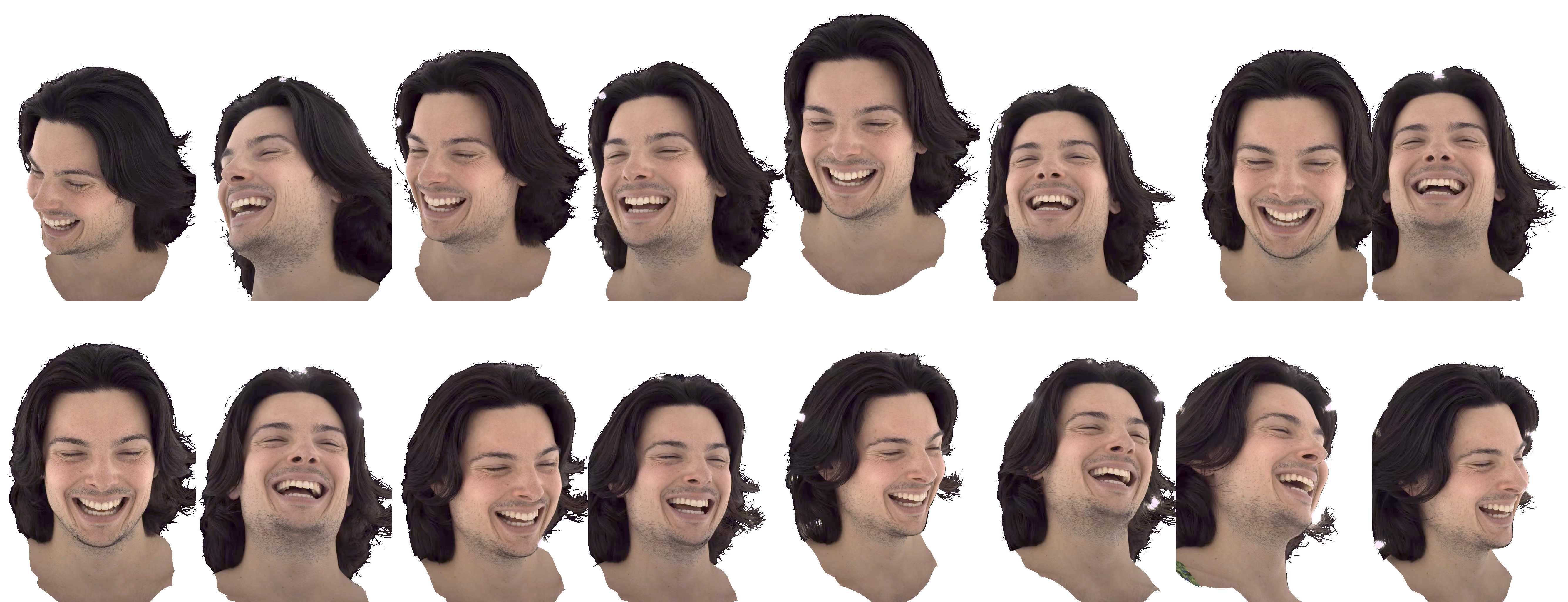}
  \includegraphics[width=\linewidth]{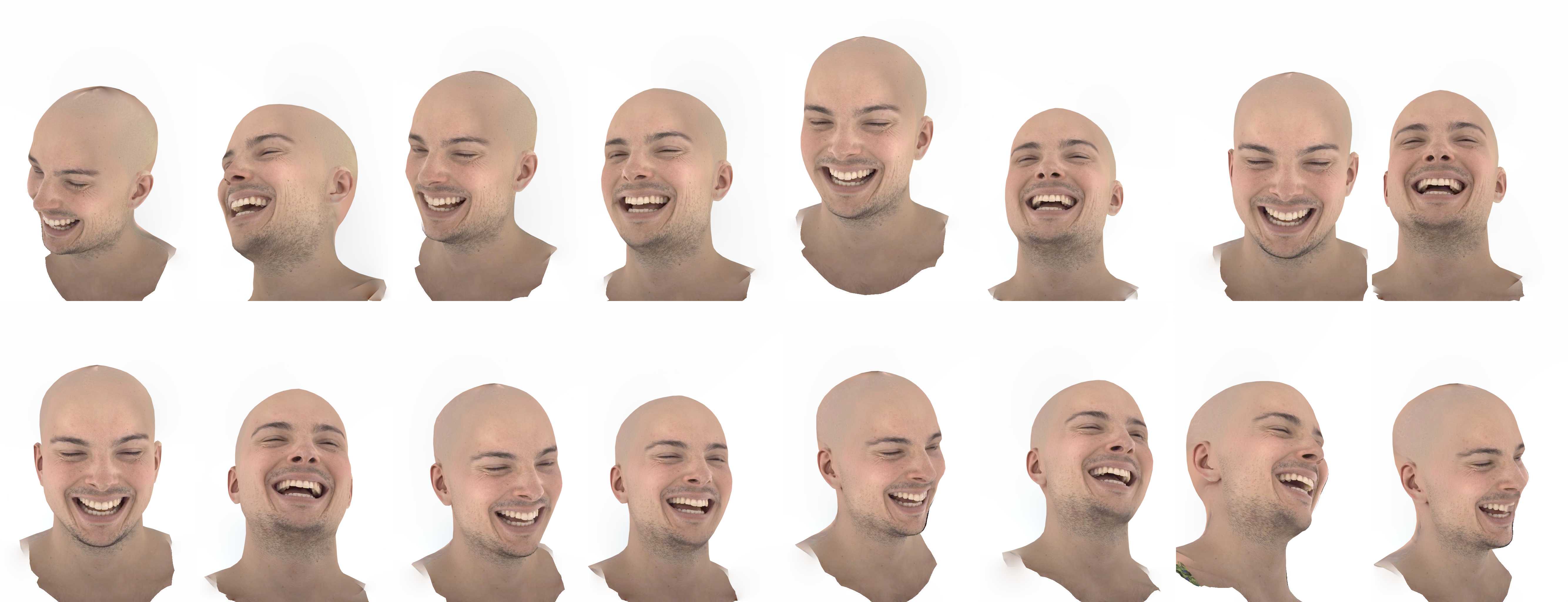}
  \caption{\textbf{Additional VLM-Based Bald Image Generation Results.}}
  \label{fig:bald1}
\end{figure*}

\begin{figure*}[t]
  \centering
  \includegraphics[width=\linewidth]{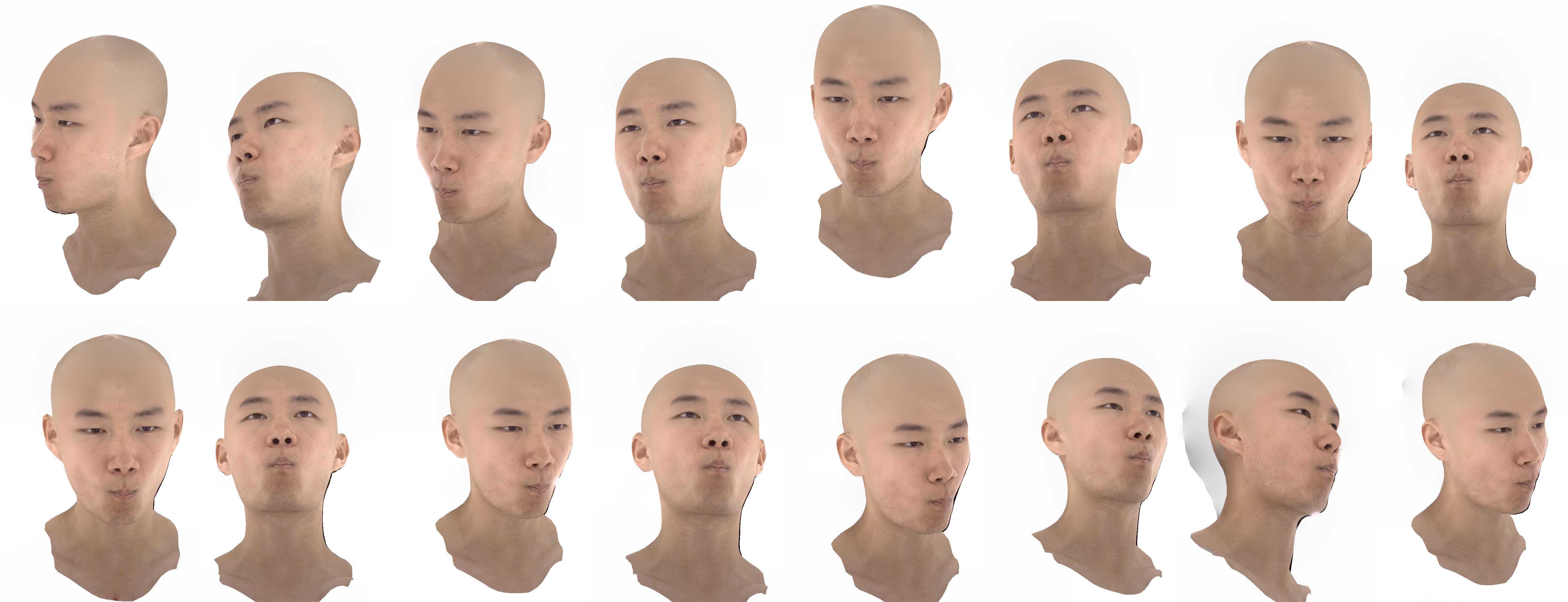}
  \includegraphics[width=\linewidth]{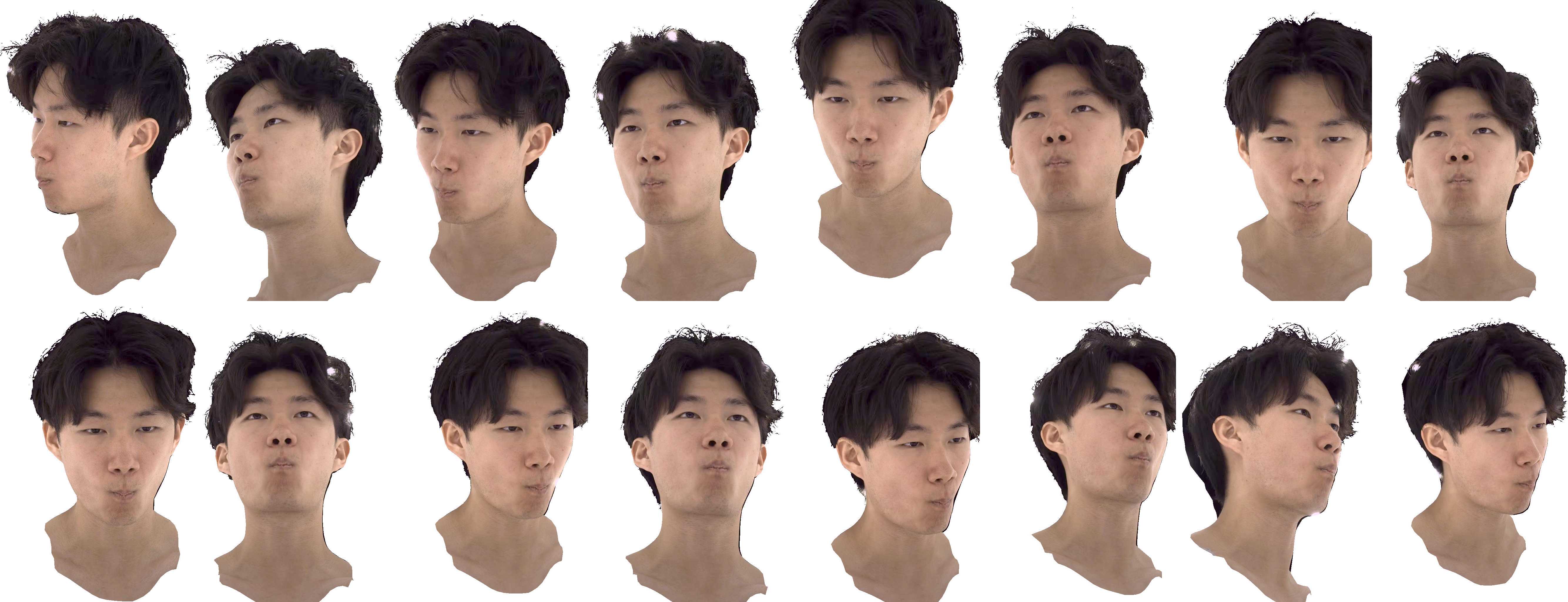}
  \caption{\textbf{Additional VLM-Based Bald Image Generation Results.}}
  \label{fig:bald2}
\end{figure*}

\begin{figure*}[t]
  \centering
  \includegraphics[width=\linewidth]{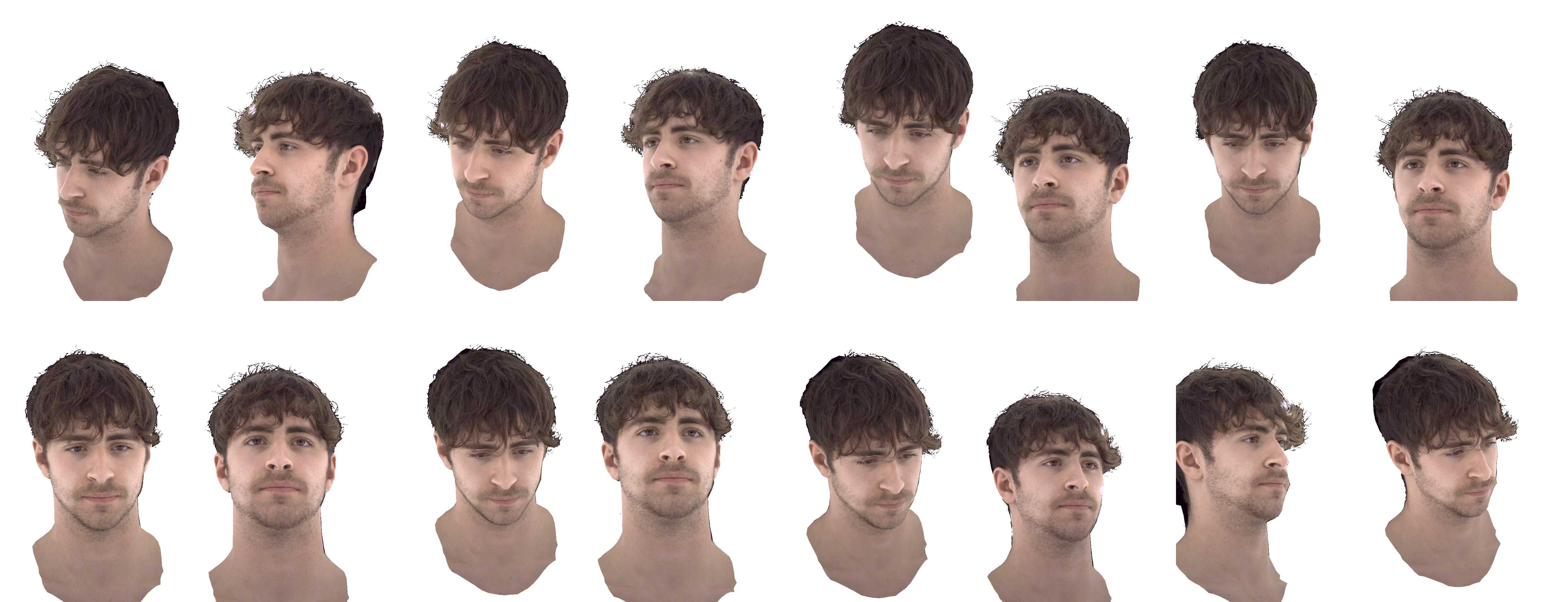}
  \includegraphics[width=1.0\linewidth]{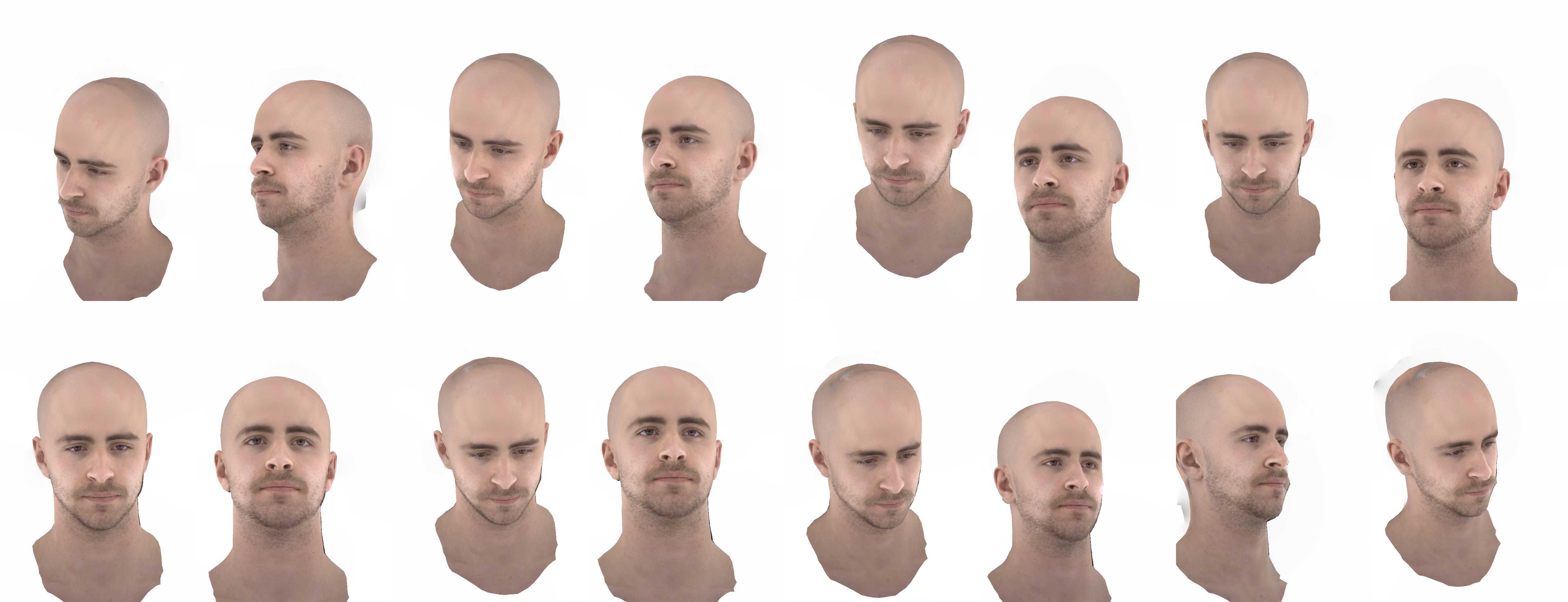}
  \caption{\textbf{Additional VLM-Based Bald Image Generation Results.}}
  \label{fig:bald3}
\end{figure*}

We provide additional comparisons of bald avatar generation using our method and HairCUP~\cite{kim2025haircup} in Figure~\ref{fig:haircup_supmat}. Note that HairCUP~\cite{kim2025haircup} code is not publicly available, so we use images provided in the paper. We noticed that HairCUP~\cite{kim2025haircup} exhibits limited skin-tone variability across different subjects, that might be an effect of using SDS loss. Our approach helps to better preserve color and produce consistent results. Furthermore, we show more results of our method in Figures~\ref{fig:bald1}, \ref{fig:bald2}, and \ref{fig:bald3}. We also include some failure cases of the VLM model in Figure~\ref{fig:nanofail}. VLM sometimes alters the head pose, which breaks the multiview consistency required for further optimization. To mitigate this issue, we filter out such cases.

\subsection{Qualitative Comparison with Baselines}
\begin{figure}[t]
\centering

\begin{minipage}{0.90\linewidth}
    \includegraphics[width=\linewidth,
        trim=0cm 3cm 0cm 2cm, clip]{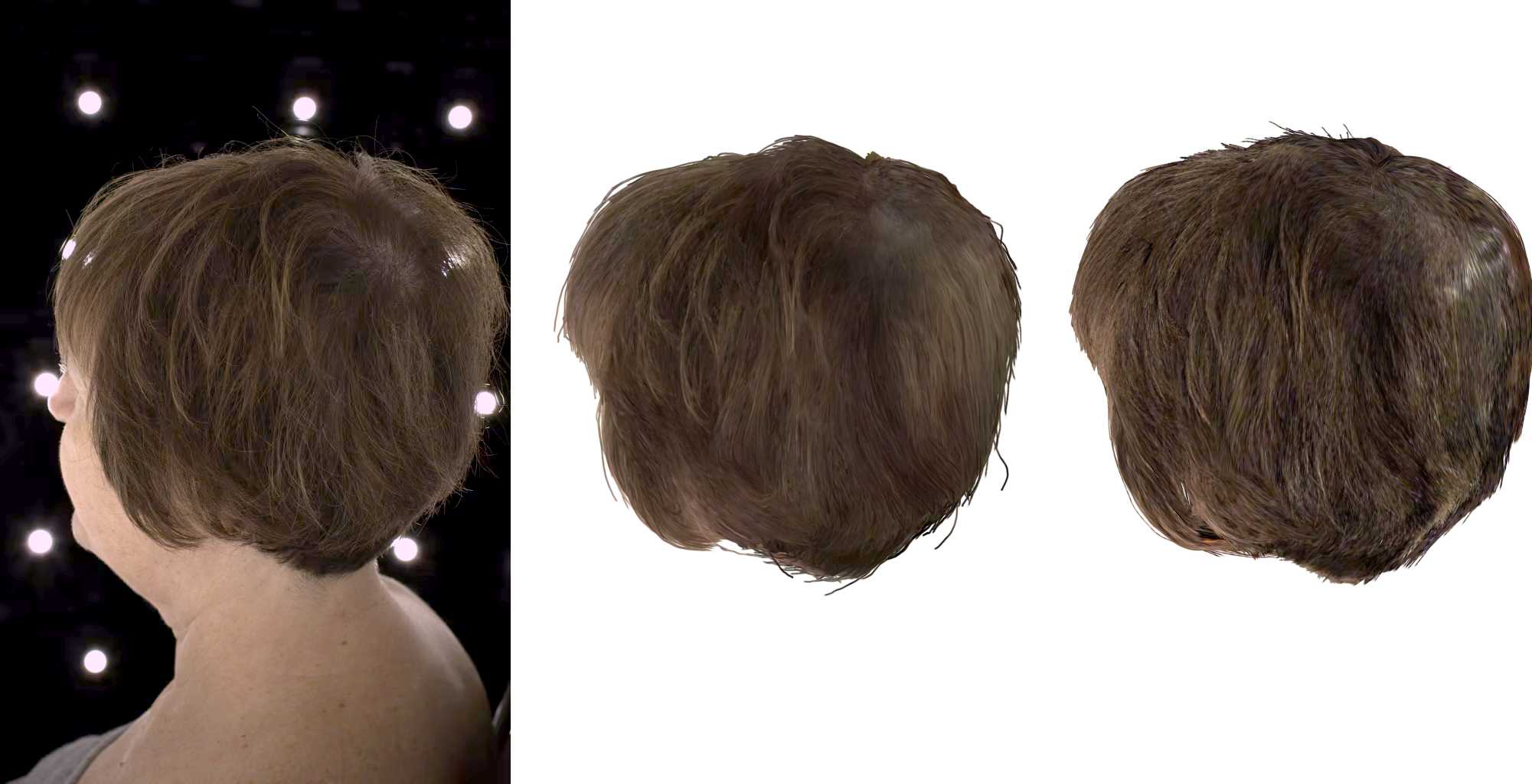}
\end{minipage}

\vspace{0.5mm}

\begin{minipage}{0.90\linewidth}
    \includegraphics[width=\linewidth,
        trim=0cm 3cm 0cm 2cm, clip]{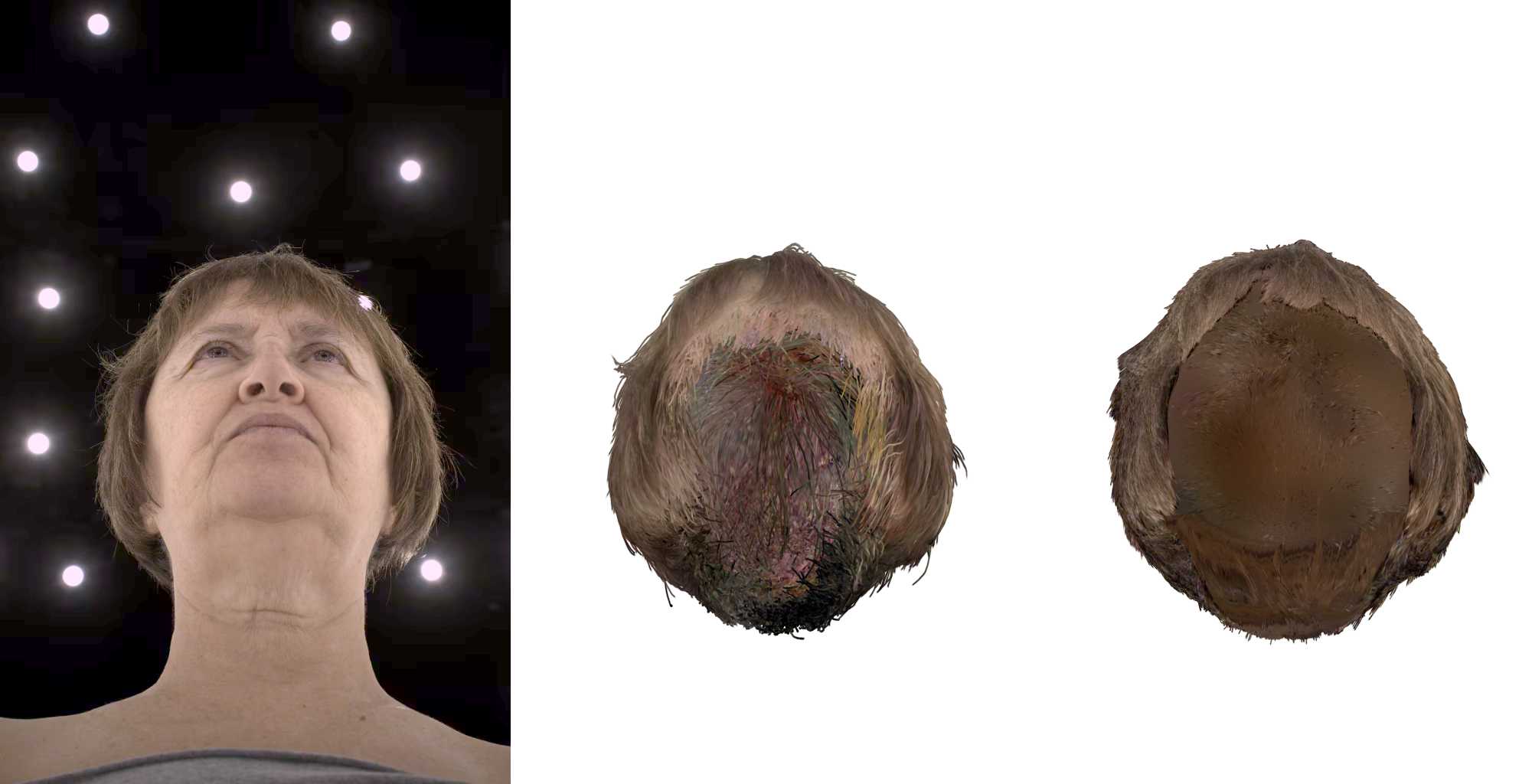}
\end{minipage}

\vspace{0.5mm}

\begin{minipage}{0.90\linewidth}
    \includegraphics[width=\linewidth,
        trim=0cm 2cm 0cm 3cm, clip]{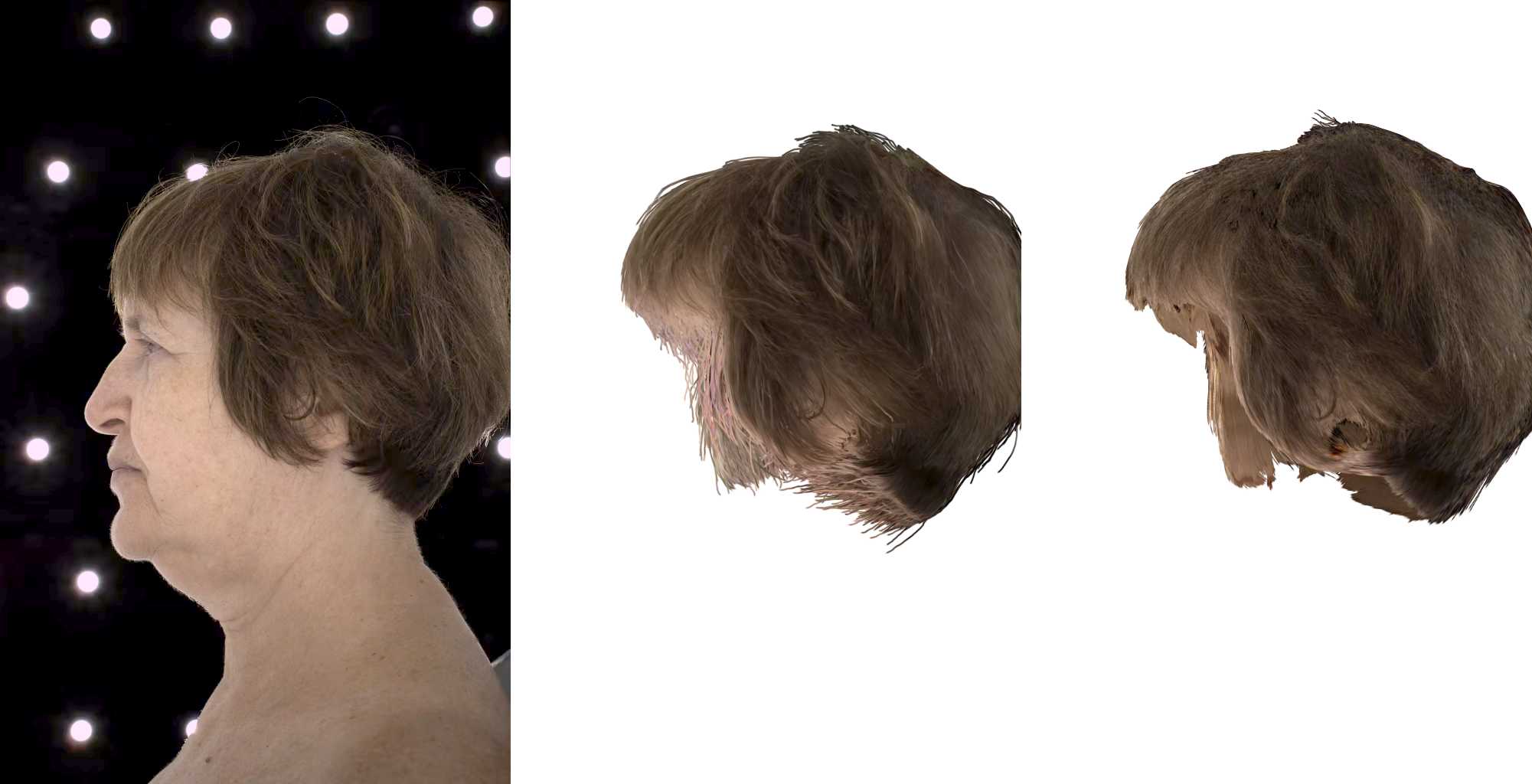}
\end{minipage}

\vspace{0.5mm}

\begin{minipage}{0.90\linewidth}
    \includegraphics[width=\linewidth,
        trim=0cm 5cm 0cm 0cm, clip]{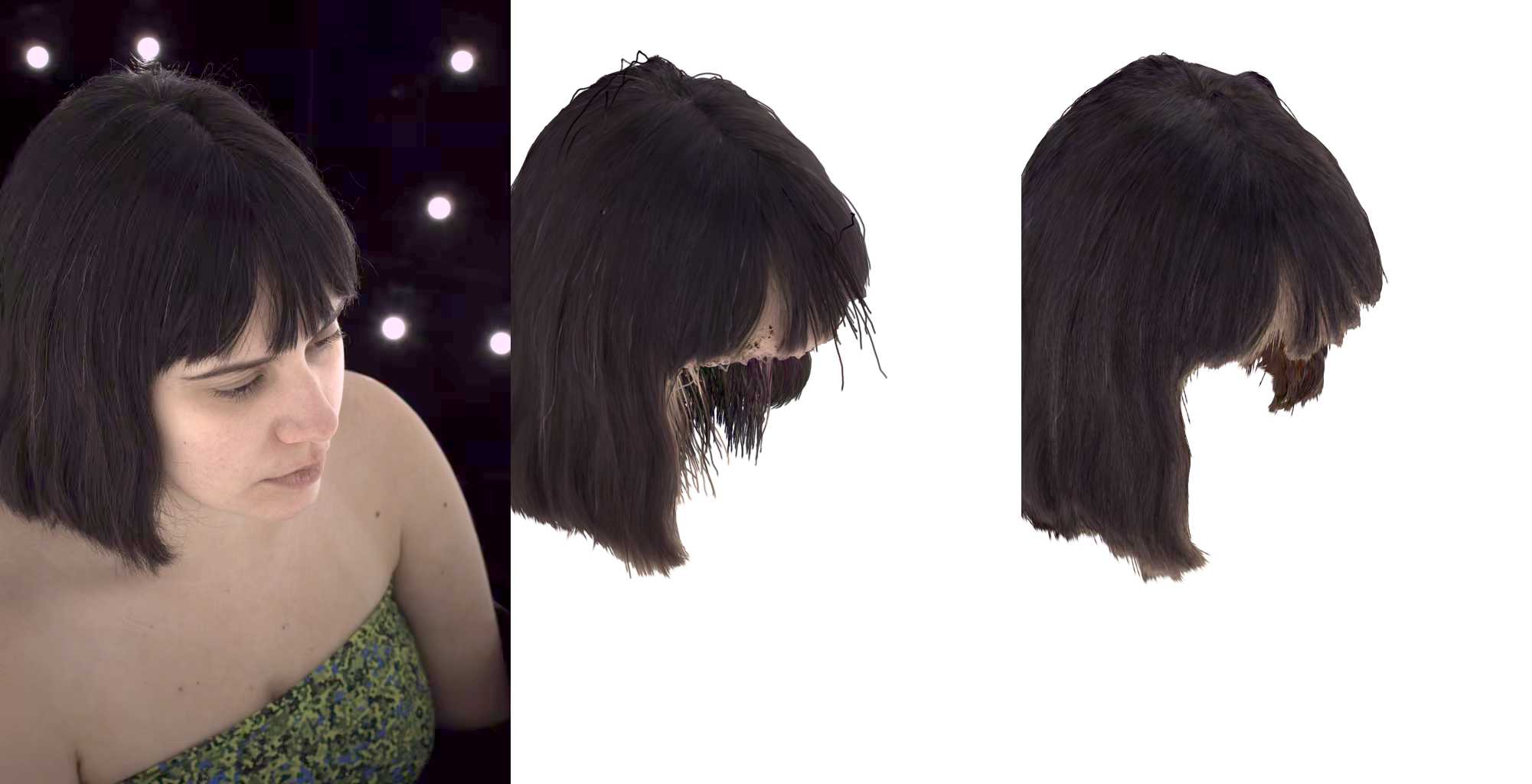}
\end{minipage}

\vspace{0.5mm}

\begin{minipage}{0.90\linewidth}
    \includegraphics[width=\linewidth,
        trim=0cm 4cm 0cm 1cm, clip]{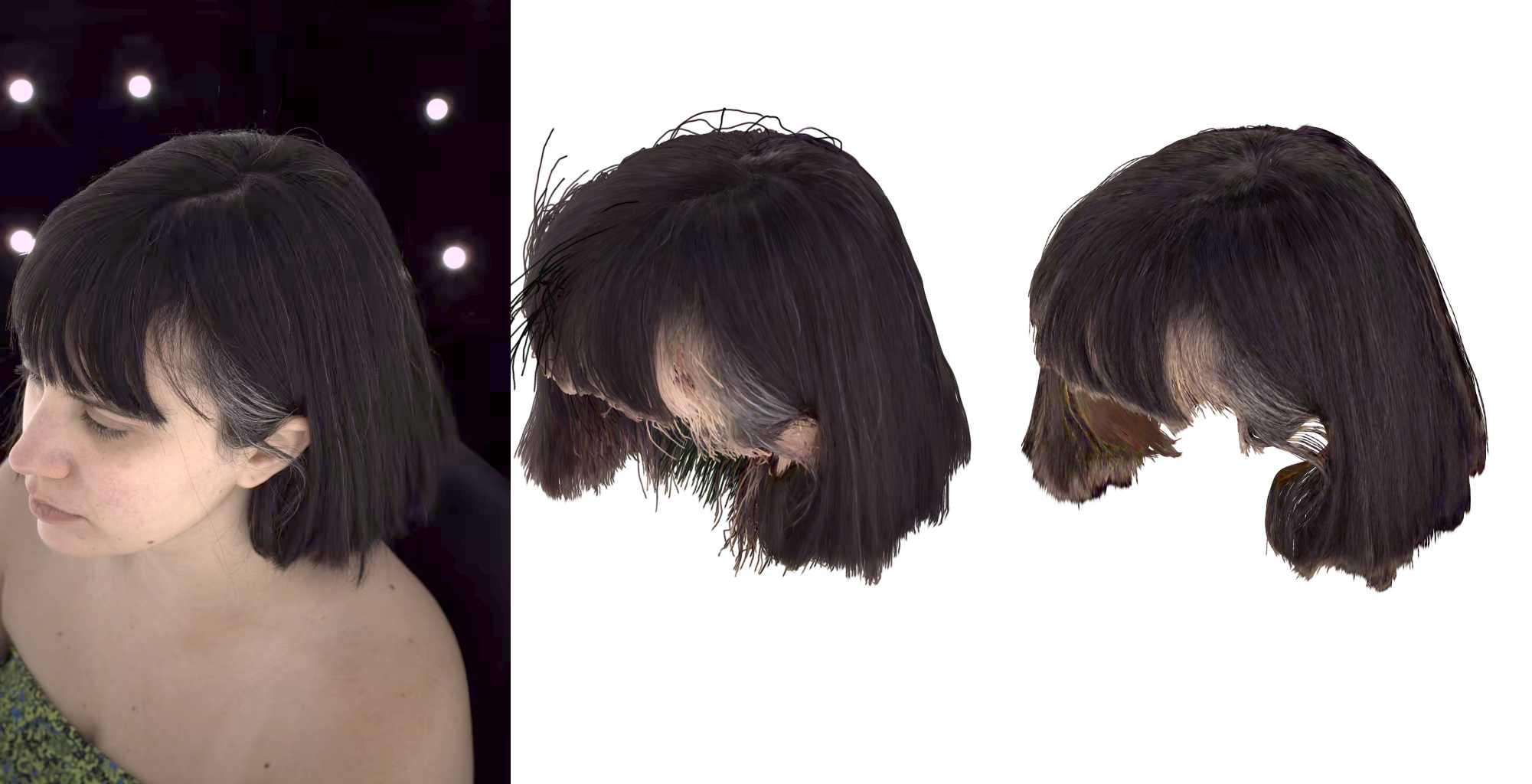}
\end{minipage}

\vspace{0.5mm}

\begin{minipage}{0.90\linewidth}
    \includegraphics[width=\linewidth,
        trim=0cm 4cm 0cm 1cm, clip]{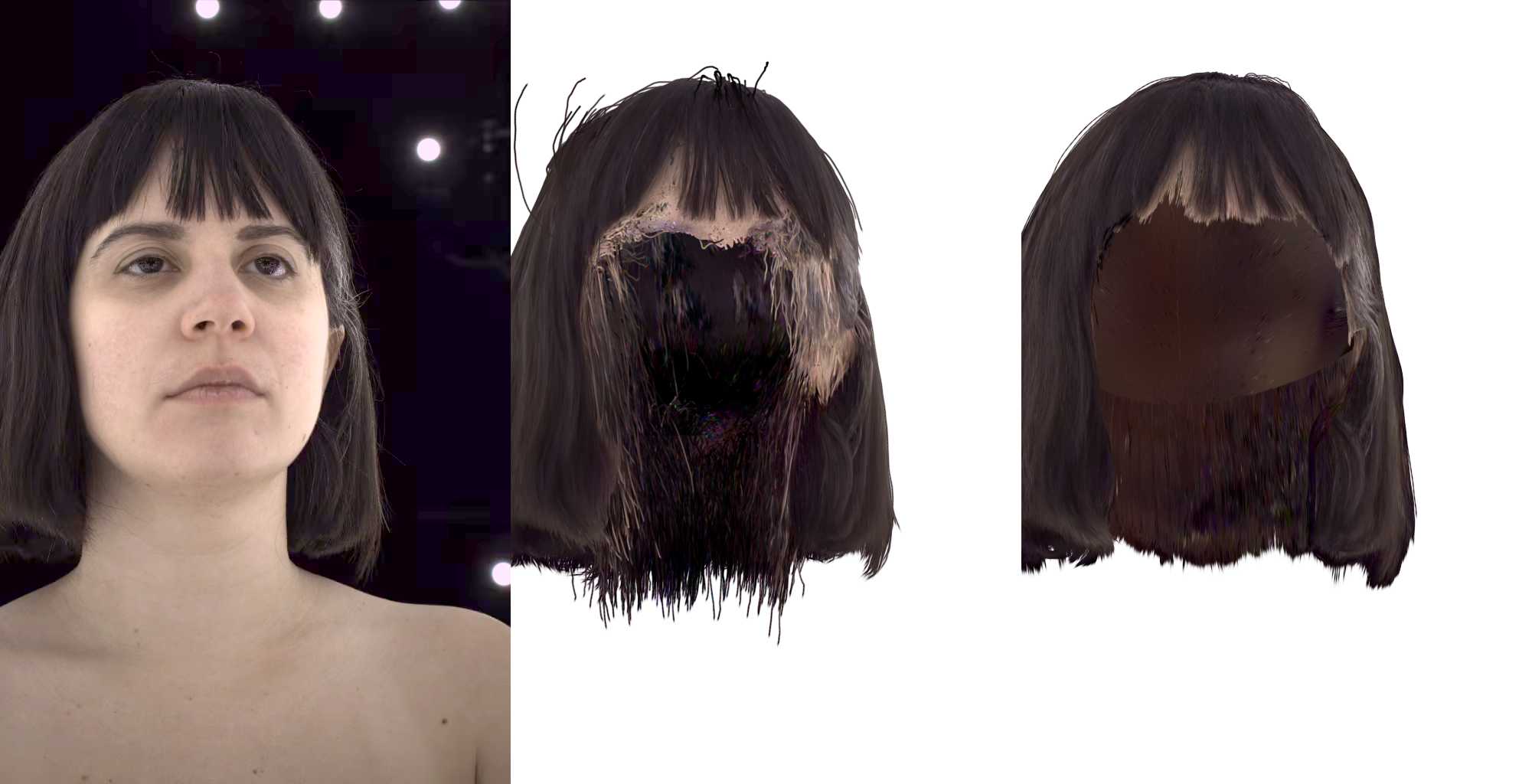}
\end{minipage}

\vspace{0.5mm}

\vspace{2mm}
\noindent
\makebox[\linewidth][c]{%
\parbox{0.32\linewidth}{\centering Ground Truth}%
\hfill
\parbox{0.32\linewidth}{\centering Gaussian Haircut}%
\hfill
\parbox{0.32\linewidth}{\centering PhysHead (Ours)}%
}

\caption{\textbf{Extended Gaussian Haircut (GH)~\cite{zakharov2024gh} comparison.} Gaussian Haircut captures the overall outer hair color nicely but exhibits artifacts in invisible regions. Our method propagates the outer color into inner strands, enabling consistent hair appearance.}
\label{fig:gh_additional}

\end{figure}

First, we compare our learned hair appearance with Gaussian Haircut~\cite{zakharov2024gh} in Figure~\ref{fig:gh_additional}, showing that our regularization loss significantly improves the realism of hair, particularly its internal structure. Finally, we present additional comparisons of dynamic hair appearance with Gaussian Avatars (GA)~\cite{Qian2024gaussianavatars} and Gaussian Head Avatars (GHA)~\cite{xu2023gaussianheadavatar} in Figure~\ref{fig:maincomparison_supmat}. The main advantage of our physics-guided hair geometry is its ability to generalize to novel poses, in contrast to prior works that produce nearly static hair.

\subsection{Single Layer Representation}
\begin{figure}[h]
  \centering
  \includegraphics[width=\linewidth]{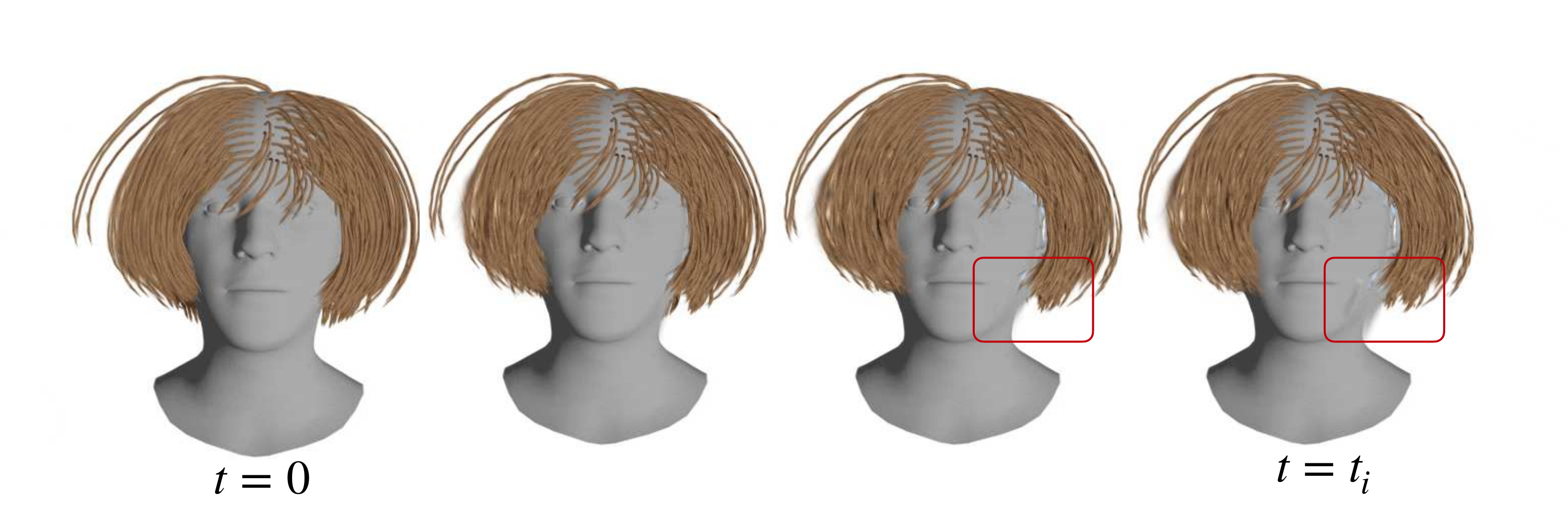}
  \caption{\textbf{Single layer animation.} Single layer of optimization results in artifacts during animation despite the usage of hair strands. }
  \label{fig:singlelayer}
\end{figure}

Jointly optimizing hair and head appearance in a single layer can produce artifacts during animation. Specifically, when animating strands with the physics engine, the skin may appear to peel along with the hair, as shown in Figure~\ref{fig:singlelayer}. To prevent this, we fully disentangle hair and head by optimizing the hair and head Gaussian layers separately, keeping the head Gaussian parameters fixed while optimizing hair.

\subsection{Limitations}
\begin{figure}[h]
  \centering
  \includegraphics[width=\linewidth]{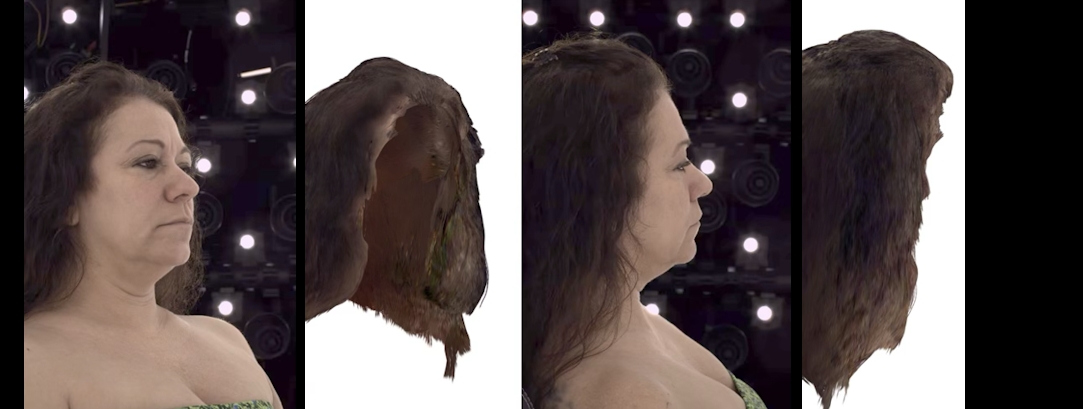}
  \caption{\textbf{Curly Hair} Curly Hair results with GT and Ours. }
  \label{fig:curly}
\end{figure}

PhysHead operates on reconstructed strand geometry. In our experiments, we utilize NH~\cite{sklyarova2023neural}, which fails to reconstruct curly hair, shown in Figure~\ref{fig:curly}. However, our method is agnostic to the specific reconstruction method, benefits directly from future improvements.

\clearpage

\end{document}